%% file: main.tex
\documentclass[10pt,twocolumn,letterpaper]{article}
\PassOptionsToPackage{dvipsnames,table}{xcolor}

\usepackage{iccv}              

\usepackage{times}
\usepackage{graphicx}
\usepackage{amsmath}
\usepackage{amssymb}
\usepackage{booktabs}
\usepackage{rotating}
\usepackage{paralist}
\usepackage{caption}
\usepackage{arydshln}
\usepackage{graphicx}
\usepackage{amsmath}
\usepackage{amssymb}
\usepackage{color, colortbl}

\input{preamble}

\definecolor{iccvblue}{rgb}{0.21,0.49,0.74}
\definecolor{LightCyan}{rgb}{0.88,1,1}
\usepackage[pagebackref,breaklinks,colorlinks,allcolors=iccvblue]{hyperref}
\usepackage{makecell}
\usepackage[accsupp]{axessibility}
\hypersetup{
    colorlinks=true,
    urlcolor=magenta,
    citecolor=blue,
}

\usepackage[T1]{fontenc}
\usepackage{multirow}
\usepackage{booktabs}
\usepackage{comment}
\usepackage{color}
\usepackage{amsmath}

\usepackage{textcomp}
\usepackage{siunitx}
\usepackage{tabulary}
\usepackage{makecell}
\usepackage{multirow}
\usepackage{booktabs}
\usepackage{hyperref}

\makeatletter
\newcommand\figref{Figure~\ref}
\newcommand{\tabref}[1]{Table~\ref{#1}}
\newcolumntype{P}[1]{>{\centering\arraybackslash}p{#1}}
\newcolumntype{M}[1]{>{\centering\arraybackslash}m{#1}}
\let\ts@includegraphics\includegraphics

\usepackage{float}
\usepackage{lipsum}   
\usepackage{rotating} 
\usepackage{stfloats} 
\usepackage[export]{adjustbox} 

\usepackage[capitalize]{cleveref}
\crefname{section}{Sec.}{Secs.}
\Crefname{section}{Section}{Sections}
\Crefname{table}{Table}{Tables}
\crefname{table}{Tab.}{Tabs.}

\def\paperID{4541} 
\def\confName{ICCV}
\def\confYear{2025}

\title{Exploring Probabilistic Modeling Beyond Domain Generalization for Semantic Segmentation}

\vspace{-20mm}\author{
    I-Hsiang Chen\textsuperscript{1,2*,†}
    \quad Hua-En Chang\textsuperscript{1*}
    \quad Wei-Ting Chen\textsuperscript{3}
    \quad Jenq-Neng Hwang\textsuperscript{2} 
    \quad Sy-Yen Kuo\textsuperscript{1,4} \\
    \\
    \textsuperscript{1}National Taiwan University\quad \textsuperscript{2}University of Washington\quad \textsuperscript{3}Microsoft\quad\textsuperscript{4}Chang Gung University
}

\begin{document}

\twocolumn[{%
\maketitle
\vspace{-5mm}
\begin{figure}[H]
    \centering
    \hsize=\textwidth 
    \footnotesize
    \begin{subfigure}{1.33\linewidth}     
       \setlength{\tabcolsep}{2pt}
       \begin{tabular}{cccc}
        Input & DeepLabV3Plus & PDAF & Ground Truth \\
        \includegraphics[width=.245\textwidth, height=1.6cm]{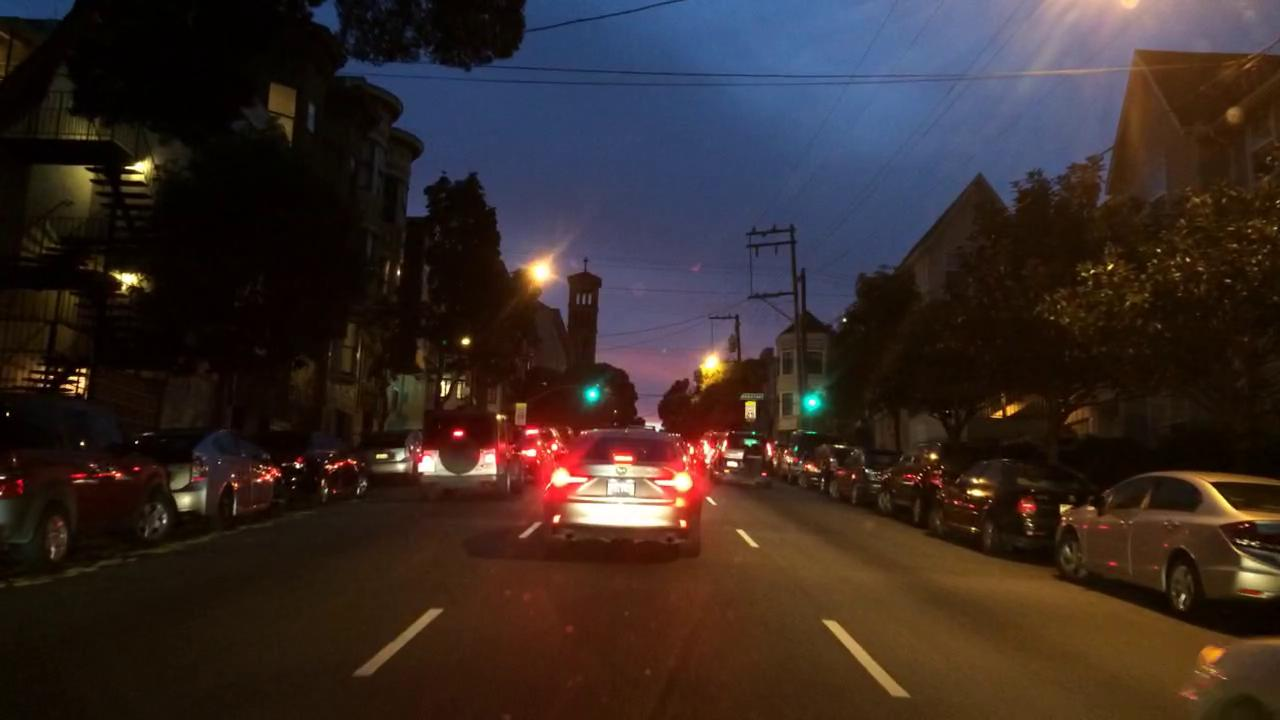} &
        \includegraphics[width=.245\textwidth, height=1.6cm]{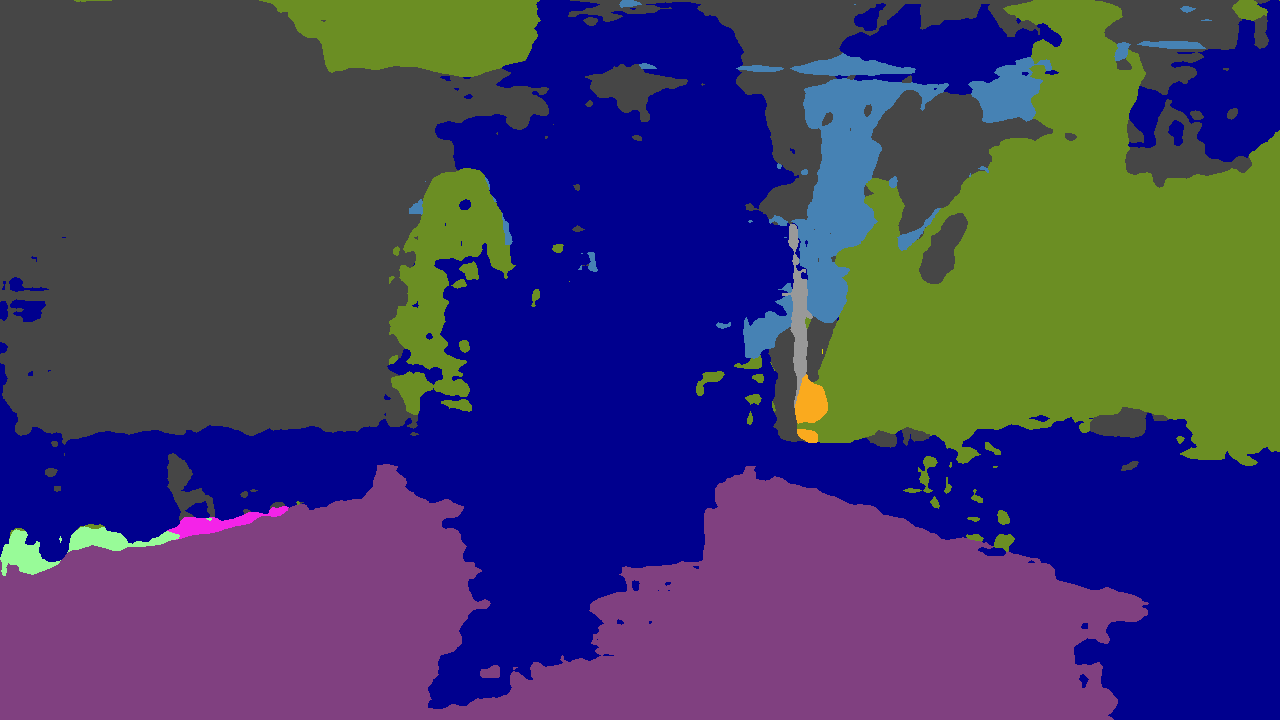} &
        \includegraphics[width=.245\textwidth, height=1.6cm]{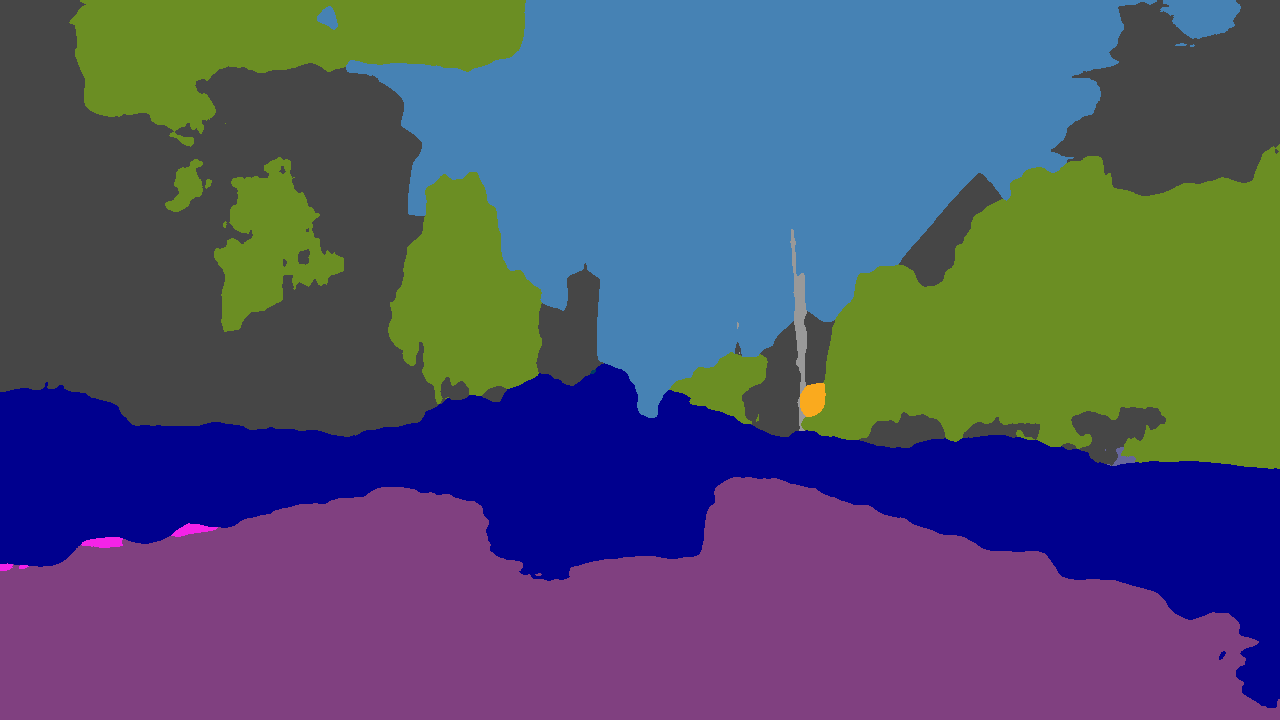} &
        \includegraphics[width=.245\textwidth, height=1.6cm]{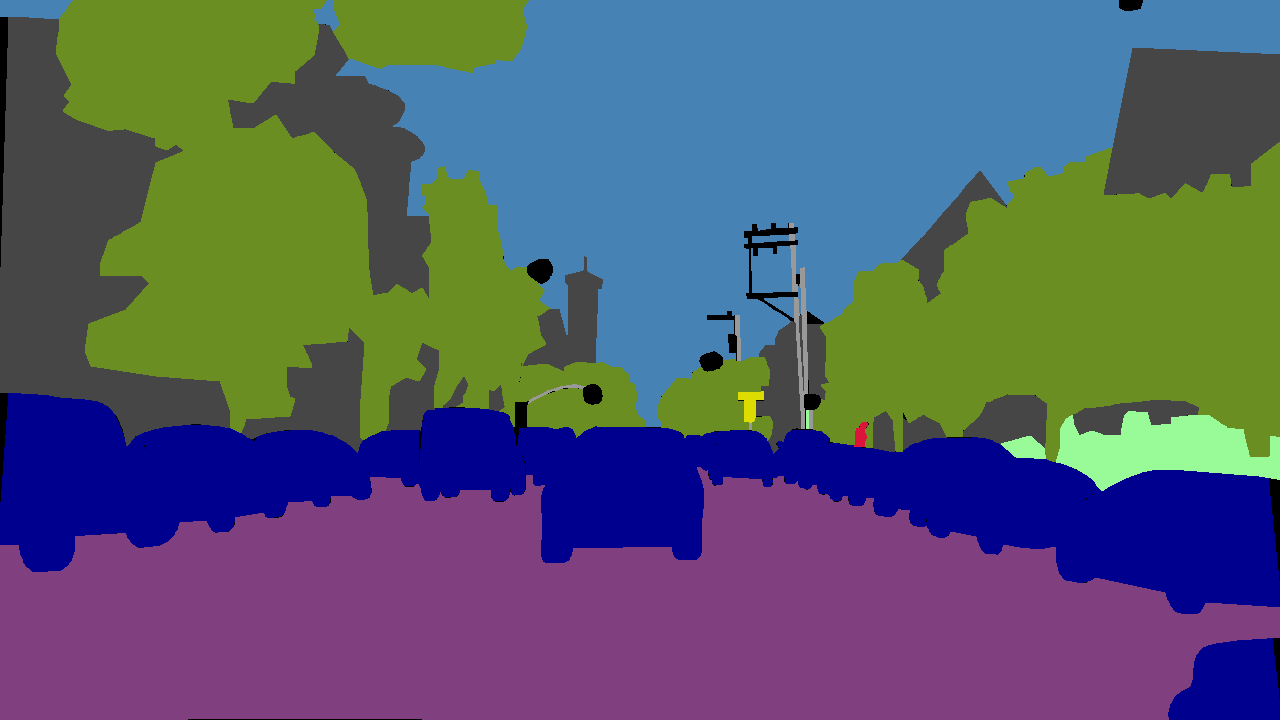} \\
        \includegraphics[width=.245\textwidth, height=1.6cm]{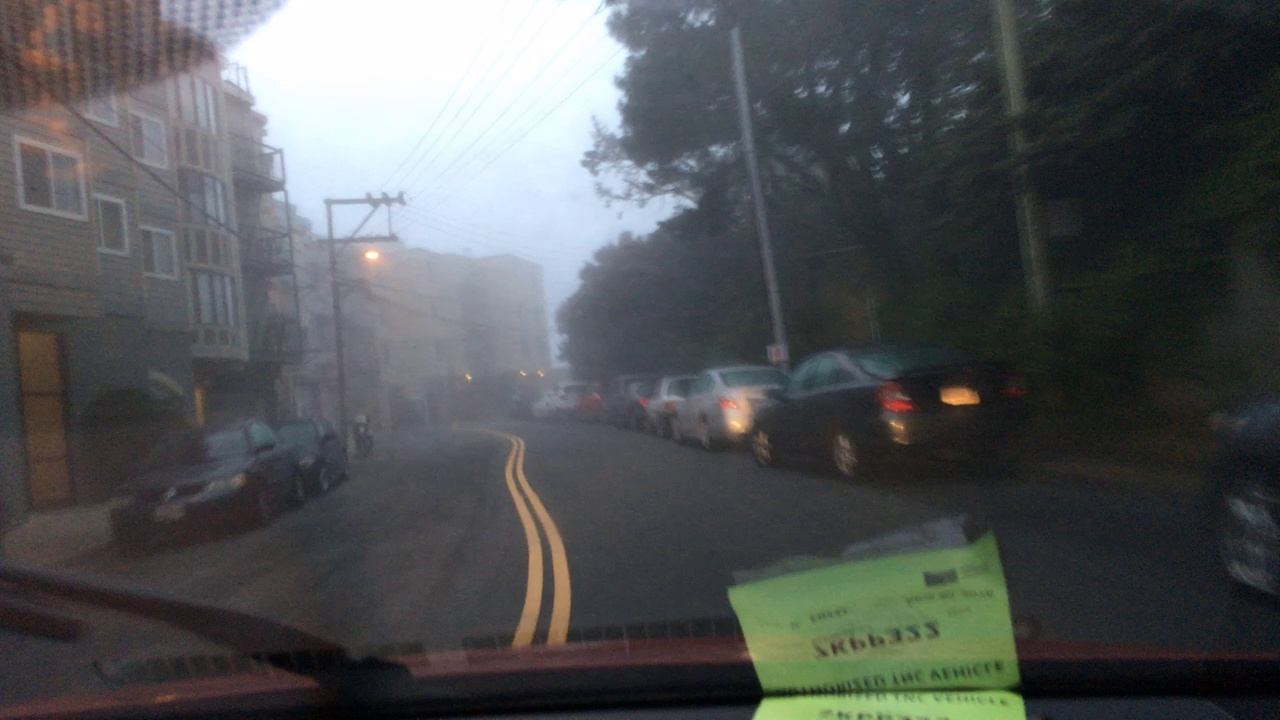} &
        \includegraphics[width=.245\textwidth, height=1.6cm]{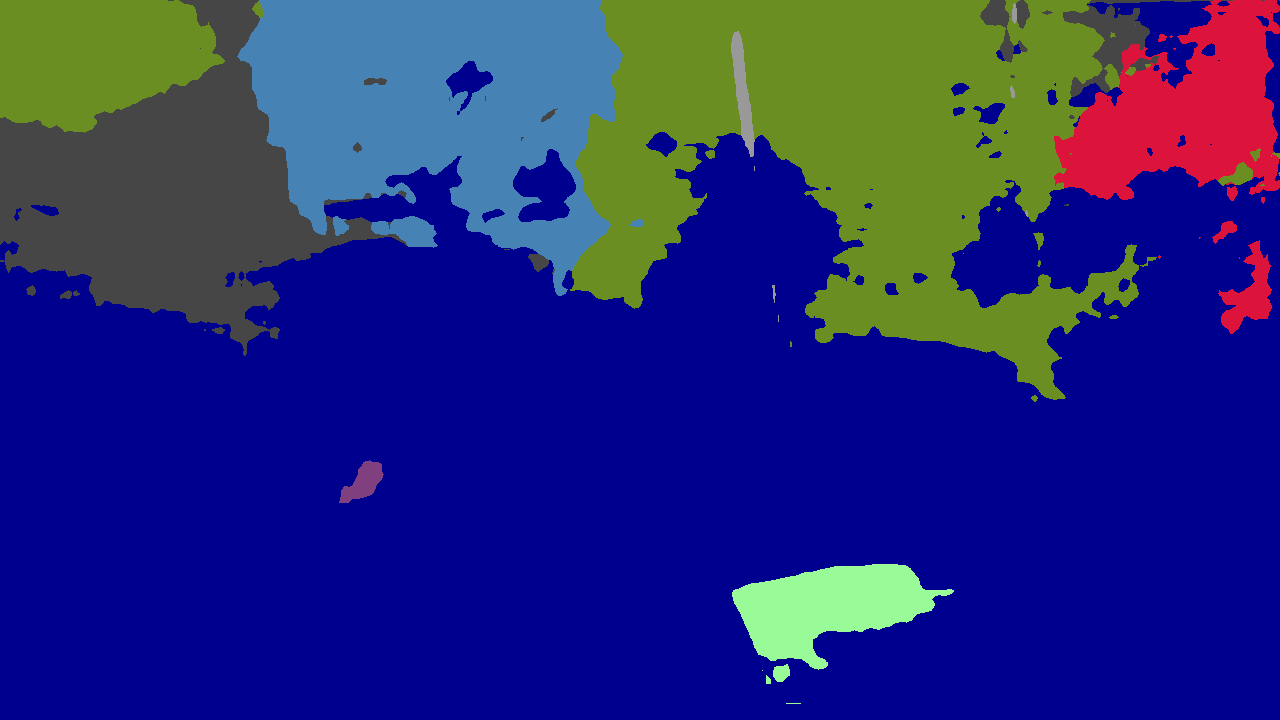} &
        \includegraphics[width=.245\textwidth, height=1.6cm]{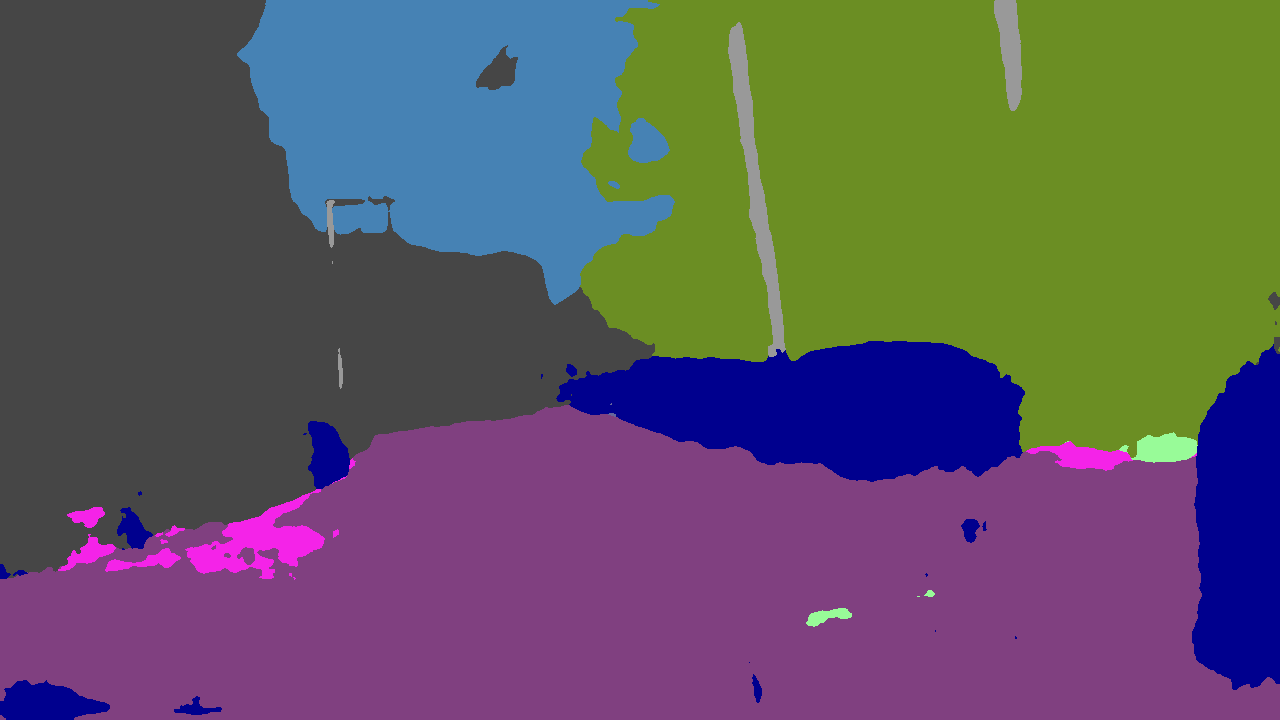} &
        \includegraphics[width=.245\textwidth, height=1.6cm]{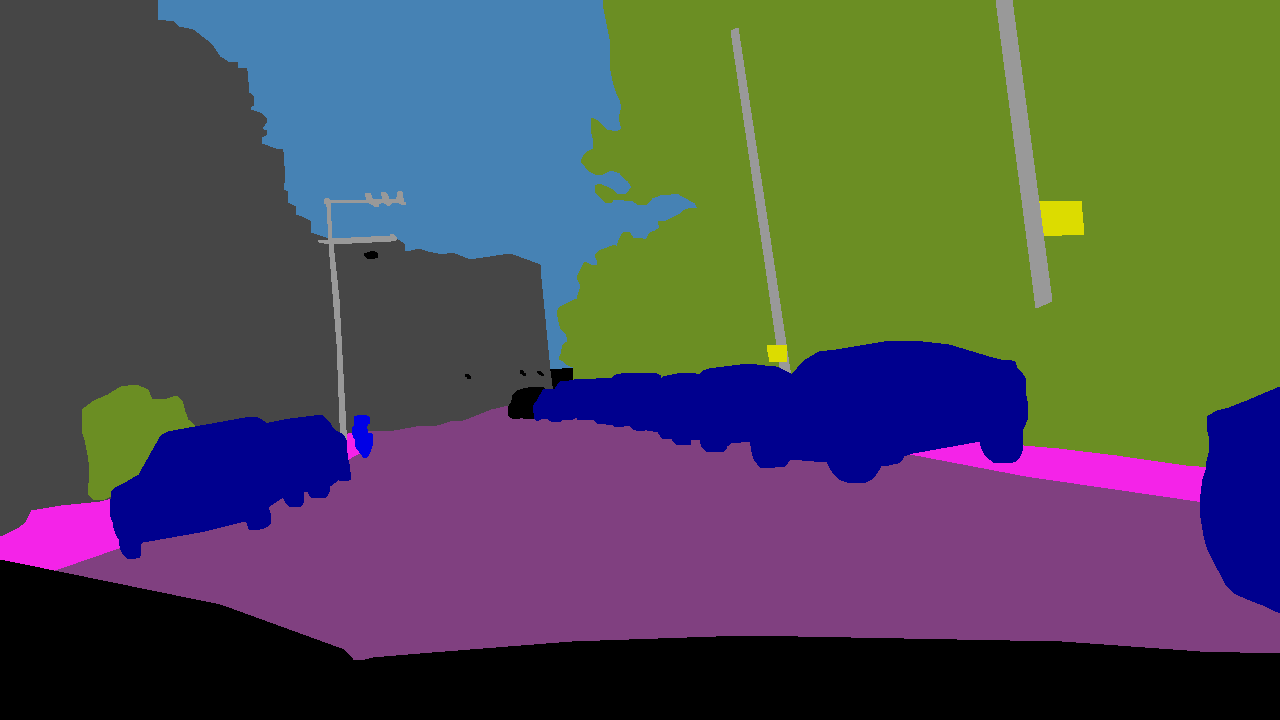} \\
        Input & Mask2Former & PDAF & Ground Truth \\
        \includegraphics[width=.245\textwidth, height=1.6cm]{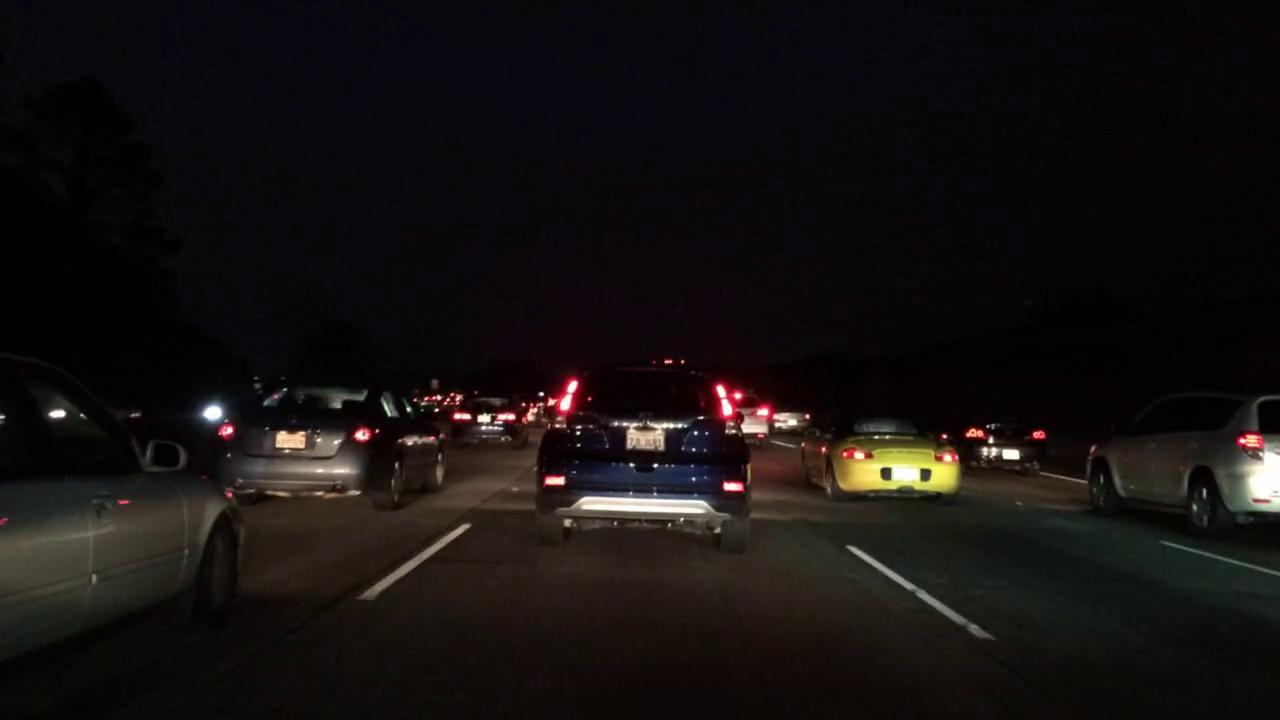} &
        \includegraphics[width=.245\textwidth, height=1.6cm]{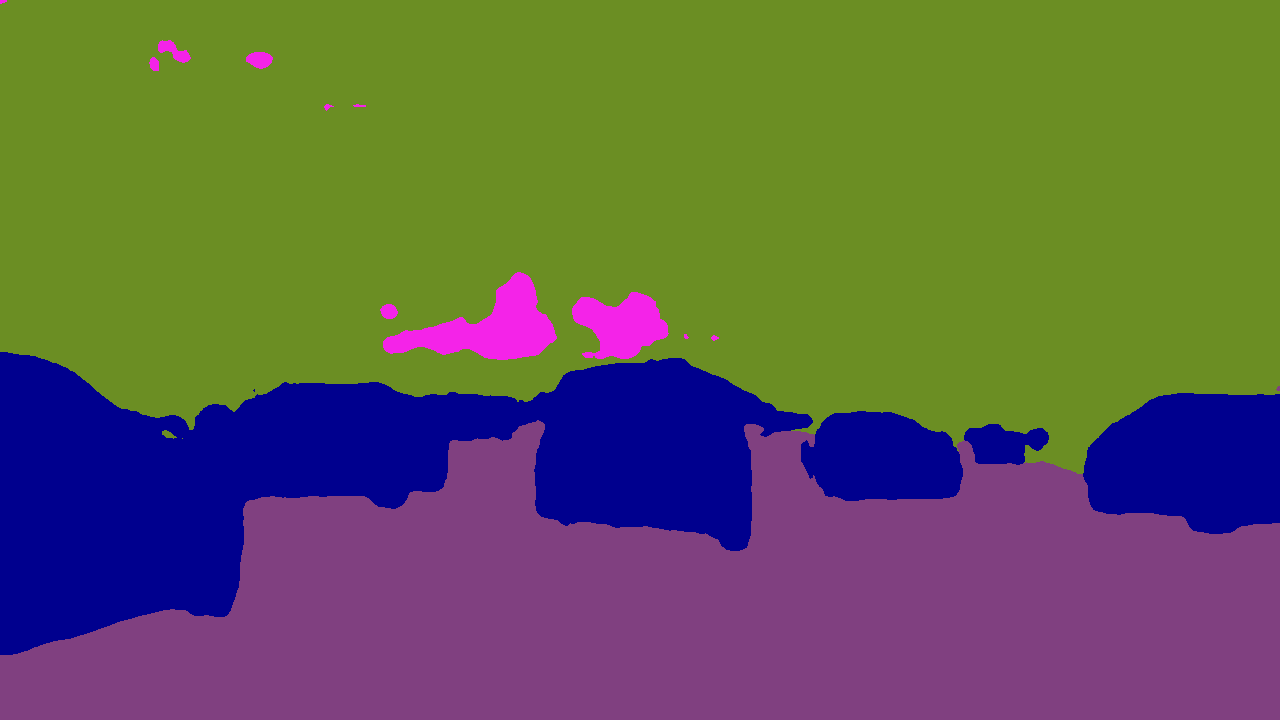} &
        \includegraphics[width=.245\textwidth, height=1.6cm]{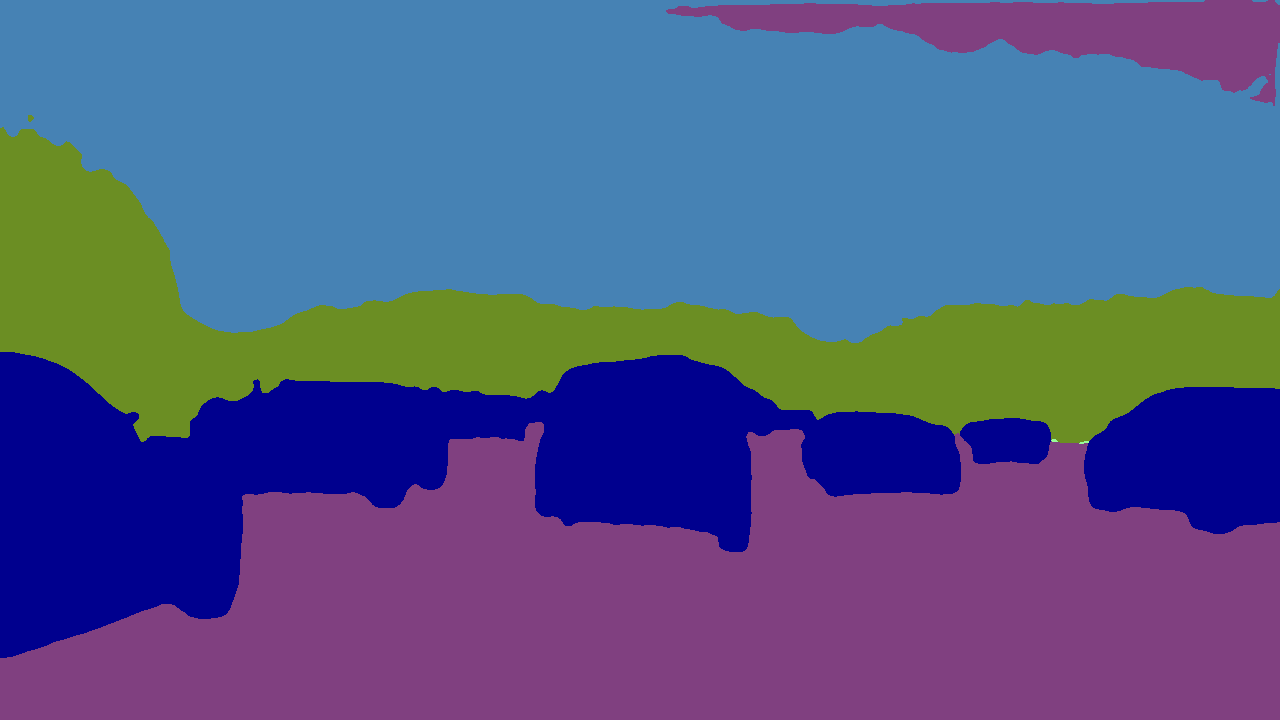} &
        \includegraphics[width=.245\textwidth, height=1.6cm]{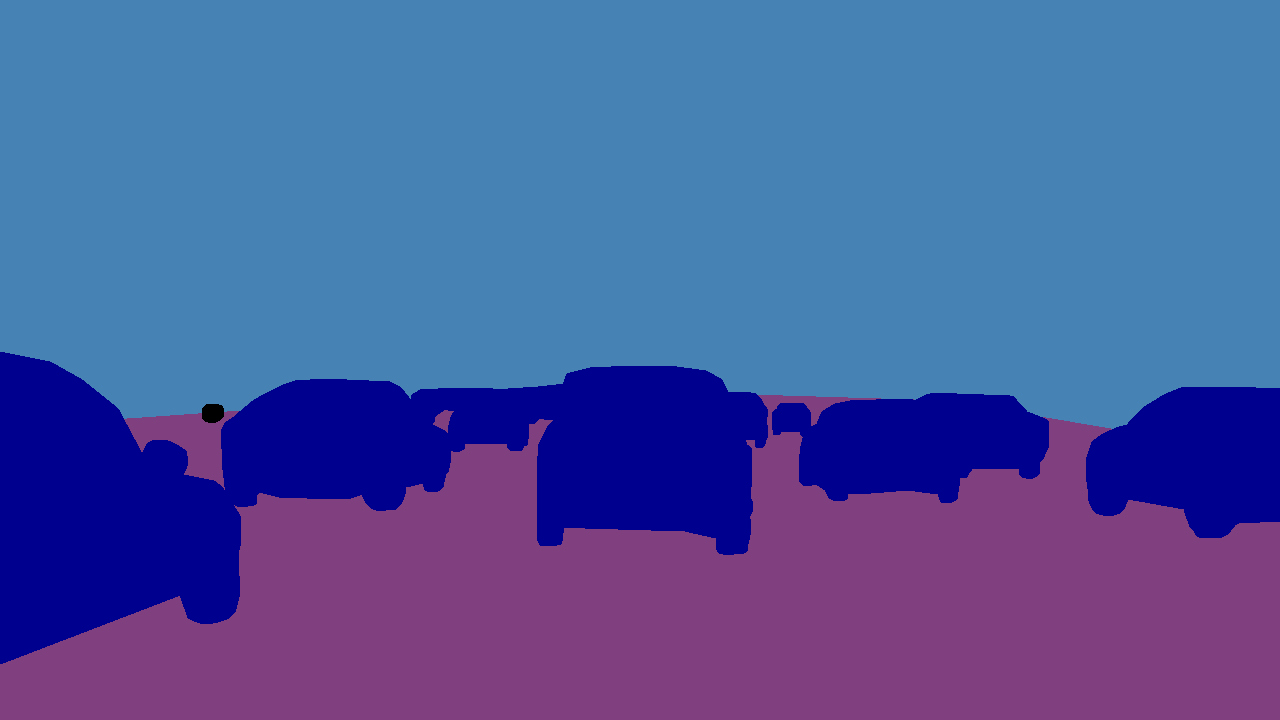} \\
        \includegraphics[width=.245\textwidth, height=1.6cm]{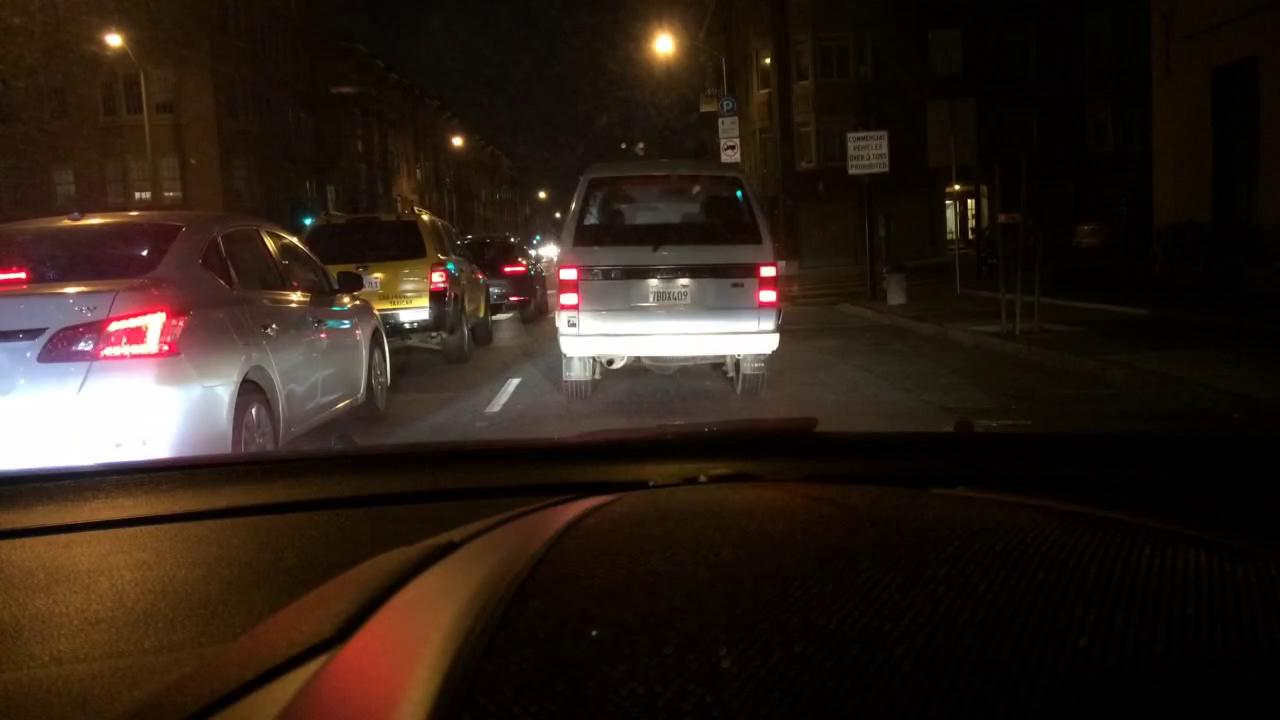} &
        \includegraphics[width=.245\textwidth, height=1.6cm]{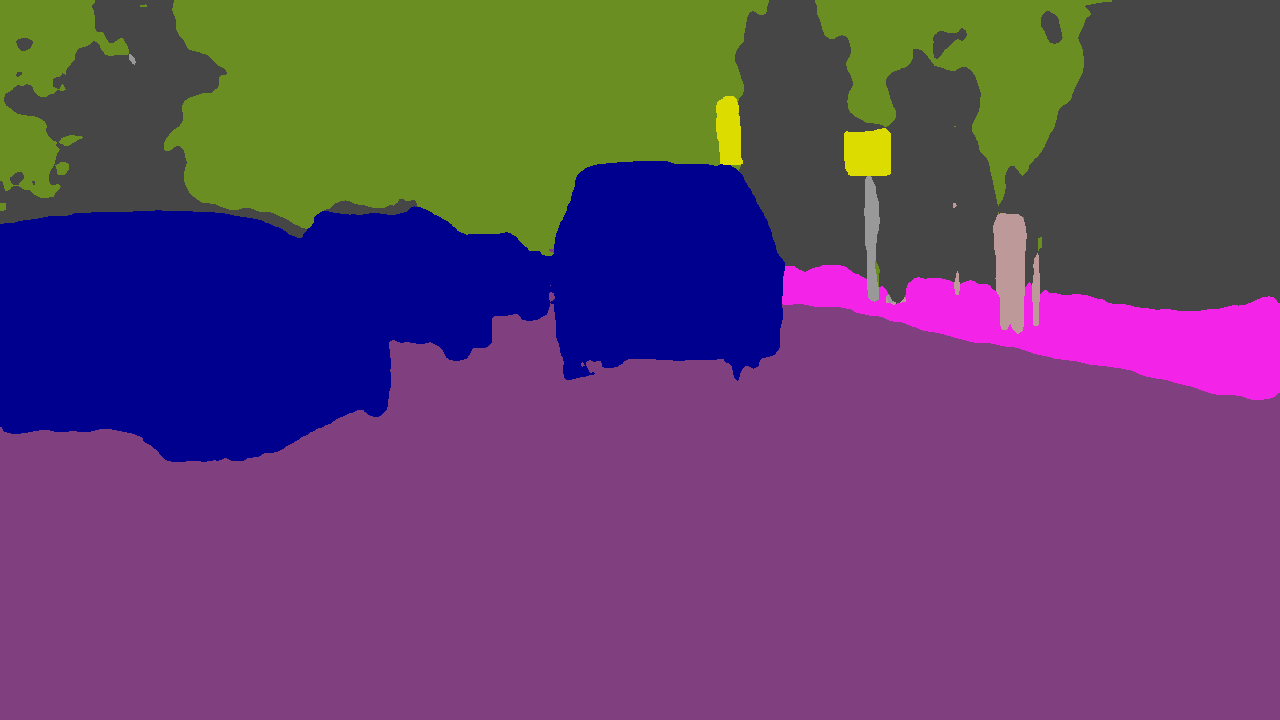} &
        \includegraphics[width=.245\textwidth, height=1.6cm]{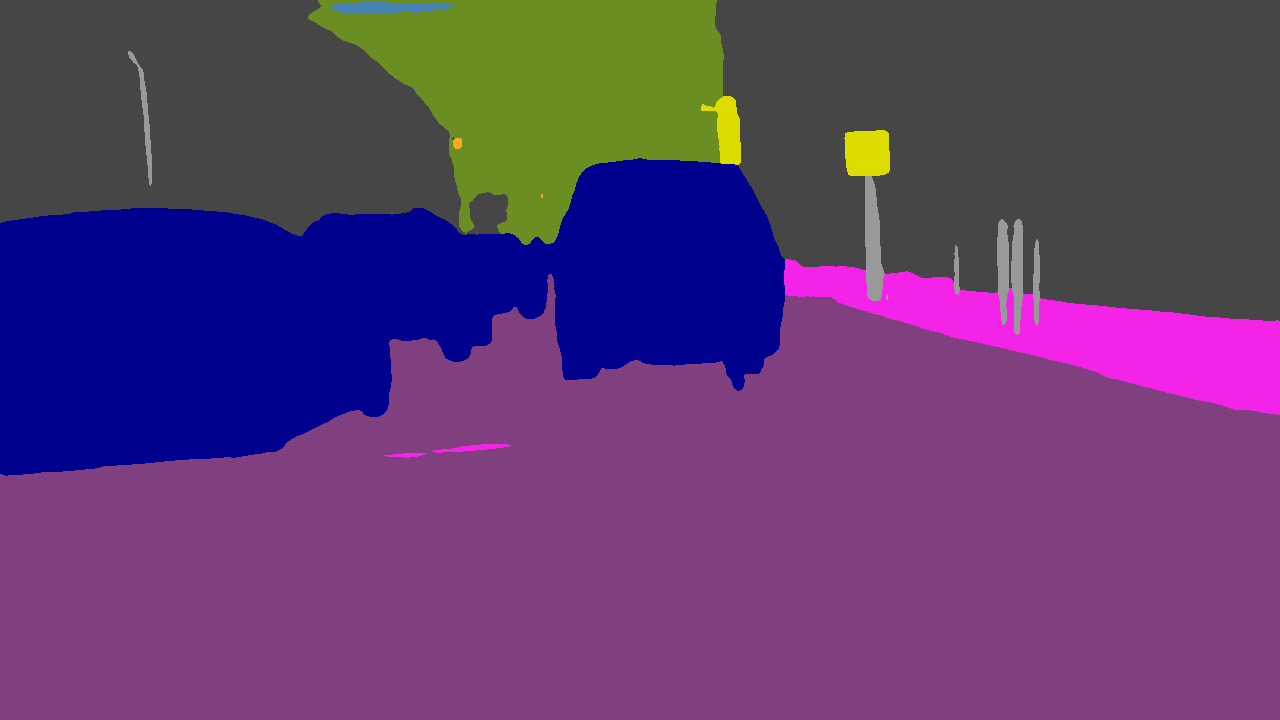} &
        \includegraphics[width=.245\textwidth, height=1.6cm]{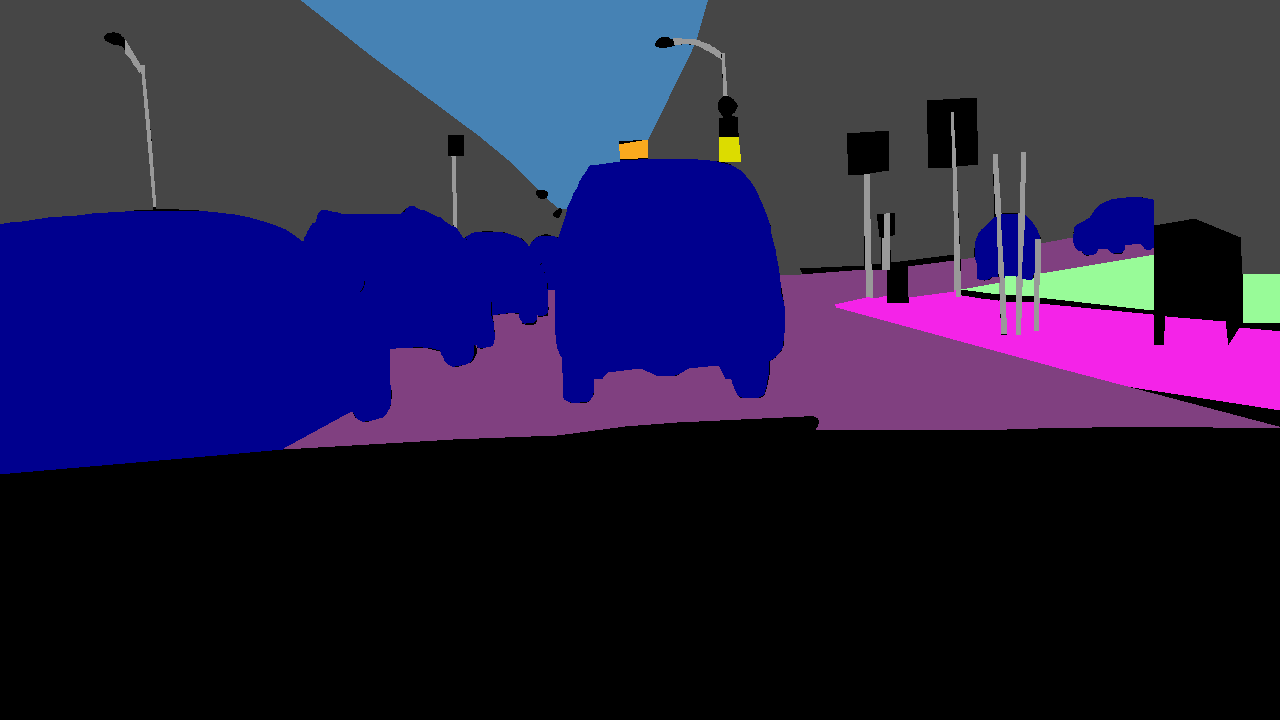} \\
      \end{tabular}
      \caption{Qualitative comparison between \method{} and its backbones}
      \label{fig:teaser_samples}
    \end{subfigure}
    \hfill
    \begin{subfigure}{0.65 \linewidth}
        \includegraphics[width=1.0\linewidth]{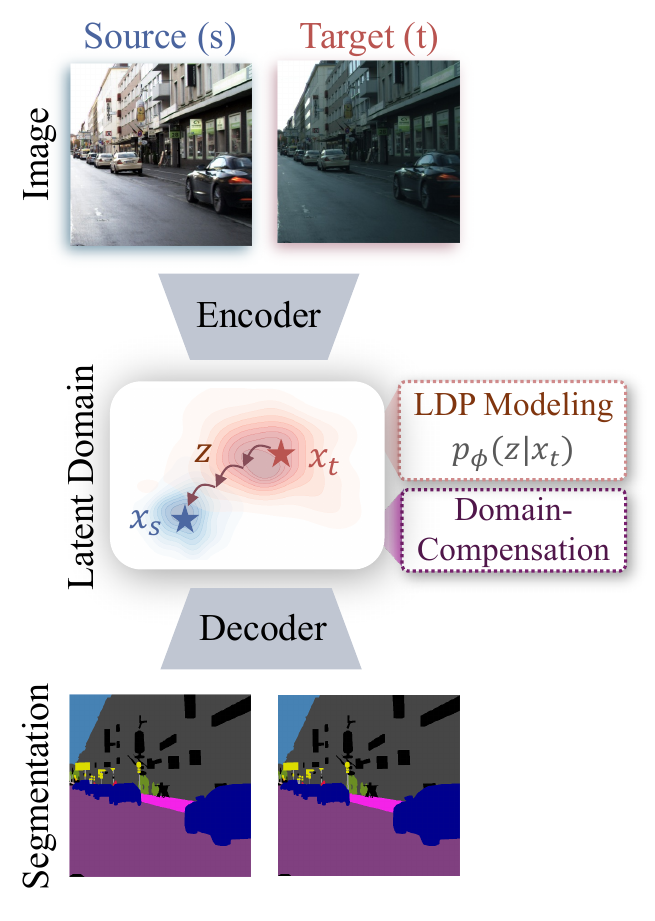}
        \caption{Overview of PDAF framework.}
        \label{fig:teaser_pdaf}
    \end{subfigure}
    \caption{The proposed \method{} enhances existing segmentation backbones (DeepLabV3Plus~\cite{Chen2017RethinkingAC} and Mask2Former~\cite{Cheng2021MaskedattentionMT}) by leveraging latent domain prior modeling as a compensation mechanism, yielding consistent improvements on unseen target domains such as BDD100K~\cite{Yu2018BDD100KAD}.}
    \label{fig:teaser}
\end{figure}
}]

\newcommand\blfootnote[1]{%
\begingroup
\renewcommand\thefootnote{}\footnote{#1}%
\addtocounter{footnote}{-1}%
\endgroup
}
\blfootnote{Project page: \url{https://pdaf-iccv.github.io}}
\blfootnote{* Indicates equal contribution.}
\blfootnote{† This work was conducted while I-Hsiang Chen was an intern at University of Washington.}

\input{Abstract}    
\input{Introduction}
\input{Related_Work}

\input{Method}
\input{Implementation_Detail}
\input{Experimental_Results}
\input{Conclusion}
\section*{Acknowledgements}
We thank to National Center for High-performance Computing (NCHC) for
providing computational and storage resources. This research was supported by Taiwan's National Science and Technology Council under Grant NSTC 111-2221-E-002-136-MY3 and NSTC 114-2221-E-002-067-MY3.

{
    \small
    \bibliographystyle{ieeenat_fullname}
    \bibliography{main}
}
\end{document}

%% file: preamble.tex
\newcommand{\method}[1]{PDAF}
\newcommand{\LDP}[1]{LDP}

\newcommand{\LPE}[1]{LPE} 
\newcommand{\DCM}[1]{DCM} 
\newcommand{\DPE}[1]{DPE} 

\newcommand{\C}[1]{Cityscapes}
\newcommand{\B}[1]{BDD-100K}
\newcommand{\M}[1]{Mapillary}
\newcommand{\G}[1]{GTAV}
\newcommand{\SYN}[1]{SYNTHIA}


%% file: Abstract.tex
\begin{abstract}

Domain Generalized Semantic Segmentation (DGSS) is a critical yet challenging task, as domain shifts in unseen environments can severely compromise model performance. 
While recent studies enhance feature alignment by projecting features into the source domain, they often neglect intrinsic latent domain priors, leading to suboptimal results.
In this paper, we introduce \method{}, a Probabilistic Diffusion Alignment Framework that enhances the generalization of existing segmentation networks through probabilistic diffusion modeling. 
\method{} introduces a Latent Domain Prior (\LDP{}) to capture domain shifts and uses this prior as a conditioning factor to align both source and unseen target domains.
To achieve this, \method{} integrates into a pre-trained segmentation model and utilizes paired source and pseudo-target images to simulate latent domain shifts, enabling \LDP{} modeling.
The framework comprises three modules: the Latent Prior Extractor (\LPE{}) predicts the \LDP{} by supervising domain shifts; the Domain Compensation Module (\DCM{}) adjusts feature representations to mitigate domain shifts; and the Diffusion Prior Estimator (\DPE{}) leverages a diffusion process to estimate the \LDP{} without requiring paired samples. 
This design enables \method{} to iteratively model domain shifts, progressively refining feature representations to enhance generalization under complex target conditions.
Extensive experiments validate the effectiveness of \method{} across diverse and challenging urban scenes.

\end{abstract}

%% file: Introduction.tex
\section{Introduction}
Domain Generalized Semantic Segmentation (DGSS) remains a fundamental challenge in computer vision, primarily due to inevitable domain shifts encountered in unseen environments~\cite{qu2024lead, zhang2024improving}.
Segmentation models trained on a source domain frequently degrade on the unseen target domains~\cite{Balaji2018MetaRegTD, Li2018DeepDG, chen2022rvsl}. 
This challenge is particularly critical in real-world applications such as autonomous driving~\cite{bartoccioni2023lara, hu2023planning} and robotic systems~\cite{nilsson2021embodied, onozuka2021autonomous}, where accurate semantic segmentation is essential for ensuring operational safety.
Factors such as lighting variations, weather changes, and other underrepresented conditions in training data exacerbate the generalization problem. 
Addressing the domain shifts thus remains a key obstacle in developing robust and reliable semantic segmentation models.

Within the field of DGSS, two primary methodologies have emerged: data augmentation-based approaches and domain-invariant representation learning-based approaches.
Data augmentation-based methods enhance the diversity of training data by introducing synthetic variations, thereby exposing models to a broader spectrum of potential domain shifts and improving their generalization ability~\cite{Huang2021FSDRFS, Zhong2022AdversarialSA, Zhao2022StyleHallucinatedDC}. Nevertheless, a crucial limitation of data augmentation-based methods is their significant dependence on auxiliary domains or generation models~\cite{ganin2015unsupervised, liu2024unbiased, jia2024dginstyle}. 
Conversely, domain-invariant representation learning methods aim to train models that extract consistent features across different domains, effectively mitigating the impact of domain shift~\cite{Seo2019LearningTO, Li2018DomainGW, Dou2019DomainGV, Motiian2017UnifiedDS, Ghifary2015DomainGF, chen2022sjdl, Li2017LearningTG, Balaji2018MetaRegTD}. However, the inherent entanglement between style and content poses a significant challenge, often resulting in the loss of critical semantic information during feature decomposition~\cite{Huang2023StylePC, Yang2023GeneralizedSS, Ahn2024StyleBD}.
Recent works achieve feature alignment by projecting features into a constrained feature space~\cite{Huang2023StylePC, Yang2023GeneralizedSS, Ahn2024StyleBD}, but they often overlook the intrinsic properties of latent domain priors, leading to suboptimal performance~\cite{venkat2020your, wang2024domain}.

To overcome this limitation, we propose \method{}, a Probabilistic Diffusion Alignment Framework that enhances the generalization of existing segmentation networks through probabilistic diffusion modeling. Specifically, \method{} introduces a Latent Domain Prior (\LDP{}), which models domain-specific variations and guides feature alignment between source and unseen target domains, as illustrated in~\figref{fig:teaser}.
Specifically, \method{} integrates into a pre-trained segmentation model and leverages paired source and pseudo-target images to simulate domain shifts. This process enables \LDP{} modeling, which serves as a foundation for robust domain alignment.
To leverage these priors effectively, \method{} first utilizes the Latent Prior Extractor (\LPE{}) to capture cross-domain relationships between the source and pseudo-target samples, generating an optimal \LDP{} that encodes domain-specific variations. 
Building upon this, the Domain Compensation Module (\DCM{}) conditions feature representations on the extracted \LDP{}, ensuring that domain priors guide the adaptation process to mitigate domain shifts while preserving task-relevant information. 
However, the direct estimation of \LDP{} from explicit source-target pairs may be infeasible in real-world scenarios; to overcome this, the Diffusion Prior Estimator (\DPE{}) employs probabilistic diffusion modeling to infer \LDP{} without requiring paired samples, enhancing the model’s adaptability to unseen domains.
Through this progressive refinement, \method{} not only aligns features across domains but also maintains critical variations essential for generalization, leading to improved segmentation performance in complex environments.

Extensive experiments on benchmark datasets validate the effectiveness of \method{}, demonstrating that probabilistic modeling of \LDP{} successfully enhances the domain generalization of existing segmentation networks.

Our main contributions are as follows:
\begin{compactitem}
\item We introduce \method{}, a novel probabilistic diffusion alignment framework that leverages probabilistic diffusion modeling of latent domain priors (\LDP{}) to enhance the robustness of existing segmentation models against unseen domain shifts.

\item We propose three components for \method{}: the latent prior extractor, the domain compensation module, and the diffusion prior estimator. These modules collectively estimate and leverage the \LDP{} for robust feature alignment between source and unseen target domains.

\item Extensive experiments across diverse datasets and degradation scenarios demonstrate that our method effectively improves the domain generalization of two widely adopted segmentation models (\eg, DeepLabV3Plus~\cite{Chen2017RethinkingAC} and Mask2Former~\cite{Cheng2021MaskedattentionMT}), achieving state-of-the-art performance.
\end{compactitem}

%% file: Related_Work.tex
\section{Related Work}
\noindent\textbf{Domain Generalized Semantic Segmentation.}
Domain Generalized Semantic Segmentation (DGSS) enhances model robustness to unseen domains without requiring target data. Existing approaches fall into data augmentation-based methods and domain-invariant representation learning-based methods.

Data augmentation-based methods improve generalization by synthesizing or randomizing styles. DRPC~\cite{Yue2019DomainRA} enforces pyramid consistency across stylized versions using real images as style references. GTR-LTR~\cite{peng2021global} replaces source styles with artistic paintings for global and local texture variation, while WildNet~\cite{Lee2022WildNetLD} enhances feature-level representations with ImageNet~\cite{5206848} as an external data source. Domain-invariant representation learning-based methods remove domain-specific information via normalization and whitening. IBN-Net~\cite{Pan2018TwoAO} integrates Instance Normalization (IN) and Batch Normalization (BN)~\cite{Ioffe2015BatchNA}, while Switchable Whitening (SW)~\cite{Pan2019SwitchableWF} adaptively merges whitening and standardization. ISW~\cite{choi2021robustnet} and DIRL~\cite{Xu2022DIRLDR} reduce style-sensitive covariance components, whereas SHADE~\cite{Zhao2022StyleHallucinatedDC} and SAN-SAW~\cite{Peng2022SemanticAwareDG} introduce style hallucination and semantic-aware alignment. HGFormer~\cite{Ding2023HGFormerHG} and CMFormer~\cite{Bi2023LearningCM} leverage self-attention’s visual grouping to handle style variations, demonstrating the strengths of transformer-based models.

Recent studies reveal that excessive style regularization can weaken content discrimination. SPC~\cite{Huang2023StylePC} projects target styles into a source-style space, while DPCL~\cite{Yang2023GeneralizedSS} applies multi-level contrastive learning on projected source features. BlindNet~\cite{Ahn2024StyleBD} integrates covariance alignment and semantic consistency contrastive learning to reduce style sensitivity in the encoder while enhancing decoder robustness. In addition, some methods explore Vision Foundation Models (VFMs) to improve robustness and transferability in semantic segmentation~\cite{wei2024stronger, benigmim2024collaborating, zhao2025fishertune, zhang2025mamba}. Despite progress, DGSS methods focus on feature alignment and augmentation, leaving domain prior modeling underexplored. To bridge this gap, we introduce probabilistic diffusion modeling, enabling structured domain prior estimation and improved generalization to unseen domains.

\noindent\textbf{Diffusion Models.} Diffusion models (DMs)~\cite{Ho2020DenoisingDP} have recently gained significant attention for their outstanding performance in generative modeling. By parameterizing a Markov chain and optimizing the variational lower bound on the likelihood function, DMs effectively capture complex data distributions and produce high-quality samples\cite{Ho2020DenoisingDP, Kingma2021VariationalDM, zhang2023adding, zhao2023uni, zhang2024c3net, wang2024exploiting, jiang2024scedit}, outperforming conventional frameworks like GANs~\cite{Dhariwal2021DiffusionMB}. To reduce computational overhead, LDM~\cite{Rombach2021HighResolutionIS} and Würstchen~\cite{Pernias2023WuerstchenAE} conduct the diffusion process in latent space while retaining high-quality outputs. In addition to this, several studies have leveraged DMs to encode prior knowledge that facilitate downstream tasks. For instance, DiffIR~\cite{xia2023diffir} introduces a diffusion process applied to a compact image prior representation, enhancing both efficiency and stability under challenging conditions, while CDFormer~\cite{Liu2024CDFormerWhenDP} jointly reconstructs content and degraded representations to improve textural fidelity, and UniRestore~\cite{chen2025unirestore} further bridges perceptual and task-oriented image restoration by leveraging diffusion priors. These advancements highlight the role of diffusion-based methods in capturing complex data distribution, enhancing robustness and adaptability in generative modeling

\noindent\textbf{Diffusion Models for Domain Generalization.} Recent works explore diffusion models for domain generalization in semantic segmentation. DatasetDM~\cite{wu2023datasetdm} synthesizes diverse image-annotation pairs, while DGInStyle~\cite{jia2024dginstyle} employs semantic guidance to produce consistently paired image-label samples. Gong \textit{et al.}~\cite{gong2023prompting} adopt prompt randomization with pretrained diffusion features, and DIFF~\cite{ji2024diffusion} integrates sampling and fusion from pre-trained diffusion models to learn universal feature representations. These methods demonstrate the effectiveness of external diffusion-based models in enhancing domain generalization, but they typically incur substantial computational overhead. By contrast, our approach adopts an efficient probabilistic diffusion modeling framework to explicitly model latent domain priors, capturing diverse domain variations without imposing excessive resource demands. Moreover, this design seamlessly integrates with existing segmentation network, enabling robust out-of-distribution performance with minimal additional overhead.

%% file: Method.tex
\begin{figure*}
    \centering
    \includegraphics[width=1.0\linewidth]{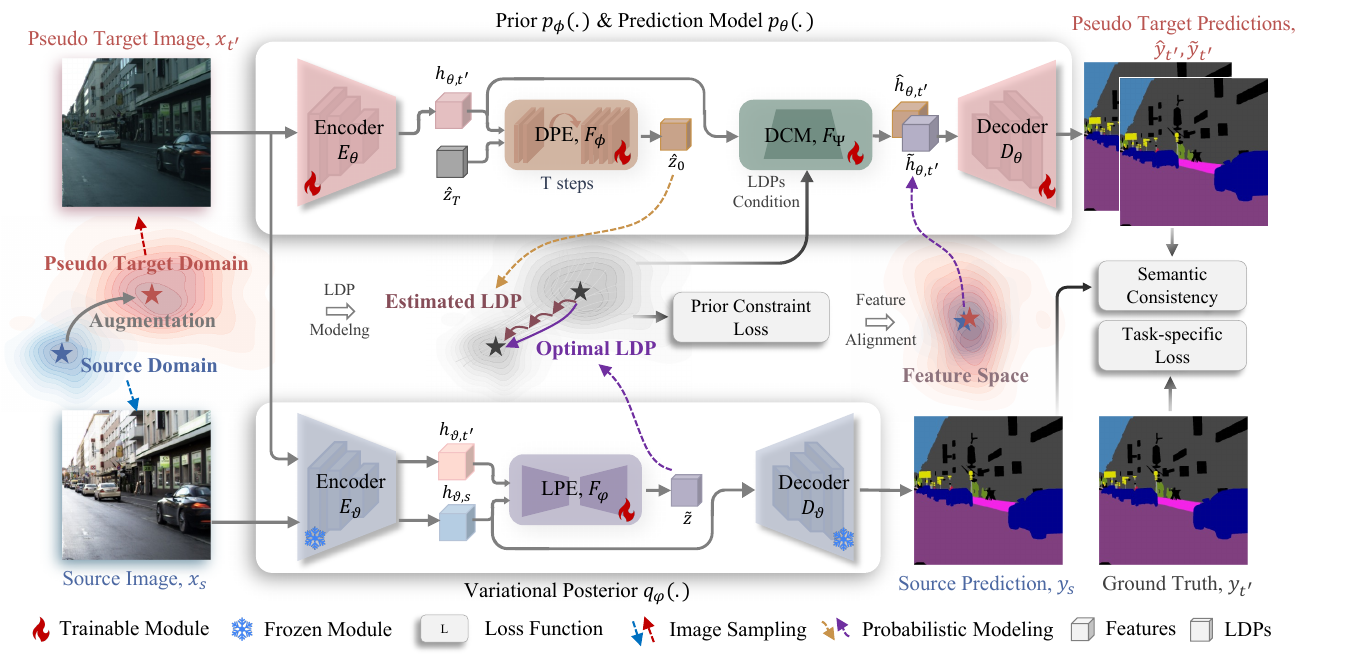}
    \caption{\textbf{Overall architecture of \method{}.} 
    \method{} augments a pre-trained segmentation network by introducing \LDP{} modeling to enhance its domain generalization. The \LPE{} learns the optimal \LDP{} by modeling cross-domain relationships between source and pseudo-target domains, while \DPE{} is employed to estimate \LDP{} using only target inputs. Finally, the \DCM{} enhances segmentation network with \LDP{} guidance, refining feature alignment and improving domain generalization.}
    \label{fig:overview}
\end{figure*}

\section{Method}
We first introduce the Latent Domain Prior (\LDP{}) Modeling (Sec.~\ref{sec: Latent Domain Prior Modeling}), where we formulate domain generalized semantic segmentation (DGSS) as a probabilistic learning problem and derive the variational inference framework for \LDP{} estimation.
Based on this, we develop the Probabilistic Diffusion Alignment Framework (Sec.~\ref{sec: Probabilistic Diffusion Alignment Framework}), which leverages \LDP{}s to model domain shifts and guide the existing segmentation model in adapting feature representations. 
Finally, we describe Model Optimization (Sec.~\ref{sec: Model Optimization}), detailing the training objectives and loss functions designed to enhance robust domain generalization. 

\subsection{Latent Domain Prior Modeling}
\label{sec: Latent Domain Prior Modeling} 
We propose a probabilistic modeling framework that explicitly integrates \LDP{} to model domain shifts in the DGSS task. 
By formulating \LDP{} within a variational inference framework, our method enables principled estimation of domain prior that serves as conditioning variables, thereby guiding segmentation models to generalize more effectively on unseen target domains.
\\\noindent\textbf{Probabilistic Problem Formulation.} To generalize across domain shifts, we cast DGSS as a probabilistic learning problem. 
We introduce a Latent Domain Prior (\LDP{}) as a latent variable \(z\) to capture domain-specific variations that are not directly observable. 
Given the target domain distribution \(p(x_t, y_t)\), where \(y_t\) denotes the ground truth, we infer \(z\) solely from the available target-domain image \(x_t\). Consequently, the prediction function of \method{} is formulated as:
\begin{equation}
    \label{eqn:probabilistic_guidance}
    p_{\theta,\phi}(y_t|x_t) = \int p_\theta(y_t|x_t,z)p_\phi(z|x_t)dz,
\end{equation}
where \(\theta\) and \(\phi\) represent the parameters of the segmentation model and the network designed to estimate \(z\) (i.e., \LDP{}), respectively. 
By incorporating \LDP{}, \method{} regularizes feature adaptation, consequently mitigating domain shifts and enhancing model robustness.

\noindent\textbf{Variational Inference for \LDP{} Estimation.} To facilitate tractable inference of \LDP{}, we introduce a variational posterior $q_\varphi(z|x_t, x_s)$ to approximate the true posterior distribution $p(z|x_t,x_s)$, where $\varphi$ represents the parameters of the inference network and $x_s$ indicates the source domain counterpart of target domain image $x_t$. 
The pair of $x_s$ and $x_t$ preserves semantic consistency while exhibiting domain-specific variations. 
By leveraging variational inference, we construct a principled optimization framework that estimates \LDP{} to enhance domain alignment.
The Evidence Lower Bound (ELBO) of the predictive function is formulated as:
\begin{equation}
    \begin{aligned}
        \log p_{\theta, \phi}(y_t | x_t) &\geq \mathbb{E}_{q_\varphi(z | x_t, x_s)} \left[ \log p_\theta(y_t | x_t, z) \right] \\&- \mathbb{KL} \left[ q_\varphi(z | x_t, x_s) || p_\phi(z | x_t) \right].
    \end{aligned}   
\end{equation}
The variational posterior $q_\varphi(z|x_t, x_s)$ explicitly encodes cross-domain feature correlations from source-target pairs, enabling precise modeling of latent domain variation critical for effective generalization. 
The KL divergence term encourages the learned posterior to approximate the prior, ensuring structured regularization. A detailed derivation of ELBO is provided in the supplementary material.

\subsection{Probabilistic Diffusion Alignment Framework}
\label{sec: Probabilistic Diffusion Alignment Framework}

\vspace{1mm} 
\noindent\textbf{Overview of the Framework.} We introduce the Probabilistic Diffusion Alignment Framework (\method{}), which instantiates LDP modeling to obtain domain shift, thereby enhancing generalization performance of segmentation models. 
To achieve this, \method{} integrates into a pre-trained segmentation model and utilizes paired images generated through data augmentation, effectively simulating domain shifts. This process preserves semantic consistency while injecting domain-specific variations, thus facilitating the principled ELBO approximation. The optimization function of \method{} is described as follows:
\begin{equation}
    \label{eqn:optim}
\begin{aligned}
    \mathcal{L}_\text{PDAF} = &- \mathbb{E}_{q_\varphi(z' | x_{t'}, x_s)} \left[ \log p_\theta(y_{t'} | x_{t'}, z') \right]\\ &+ \mathbb{KL} \left( q_\varphi(z' | x_{t'}, x_s) \, || \, p_\phi(z' | x_t) \right),
\end{aligned}
\end{equation}
where $x_{t'}$ represents pseudo-target images generated from source images $x_s$ via data augmentation, while $y_{t'}$ is the corresponding ground truth. $z'$ denotes the \LDP{}, inferred through cross-domain relationships between source and pseudo-target domain samples. 
The prediction model $p_\theta(.)$, variational posterior $q_\varphi(.)$ and prior $p_\phi(.)$ jointly optimize segmentation accuracy while modeling domain shifts by maximizing likelihood and minimizing KL divergence, respectively. 

Corresponding to the optimization function in Eq.~\ref{eqn:optim}, \method{} introduces three components: (i) a Latent Prior Extractor (\LPE{}), which serves as $q_\varphi(.)$ and infers the optimal \LDP{} from source and pseudo-target pairs; (ii) a Domain Compensation Module (\DCM{}), integrated into a pretrained segmentation network to implement the prediction model $p_\theta(.)$, which leverages the extracted \LDP{} to refine feature representations, thereby mitigating domain shifts and improving segmentation accuracy; and (iii) a Diffusion Prior Estimator (\DPE{}), implementing $p_\phi(.)$ through probabilistic diffusion modeling to estimate \LDP{} without paired source and target data.
Notably, \method{} is compatible with any existing segmentation model, allowing frozen pre-trained encoders $E_\vartheta$ and decoders $D_\vartheta$ to discover \LDP{}, and then fine-tuning the target network  $E_\theta$ and $D_\theta$ for DGSS. The overall architecture of \method{} is illustrated in ~\figref{fig:overview}. 

\begin{figure*}
    \centering
    \begin{subfigure}{0.33\linewidth}
        \includegraphics[width=1.0\linewidth]{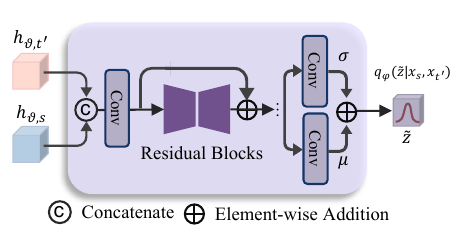}
        \caption{Details of \LPE{}}
        \label{fig:LPE}
    \end{subfigure}
    \hfill
    \begin{subfigure}{0.33\linewidth}
        \includegraphics[width=1.0\linewidth]{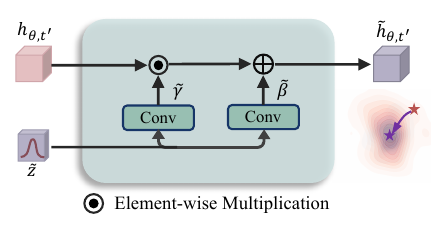}
        \caption{Details of \DCM{}}
        \label{fig:DCM}
    \end{subfigure}
    \begin{subfigure}{0.33\linewidth}
        \includegraphics[width=1.0\linewidth]{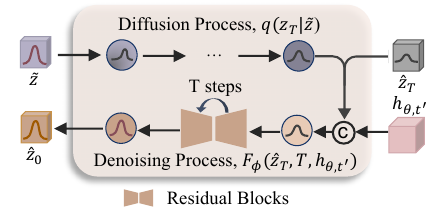}
        \caption{Details of \DPE{}}
        \label{fig:DPE}
    \end{subfigure}
    \caption{\textbf{Schematic diagrams of (a) the Latent Prior Extractor, (b) the Domain Compensation Module and (c) the Diffusion Prior Estimator.}}
    \label{fig:modules}
\end{figure*}

\vspace{1mm} 
\noindent\textbf{Latent Prior Extractor.} The \LPE{} is designed to estimate the optimal \LDP{} by supervising cross-domain feature relationships, providing effective guidance for feature alignment. 
As illustrated in~\figref{fig:LPE}, the \LPE{} first concatenates source features \( h_{\vartheta, s} \) and pseudo-target features \( h_{\vartheta, t'} \), then utilizes residual blocks to model cross-domain feature relationships.
The extracted features are then processed through two projection layers to derive the probabilistic modeling parameters, namely the mean (\(\mu\)) and variance (\(\sigma\)).
To ensure stable and effective training, we constrain the optimal \LDP{} \( \tilde{z} \) to follow a standard normal distribution~\cite{Kingma2013AutoEncodingVB}, enhancing latent space regularization and enabling more robust prior estimation.
The procedure can be formulated as: 
\begin{equation}
    \label{eqn:LPE}
\begin{aligned}
    h_{\vartheta, s} &= E_\vartheta(x_s), h_{\vartheta, t'} = E_\vartheta(x_{t'}),\\
    \mu, \sigma &= F_\varphi(h_{\vartheta, s}, h_{\vartheta, t'}),\\
    \tilde{z}&=\mu + \epsilon\sigma,\  \epsilon\sim\mathcal{N}(0,I),
\end{aligned}
\end{equation}
where $F_\varphi$ indicates \LPE{} parameterized by $\varphi$. 
The dimension of \LDP{} is $\tilde{z}\in \mathbb{R}^{c'\times h \times w}$, where $c'$ denotes a predefined number of the channel of \LDP{} and $h, w$ correspond to the spatial dimensions of the feature maps. 

\vspace{1mm} 
\noindent\textbf{Domain Compensation Module.} 
The \DCM{} enhances segmentation models for domain generalization. 
Drawing inspiration from Spatial Feature Transformation (SFT)~\cite{wang2018recovering}, \DCM{} leverages \LDP{} as a compensation mechanism, providing domain-aware modulation to improve feature alignment. 
As illustrated in~\figref{fig:DCM}, a straightforward approach involves projecting \LDP{} onto affine transformation parameters using dedicated projection layers (\(F_{\Psi}\)), yielding scale (\(\tilde{\gamma}\)) and shift (\(\tilde{\beta}\)) parameters. 
These parameters are then used to rescale and shift the feature representations, effectively normalizing domain discrepancies across spatial regions. 
The refined features \(\tilde{h}_{\theta,t'}\) are subsequently processed by the segmentation head \(D_\theta\) to generate accurate segmentation maps \(\tilde{y}_{t'}\).
The process can be formulated as:
\begin{equation}
    \label{eqn:DCM}
    \begin{aligned}
    \tilde{\gamma}, \tilde{\beta} &= F_{\Psi}(\tilde{z}), \\
    \tilde{h}_{\theta,t'}&=\tilde{\gamma} \odot h_{\theta, t'} \oplus \tilde{\beta}, \\
    \tilde{y}_{t'}&=D_\theta(\tilde{h}_{\vartheta,t'}),
    \end{aligned}
\end{equation}
where $\odot$ and $\oplus$ denote element-wise multiplication and element-wise addition, respectively. 
The dimensions of the affine transformation parameters correspond to the feature map, with $\tilde{\gamma}, \tilde{\beta} \in \mathbb{R}^{c\times h \times w}$. 

\vspace{1mm} 
\noindent\textbf{Diffusion Prior Estimator.} 
Diffusion Models (DMs) have demonstrated their effectiveness in sampling from complex target domains, significantly improving various computer vision tasks as evidenced in previous studies~\cite{Rombach2021HighResolutionIS, Pernias2023WuerstchenAE, Liu2024CDFormerWhenDP}. 
In \DPE{}, we leverage probabilistic diffusion modeling to estimate the \LDP{}, enabling arbitrary target domain and accurate prior estimation for domain alignment. 
As illustrated in~\figref{fig:DPE}, we first apply the forward diffusion process to an optimal \LDP{} $\tilde{z}$, iteratively adding Gaussian noise to obtain an intermediate latent state $z_T$: 
\begin{equation}
    \label{eqn:forward}
q(z_T|\tilde{z}) = \mathcal{N}(z_T;\sqrt{\bar{\alpha}_T}\tilde{z}, (1-\bar{\alpha}_T)I),
\end{equation}
where $ \alpha_T = 1 - \beta_T $ and $ \bar{\alpha}_T = \prod_{i=1}^T \alpha_i $. For the reverse diffusion process, the \DPE{} initializes from $\hat{z}_T$ and perform T-step denoising process to estimate the \LDP{} $\hat{z}_0$, which is formulated as: 
\begin{equation}
    \label{eqn:denoising}
    \hat{z}_{T-1}=\frac{1}{\sqrt{\alpha_T}}(\hat{z}_T-\epsilon\frac{1-\alpha_T}{\sqrt{1-\bar{\alpha}_T}}).
\end{equation}

Specifically, to estimate \LDP{} for target images, we condition the \DPE{} on the extracted target features \(h_{\theta,t'}\). 
\DPE{} (\(F_\phi\)) then performs a denoising operation at timestep \(T\), represented as \(F_\phi(\hat{z}_T, T, h_{\theta,t'})\). 
To accommodate the reduced channel dimension of \LDP{}, we employ accelerated diffusion-based optimization~\cite{song2020denoising} for efficient and rapid inference. 
Moreover, this design allows for the joint training of \DPE{} and the segmentation head \(D_\theta\), thereby optimizing \LDP{} directly for the segmentation task.
This process is defined as follows:
\begin{equation}
    \label{eqn:DPE}
    \begin{aligned}
    \hat{z}_0 &= F_{\phi}(\hat{z}_T, T, h_{\theta, t'}), \\
    \hat{\gamma}, \hat{\beta} &= F_{\Psi}(\hat{z}_0), \\
    \hat{h}_{\theta,t'}&= \hat{\gamma}\odot h_{\theta, t'} \oplus \hat{\beta}, \\
    \hat{y}_{t'}&=D_\theta(\hat{h}_{\vartheta,t'}),
    \end{aligned}
\end{equation}
where $\hat{y}_{t'}$ represents the predicted segmentation maps conditions on the estimated \LDP{} $\hat{z}_0$. 
The dimensions of the affine transformation parameters correspond to the feature map, with $\hat{\gamma}, \hat{\beta} \in \mathbb{R}^{c\times h \times w}$. 

\vspace{1mm} 
\noindent\textbf{Inference.} 
During inference, only target domain images \(x_t\) are available, so the variational posterior is omitted. 
Instead, we extract target features \(h_{\theta,t}\) using the segmentation backbone \(E_\theta(\cdot)\) as the conditioning input and sample a Gaussian noise vector \(\hat{z}_T\). 
The \DPE{} then iteratively refines this noise over \(T\) denoising steps to produce the estimated \LDP{} \(\hat{z}_0\). 
Finally, the \DCM{} leverages the estimated \LDP{} to refine the feature representations, which are used for robust segmentation prediction.

\subsection{Model Optimization} 
\label{sec: Model Optimization}
We train our framework by optimizing the likelihood estimation and enforcing posterior regularization to enhance domain generalization, as defined in~\cref{eqn:optim}. 
To achieve maximum likelihood estimation, we employ task-specific loss functions $\mathcal{L}_\text{task}$ tailored to different segmentation models. 
Specifically, we use weighted cross-entropy loss for DeepLabV3+~\cite{Chen2017RethinkingAC} and focal loss for Mask2Former~\cite{Cheng2021MaskedattentionMT}. 
To further enhance feature consistency, we introduce a semantic consistency loss $\mathcal{L}_\text{sc}$ that measures the discrepancy between the predictions of the source and pseudo-target images. 
This loss enforces semantic alignment in feature space and accelerates model convergence:
\begin{equation}
    \mathcal{L}_\text{sc} = \|f_\vartheta(x_s)-\tilde{y}_{t'}\|_2 + \|f_\vartheta(x_s)-\hat{y}_{t'}\|_2,
\end{equation}
where $f_\vartheta$ denotes the frozen segmentation network $E_\vartheta(.)$ and $D_\vartheta(.)$, $\tilde{y}_{t'}$ and $\hat{y}_{t'}$ represent the predictions from \cref{eqn:DCM} and \cref{eqn:DPE}, respectively.
To minimize the KL divergence and enforce posterior regularization, we adopt a prior constraint loss $\mathcal{L}_\text{prior}$ to align the outputs of \LPE{} and \DPE{}, thereby implicitly reducing the distributional discrepancy between the posterior and the prior: 
\begin{equation}
    \mathcal{L}_\text{prior} = \|\hat{z}_0 - \tilde{z}\|_2.
\end{equation}
where $\hat{z}_0$ and $\tilde{z}$ indicate the optimal \LDP{} and estimated \LDP{} from \cref{eqn:LPE} and \cref{eqn:DPE}, respectively. 
The overall loss is:
\begin{equation}
    \label{eqn:total_loss}
 \mathcal{L}_\text{total} = \lambda_\text{task}\mathcal{L}_\text{task} + \lambda_\text{sc} \mathcal{L}_\text{sc} + \lambda_\text{prior} \mathcal{L}_\text{prior},
\end{equation}
where $\lambda_\text{task}, \lambda_\text{sc}$, and $\lambda_\text{prior}$ represents the scaling factors.

%% file: Implementation_Detail.tex
\begin{figure*}[t!]
    \centering
    \setlength{\tabcolsep}{1pt}
    \footnotesize
      \begin{tabular}{cccccccc}
          \includegraphics[width=.12\textwidth, height=1.8cm]{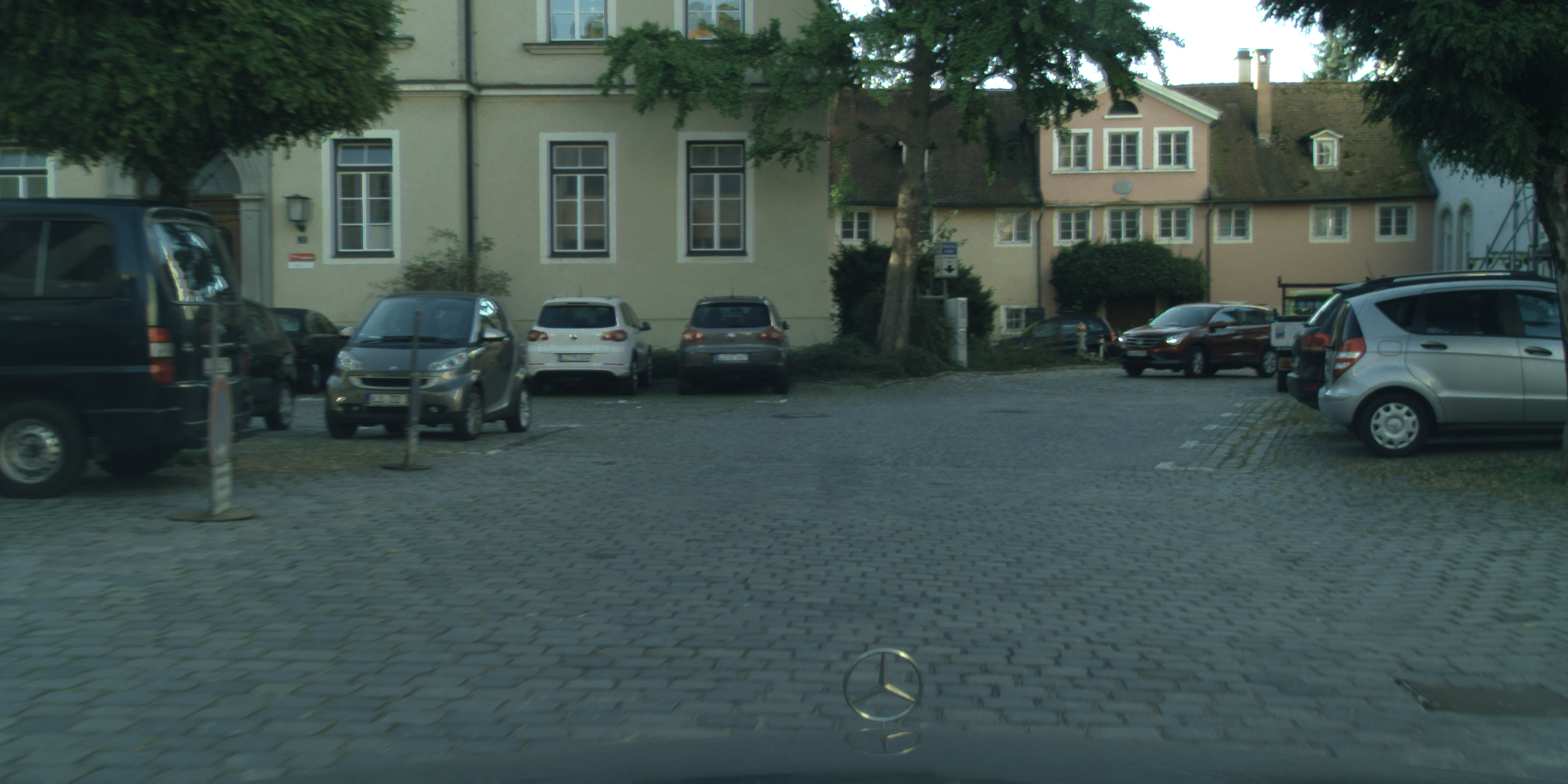} &
          \includegraphics[width=.12\textwidth, height=1.8cm]{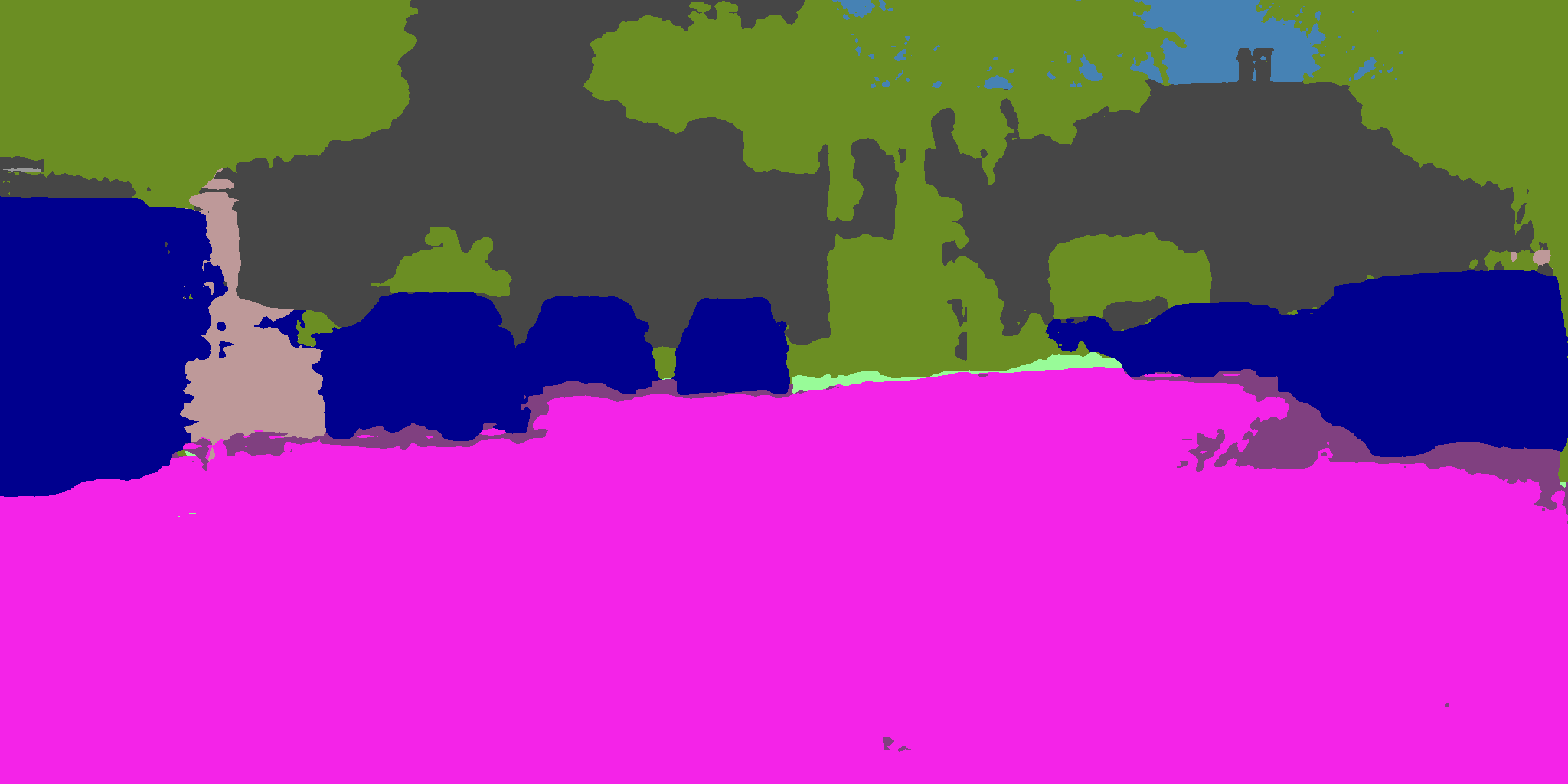} &
          \includegraphics[width=.12\textwidth, height=1.8cm]{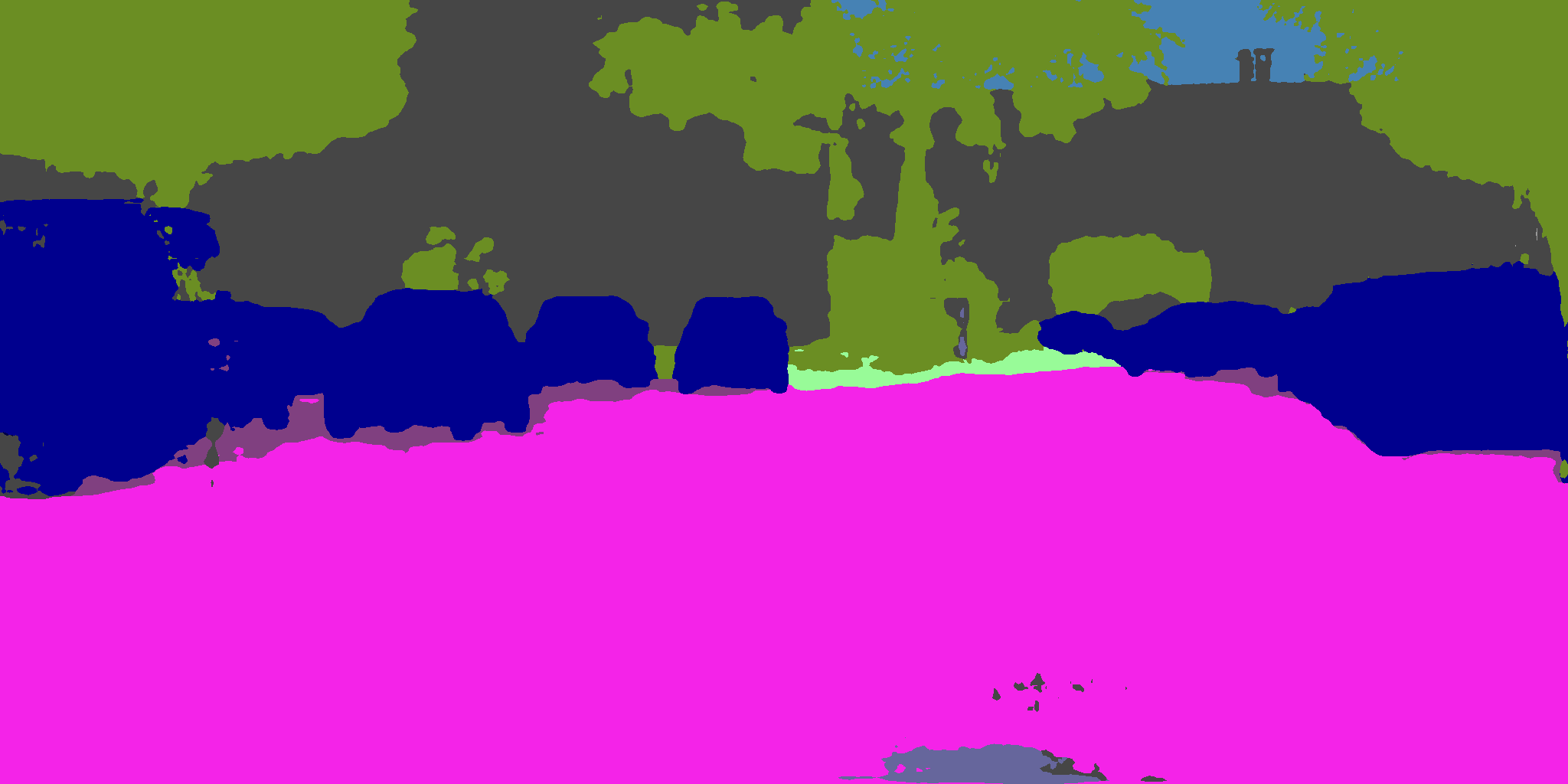} &
          \includegraphics[width=.12\textwidth, height=1.8cm]{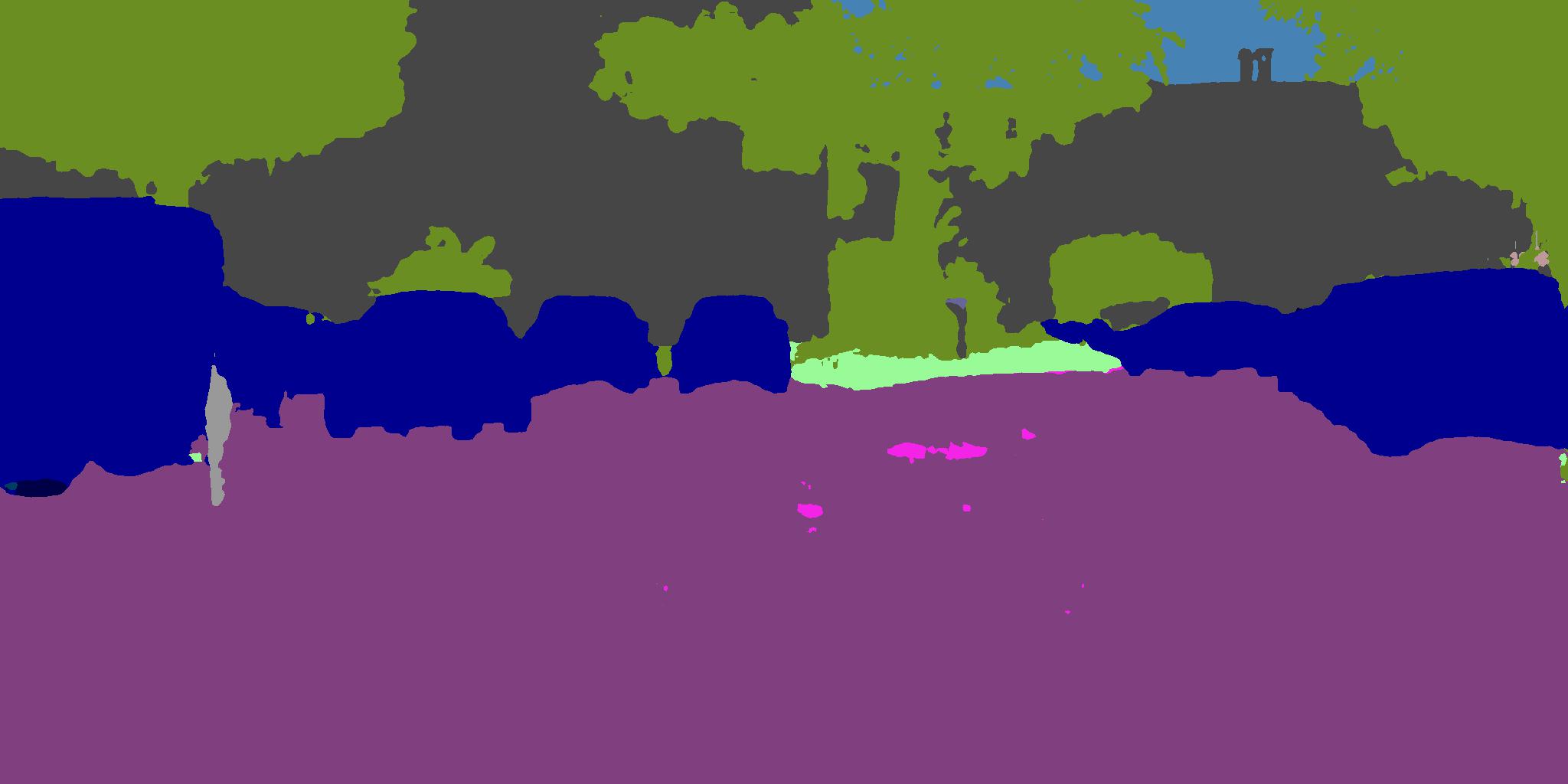} &
          \includegraphics[width=.12\textwidth, height=1.8cm]{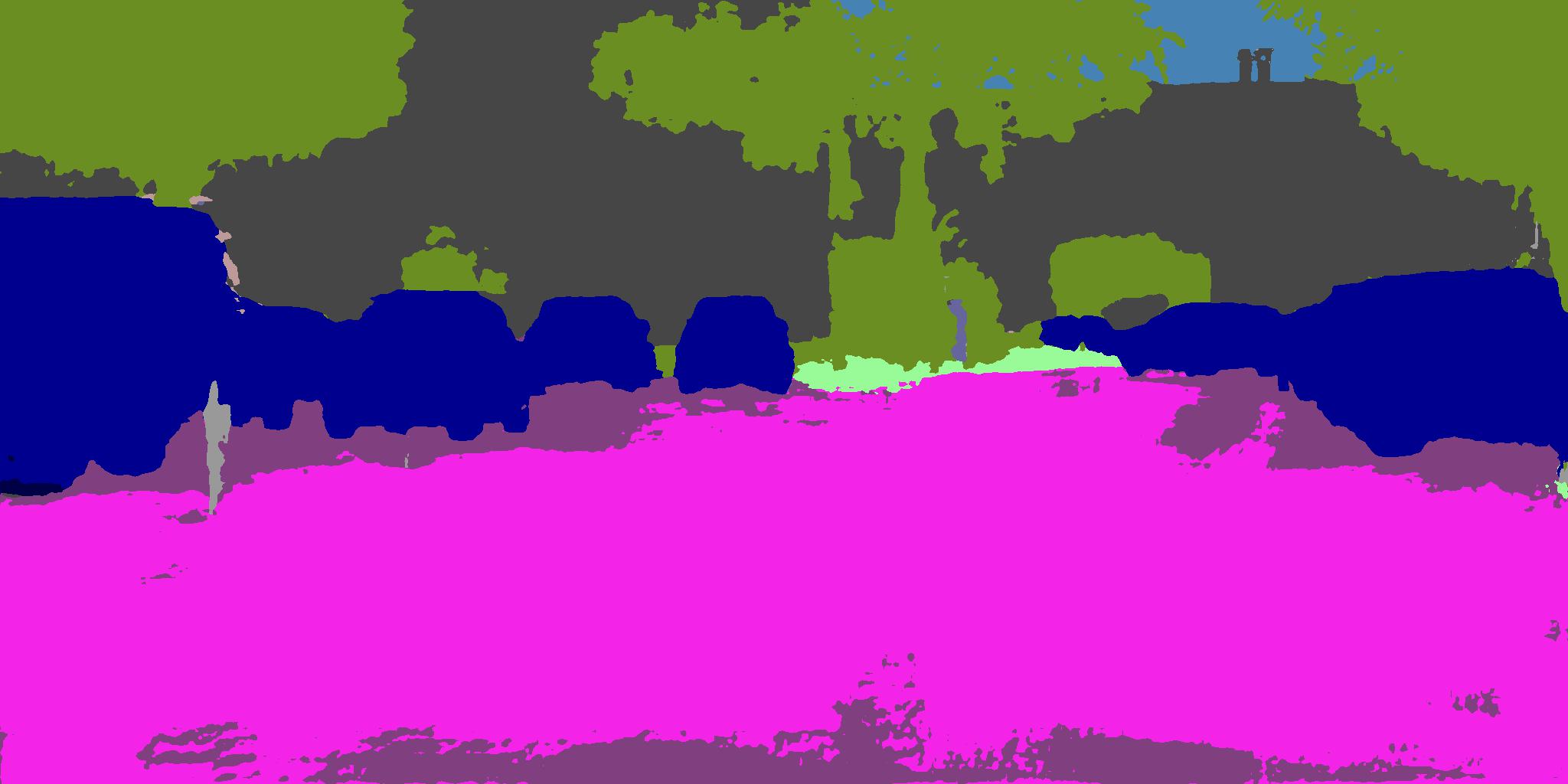} &
          \includegraphics[width=.12\textwidth, height=1.8cm]{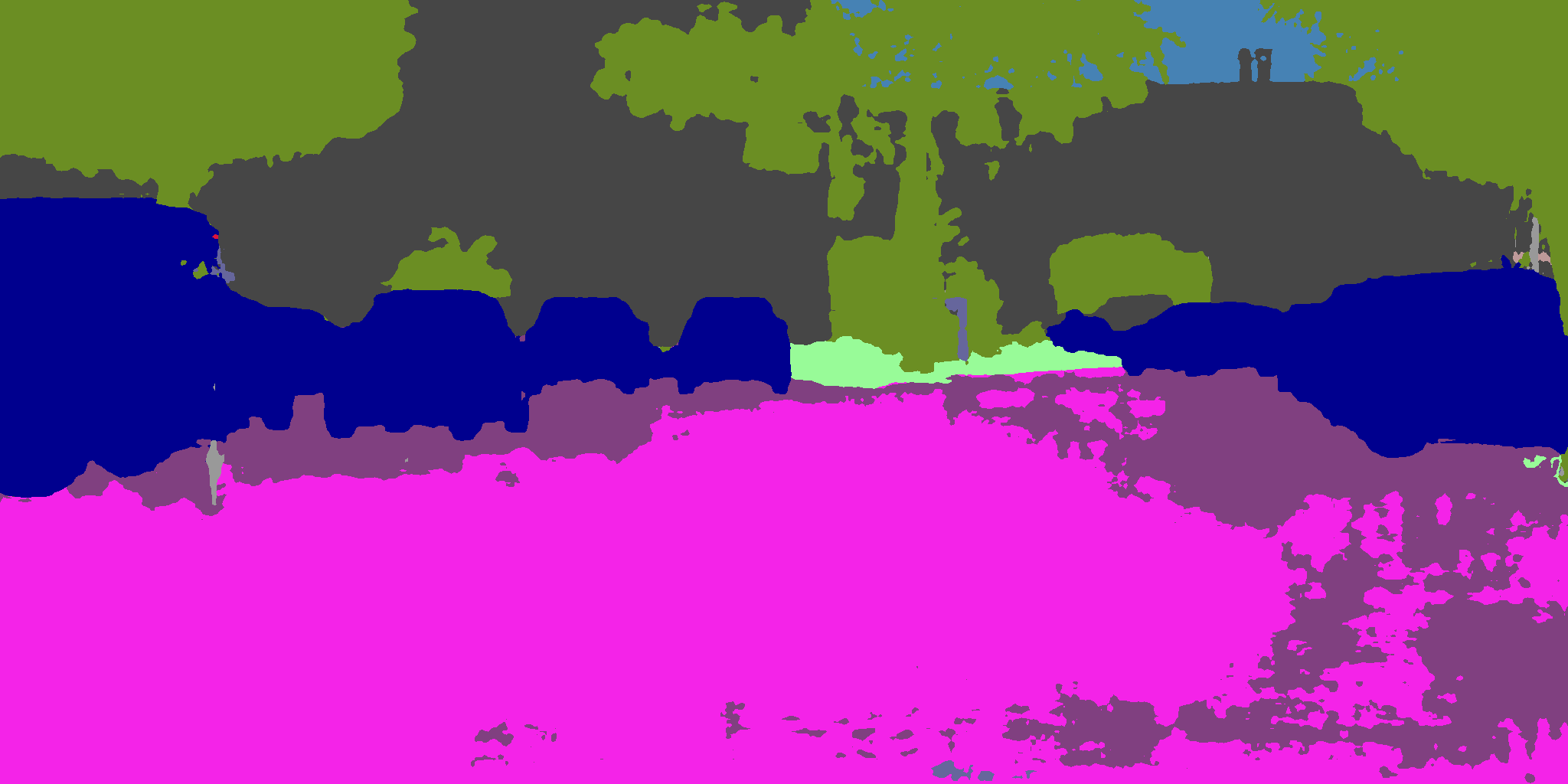} &
          \includegraphics[width=.12\textwidth, height=1.8cm]{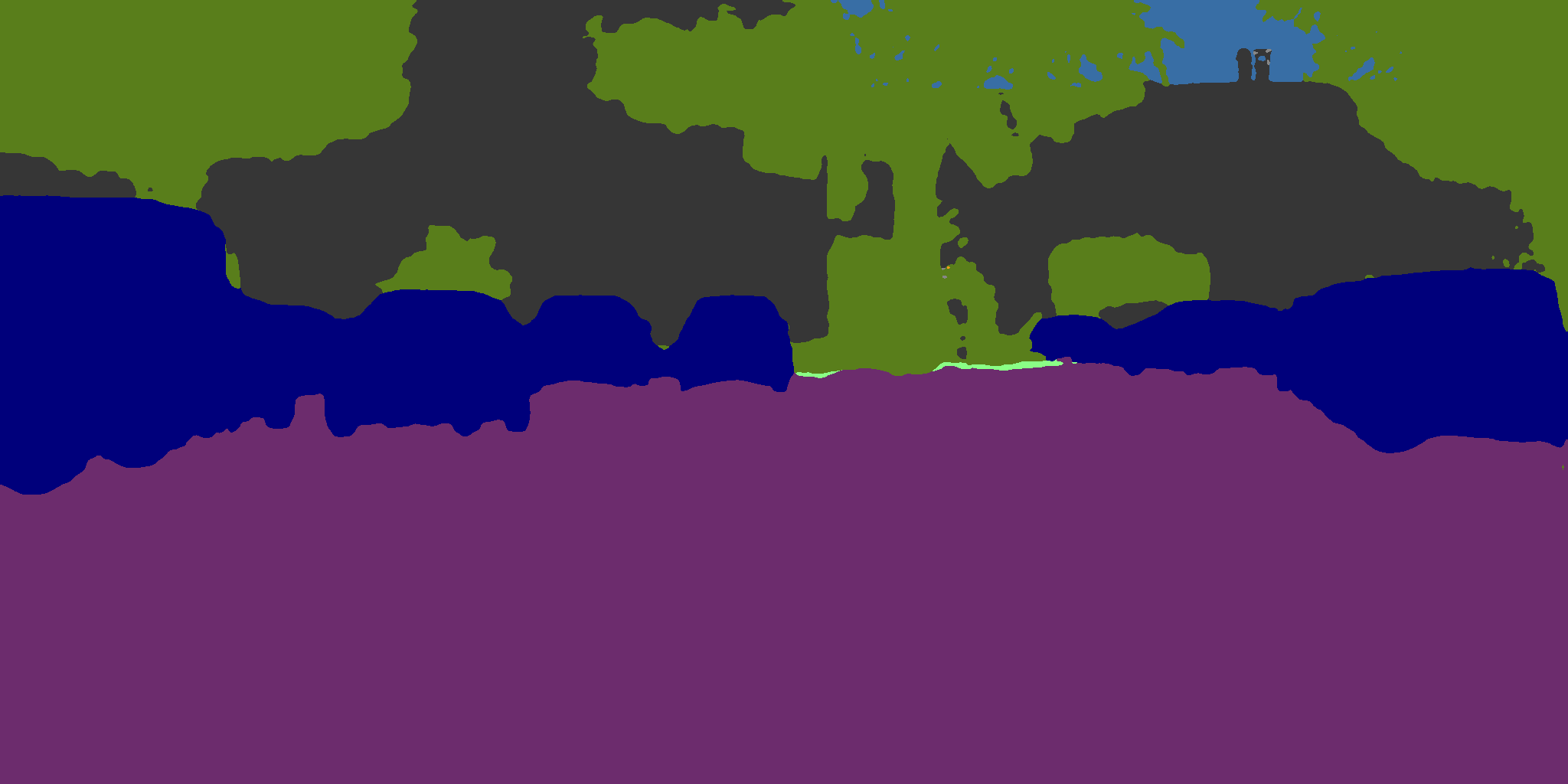} &
          \includegraphics[width=.12\textwidth, height=1.8cm]{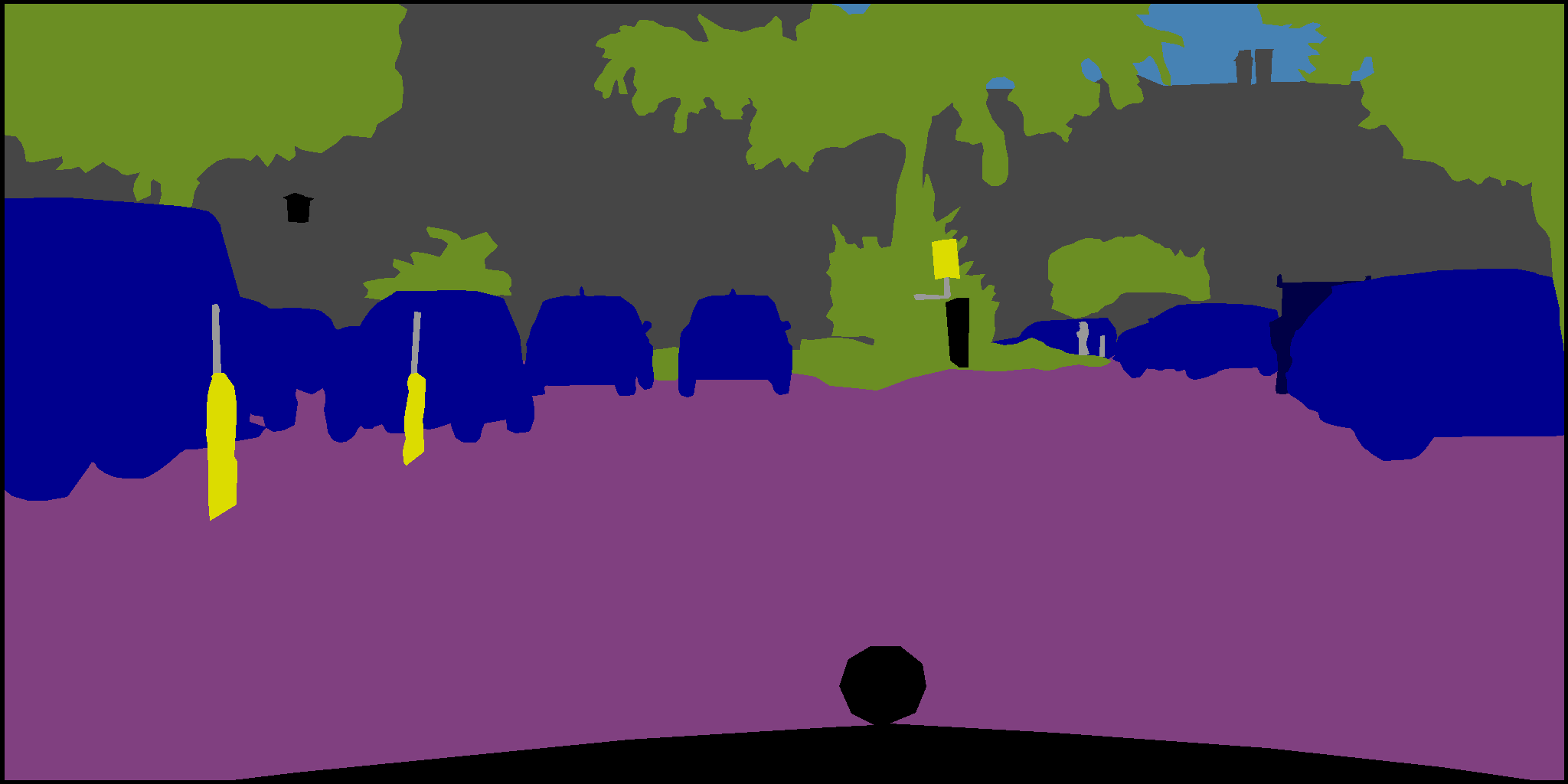} \\
          \includegraphics[width=.12\textwidth, height=1.8cm]{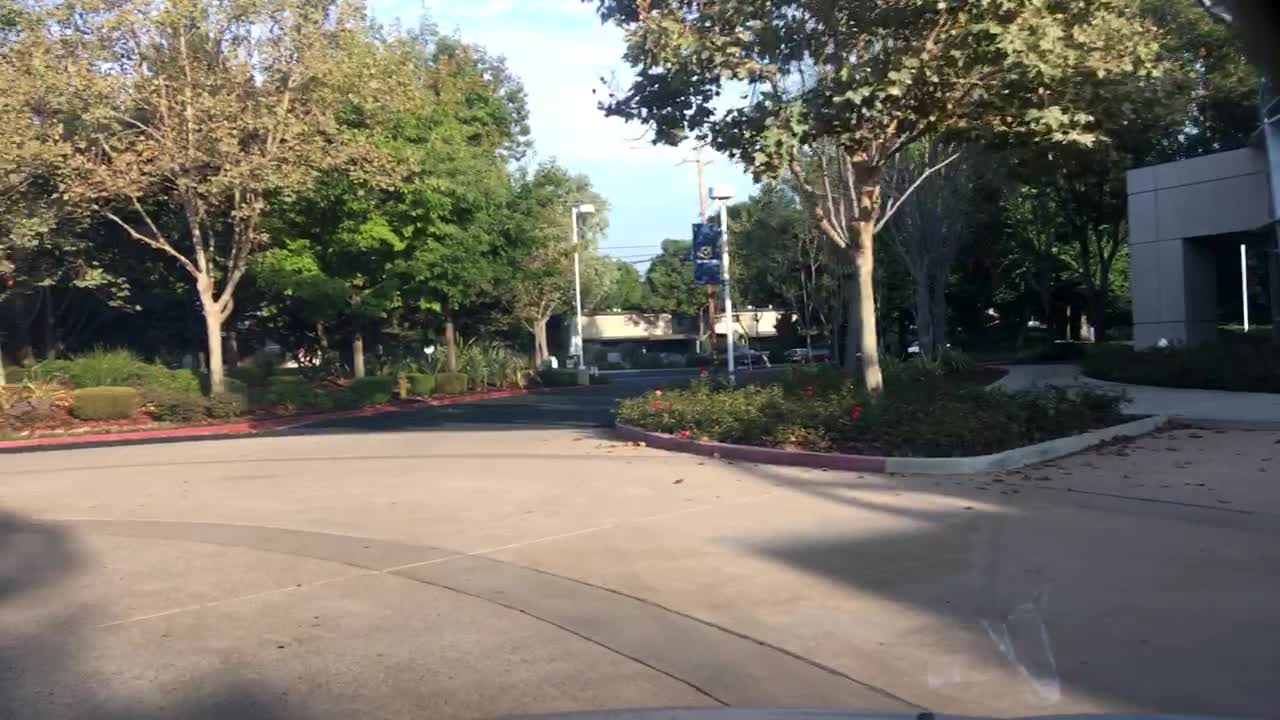} &
          \includegraphics[width=.12\textwidth, height=1.8cm]{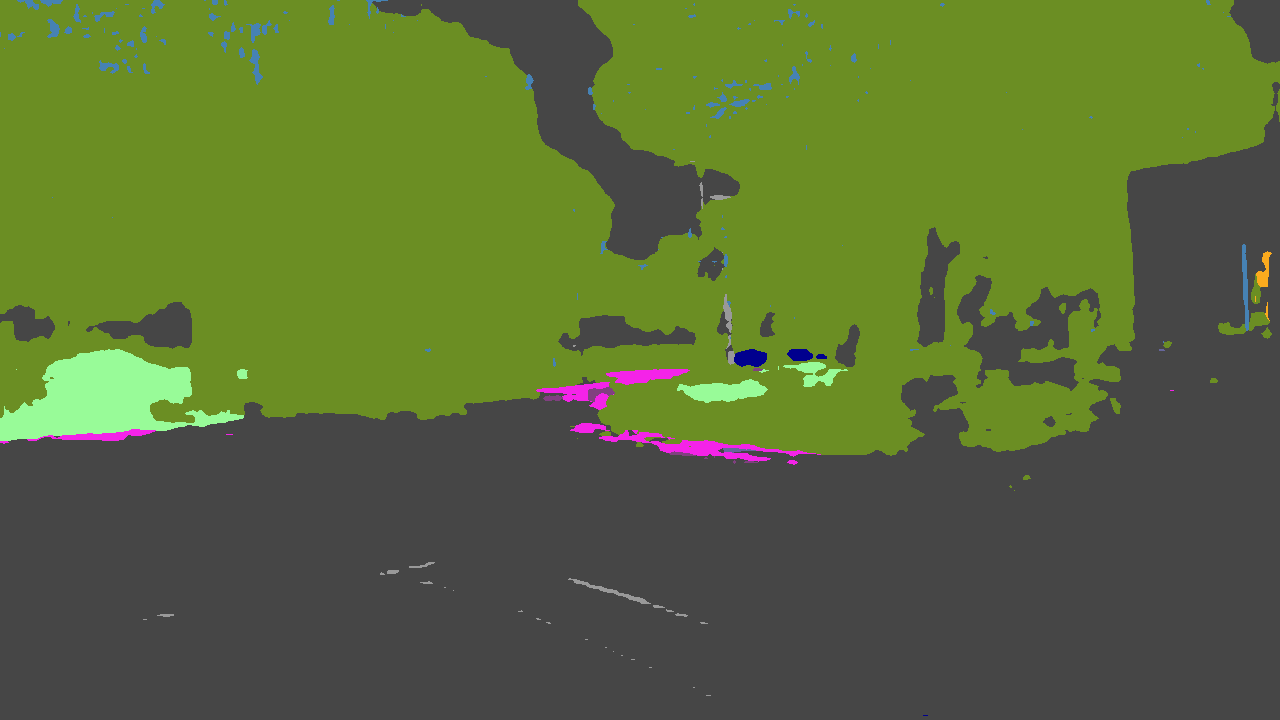} &
          \includegraphics[width=.12\textwidth, height=1.8cm]{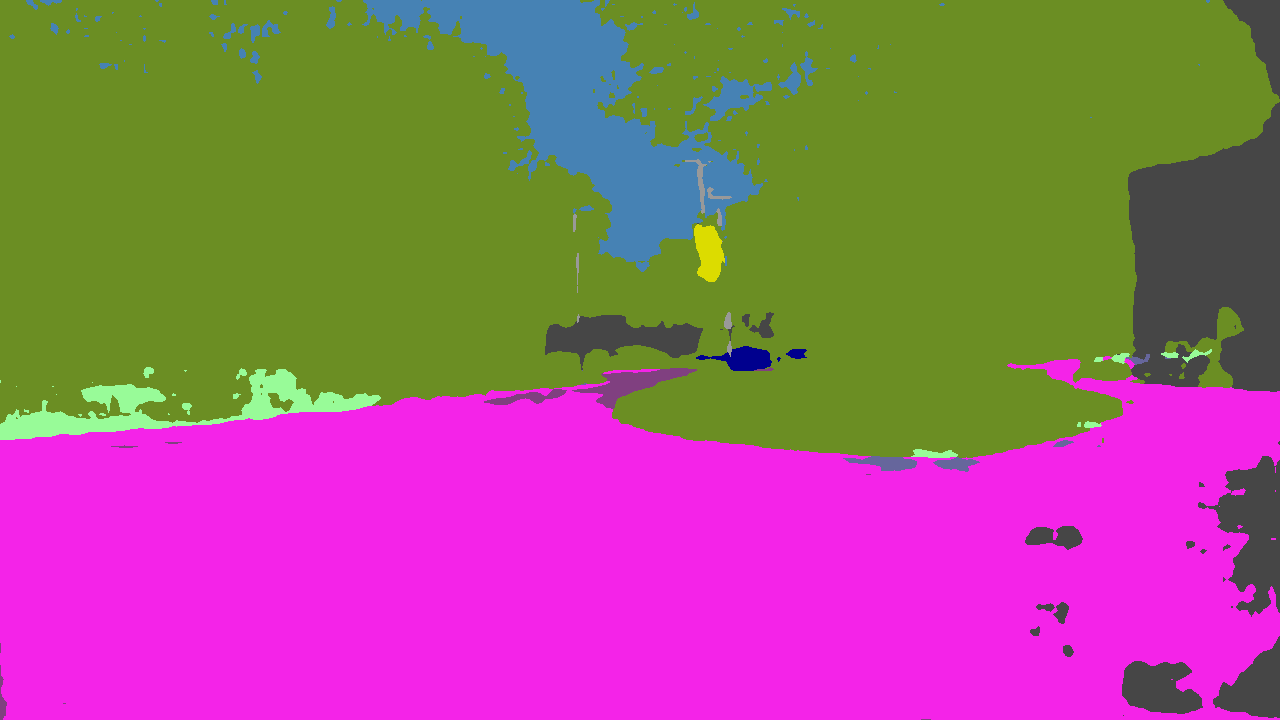} &
          \includegraphics[width=.12\textwidth, height=1.8cm]{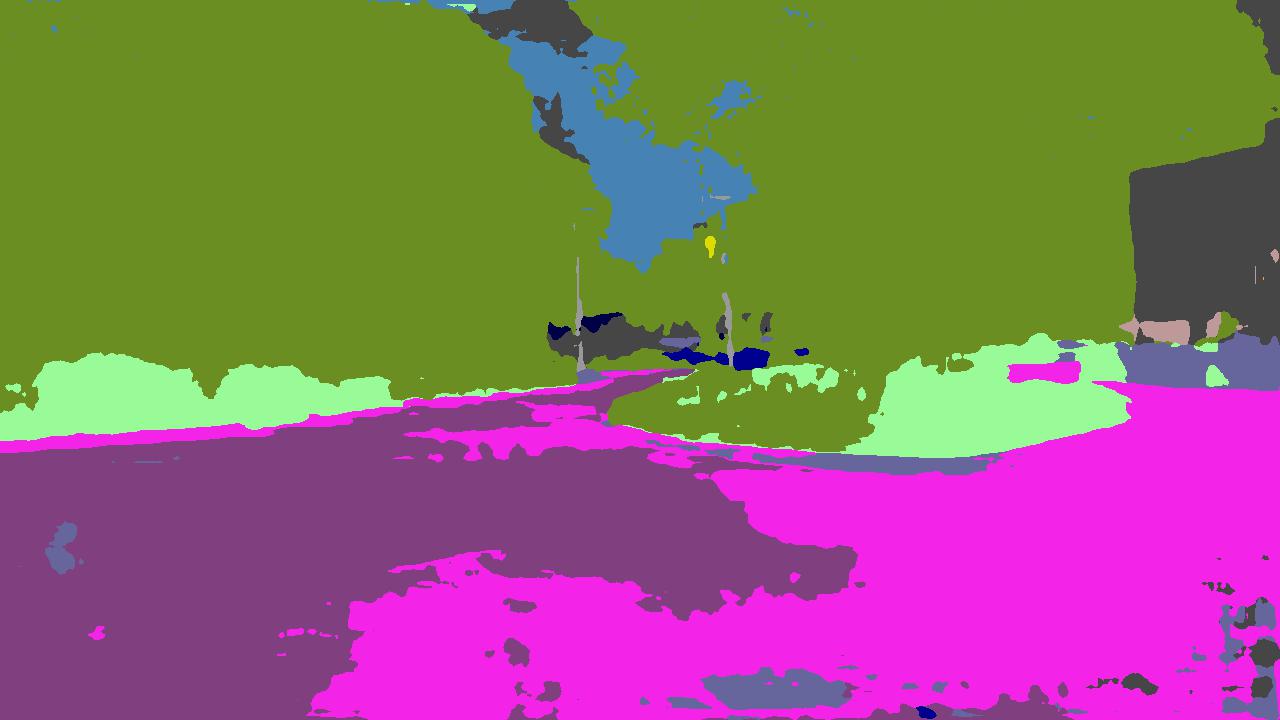} &
          \includegraphics[width=.12\textwidth, height=1.8cm]{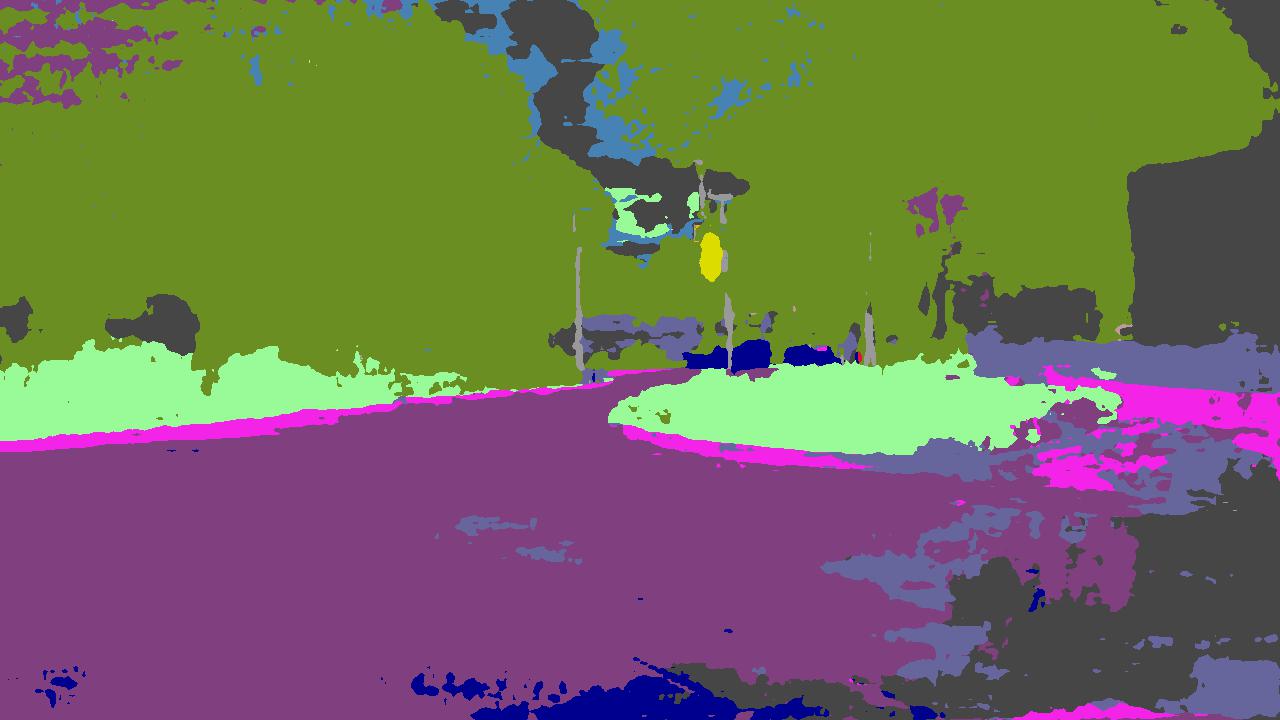} &
          \includegraphics[width=.12\textwidth, height=1.8cm]{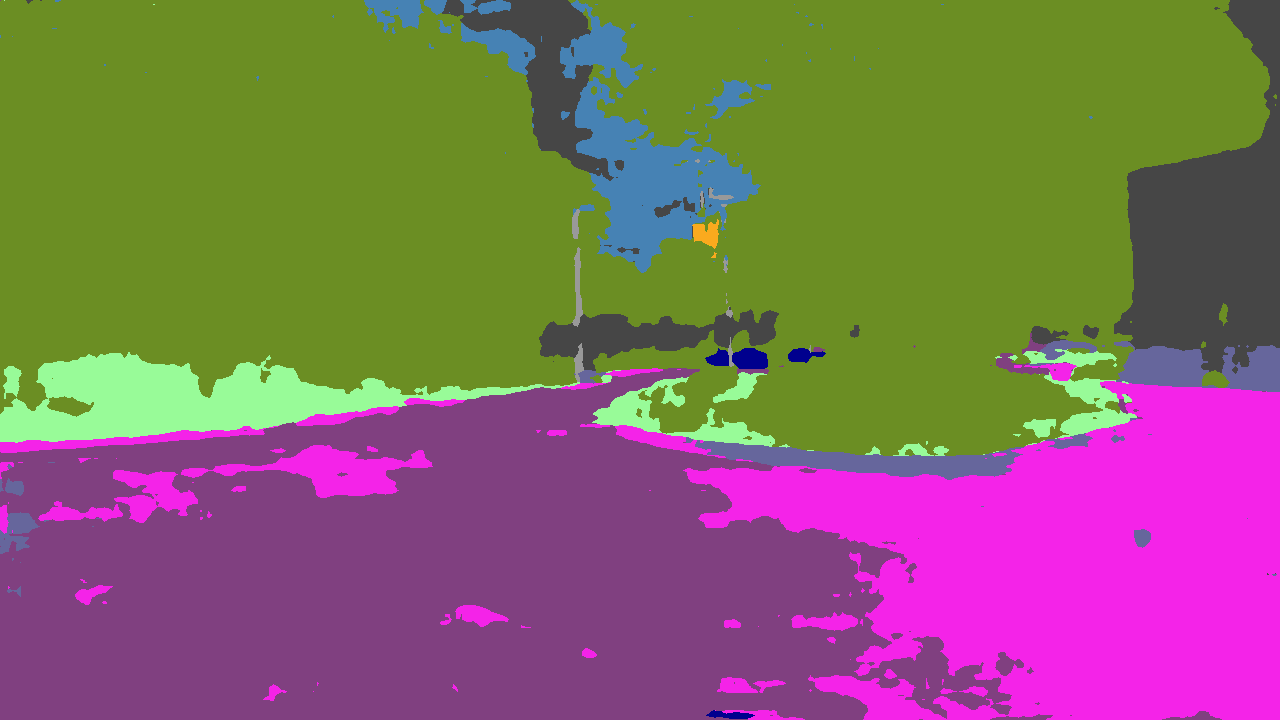} &
          \includegraphics[width=.12\textwidth, height=1.8cm]{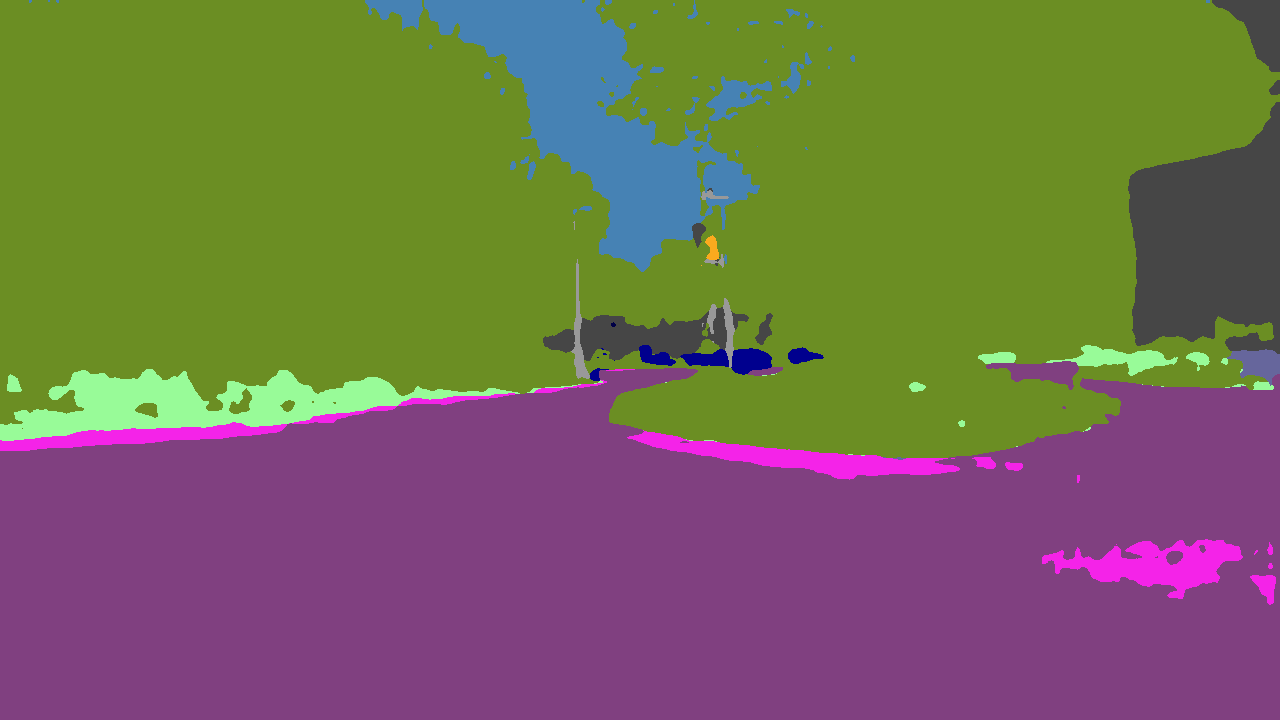} &
          \includegraphics[width=.12\textwidth, height=1.8cm]{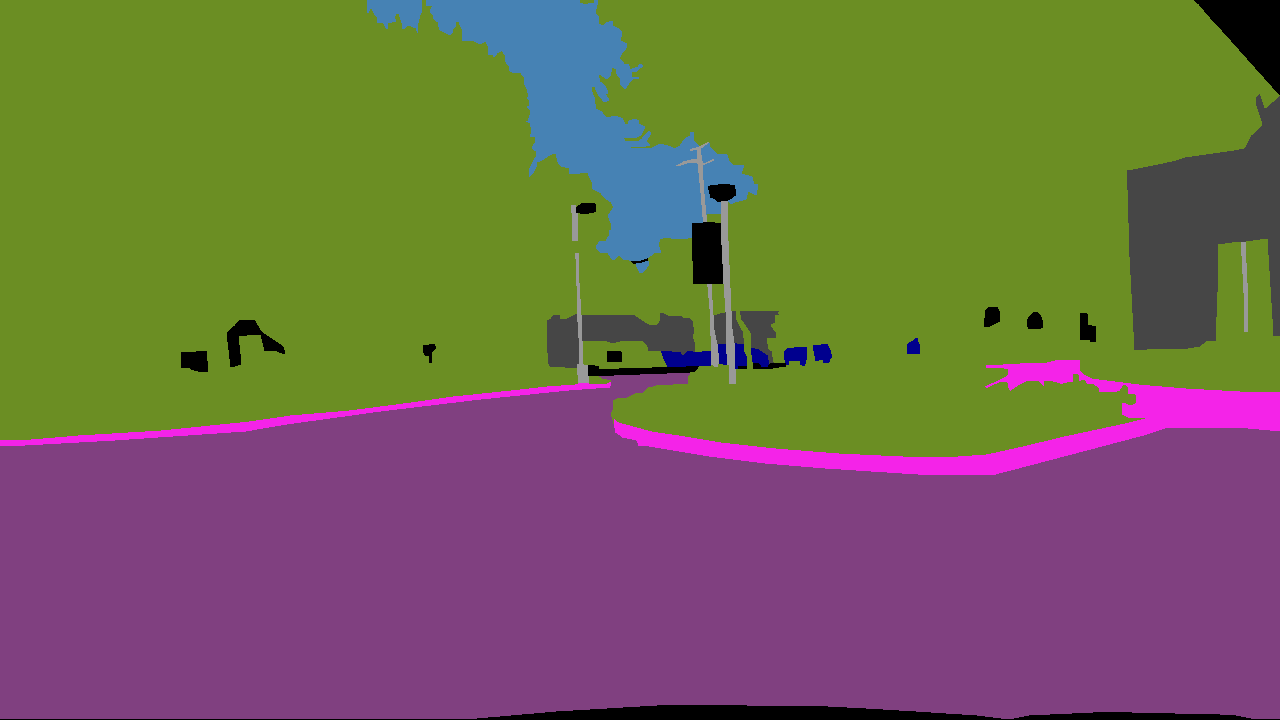} \\
  
          \includegraphics[width=.12\textwidth, height=1.8cm]{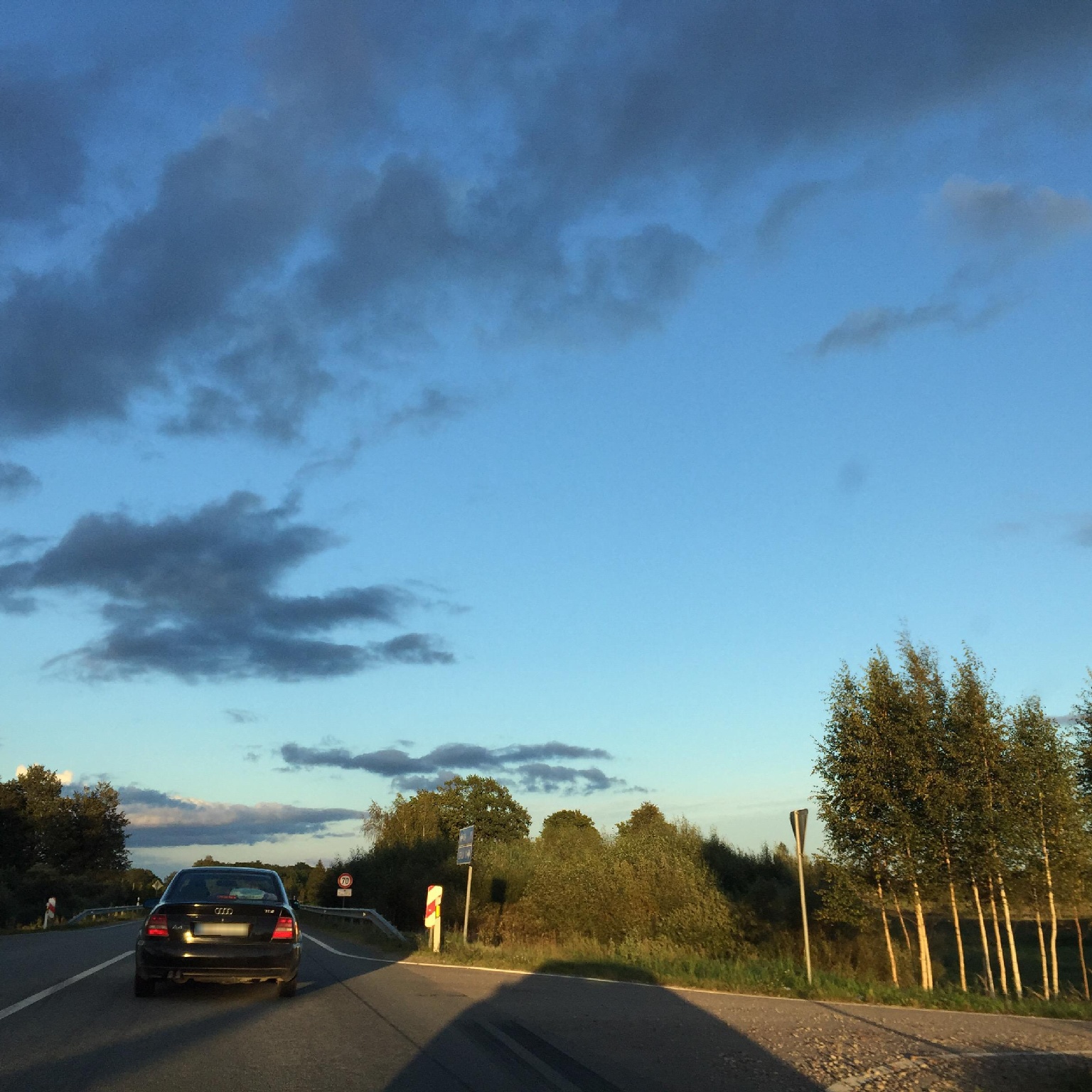} &
          \includegraphics[width=.12\textwidth, height=1.8cm]{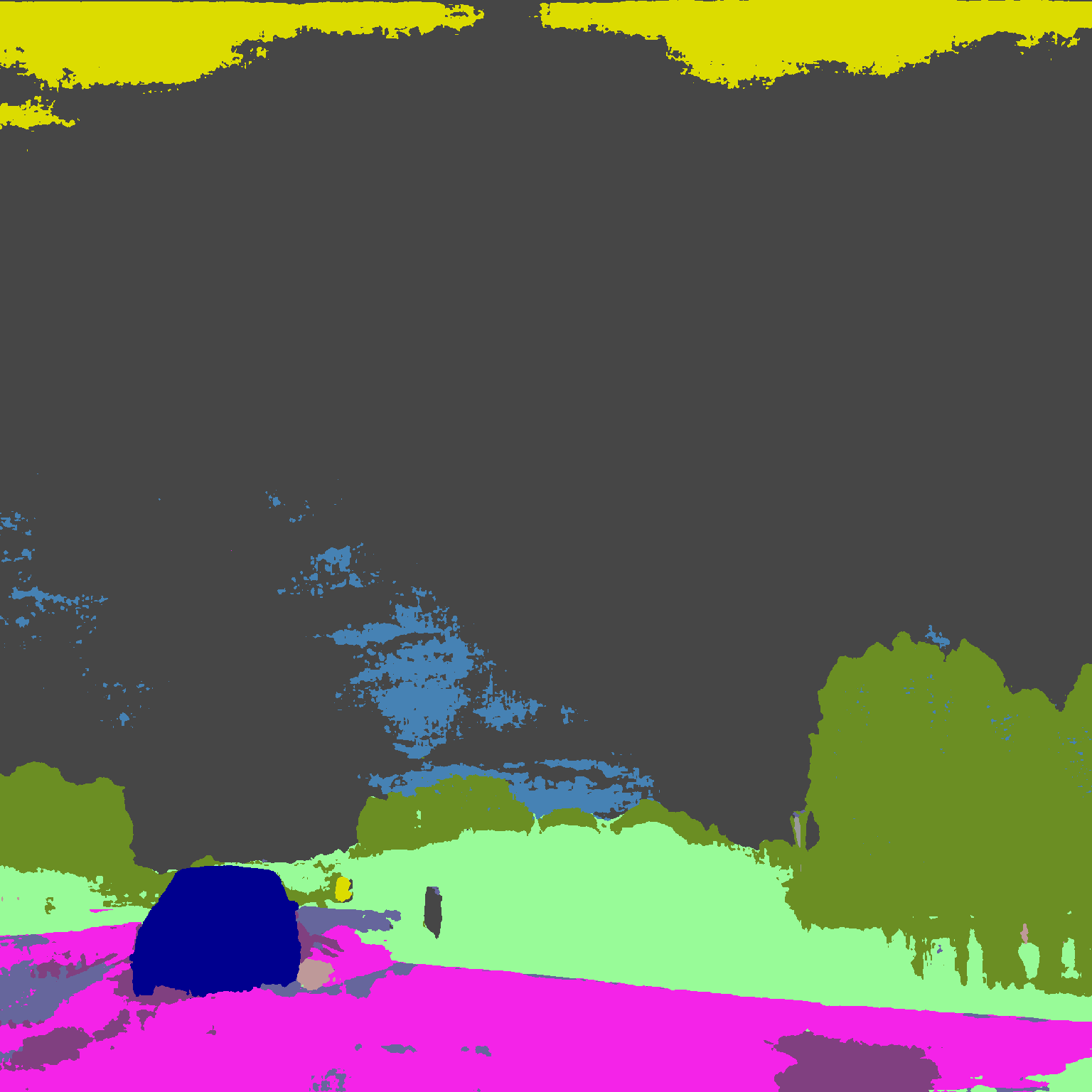} &
          \includegraphics[width=.12\textwidth, height=1.8cm]{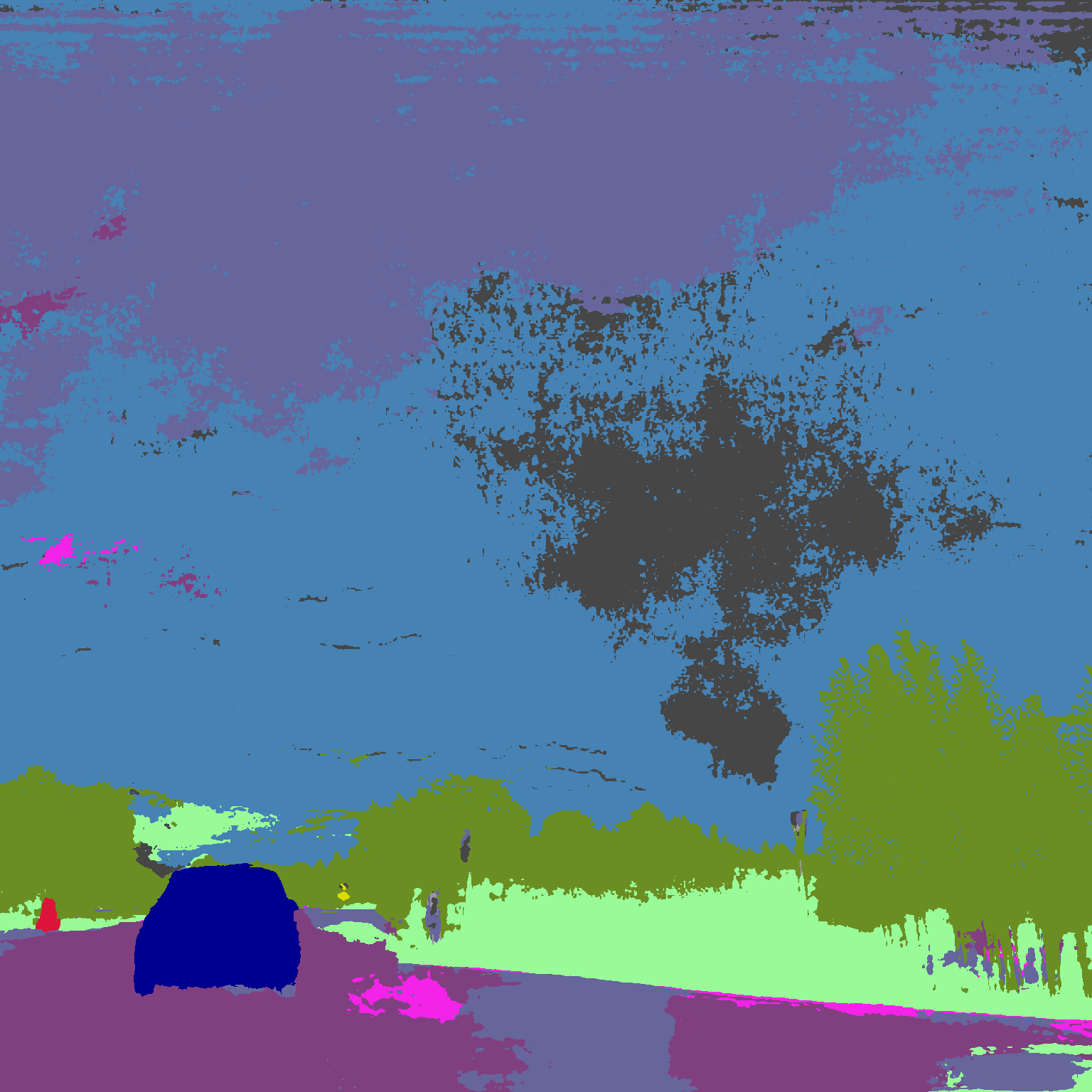} &
          \includegraphics[width=.12\textwidth, height=1.8cm]{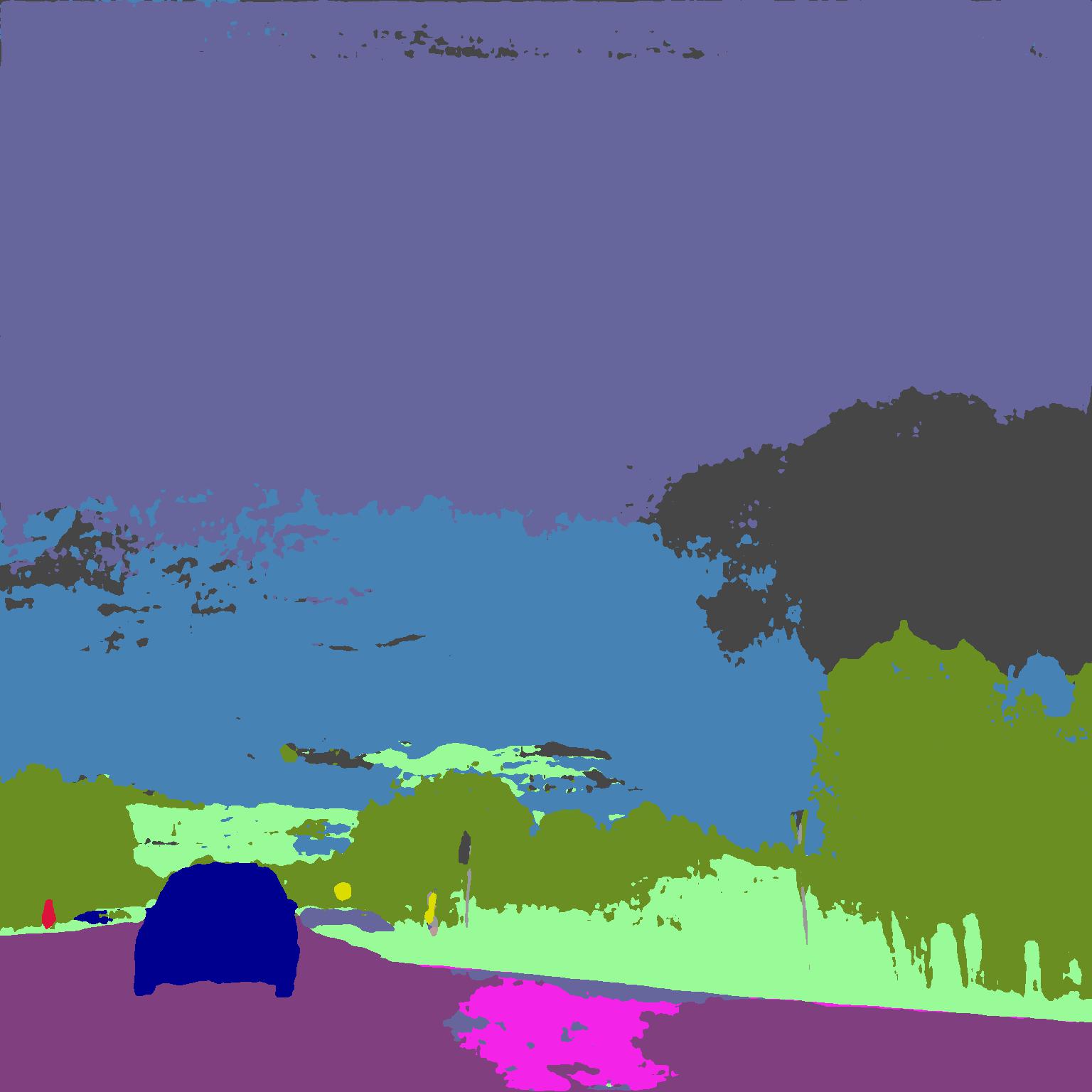} &
          \includegraphics[width=.12\textwidth, height=1.8cm]{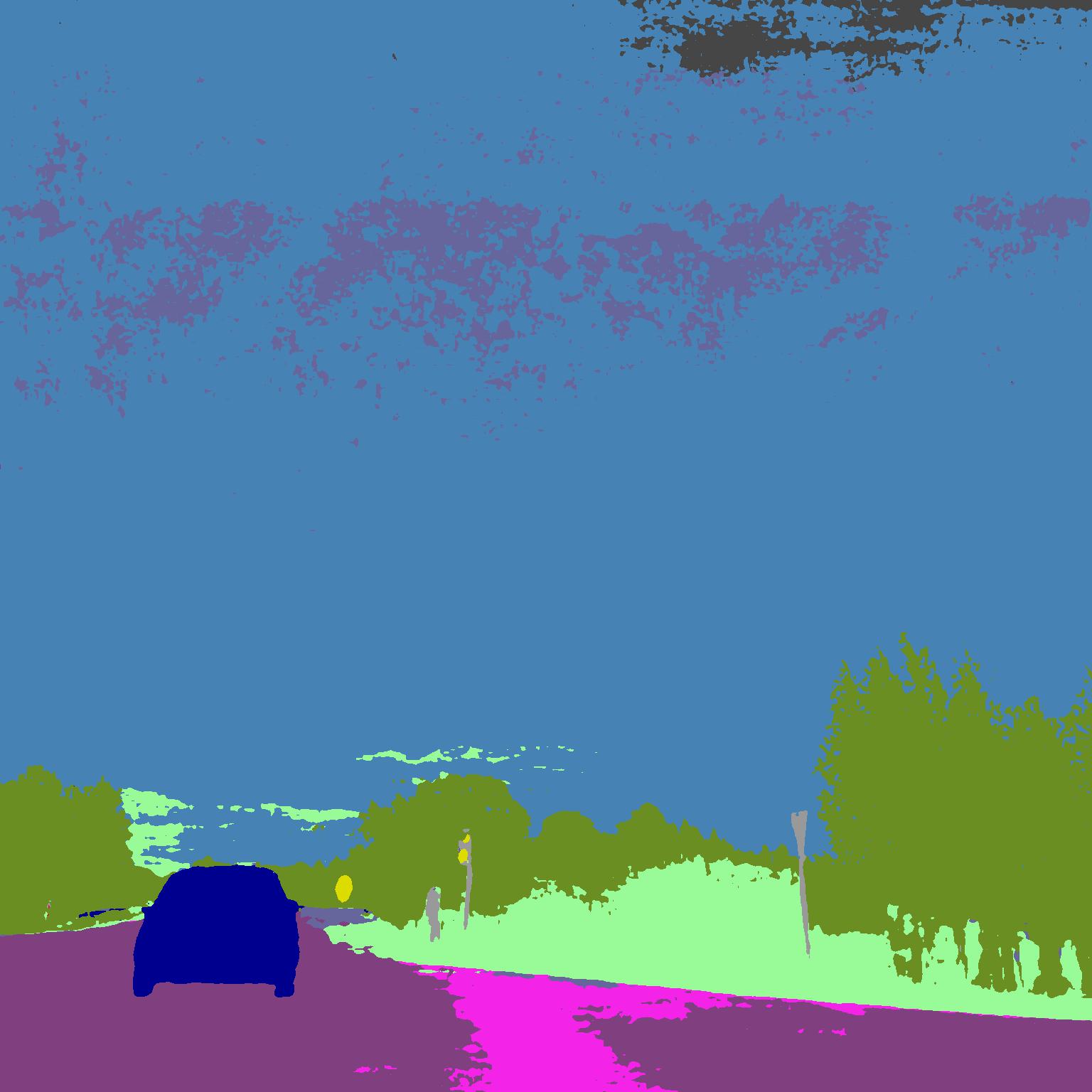} &
          \includegraphics[width=.12\textwidth, height=1.8cm]{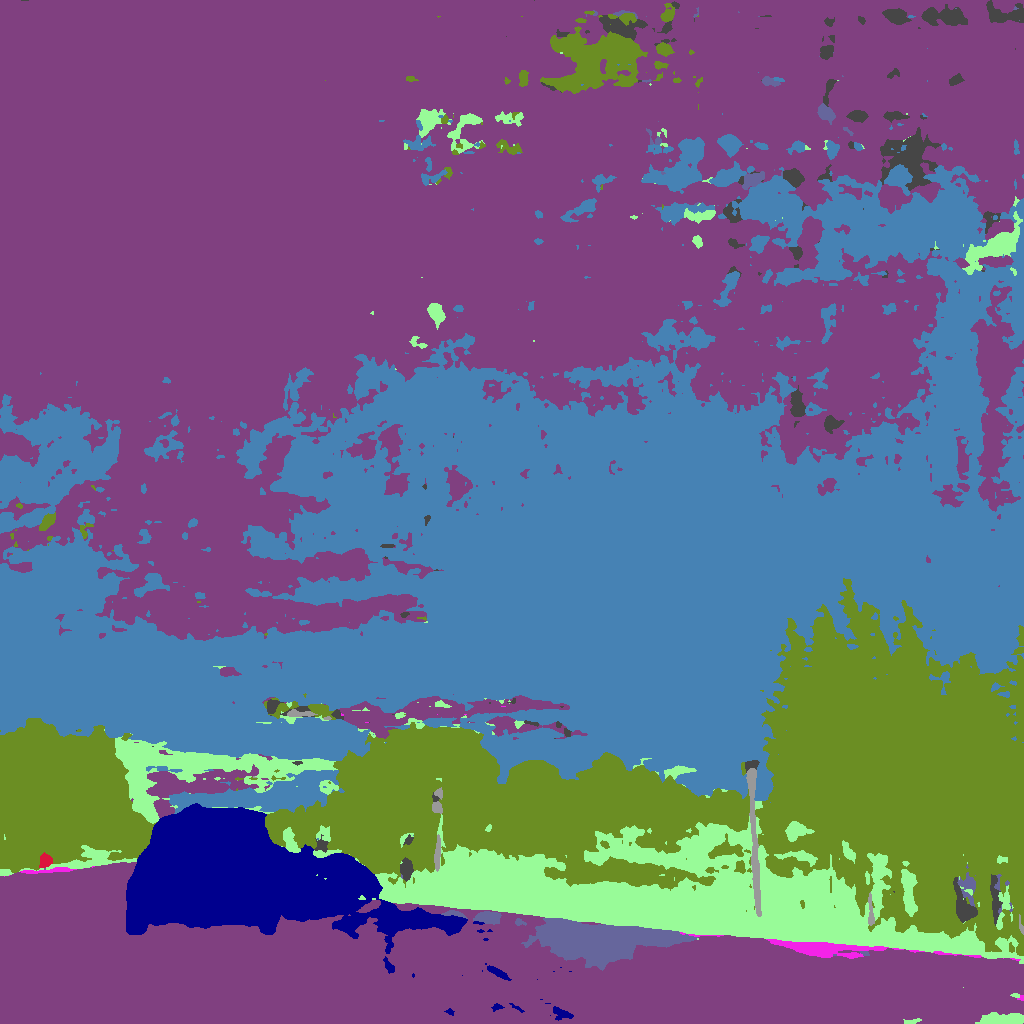} &
          \includegraphics[width=.12\textwidth, height=1.8cm]{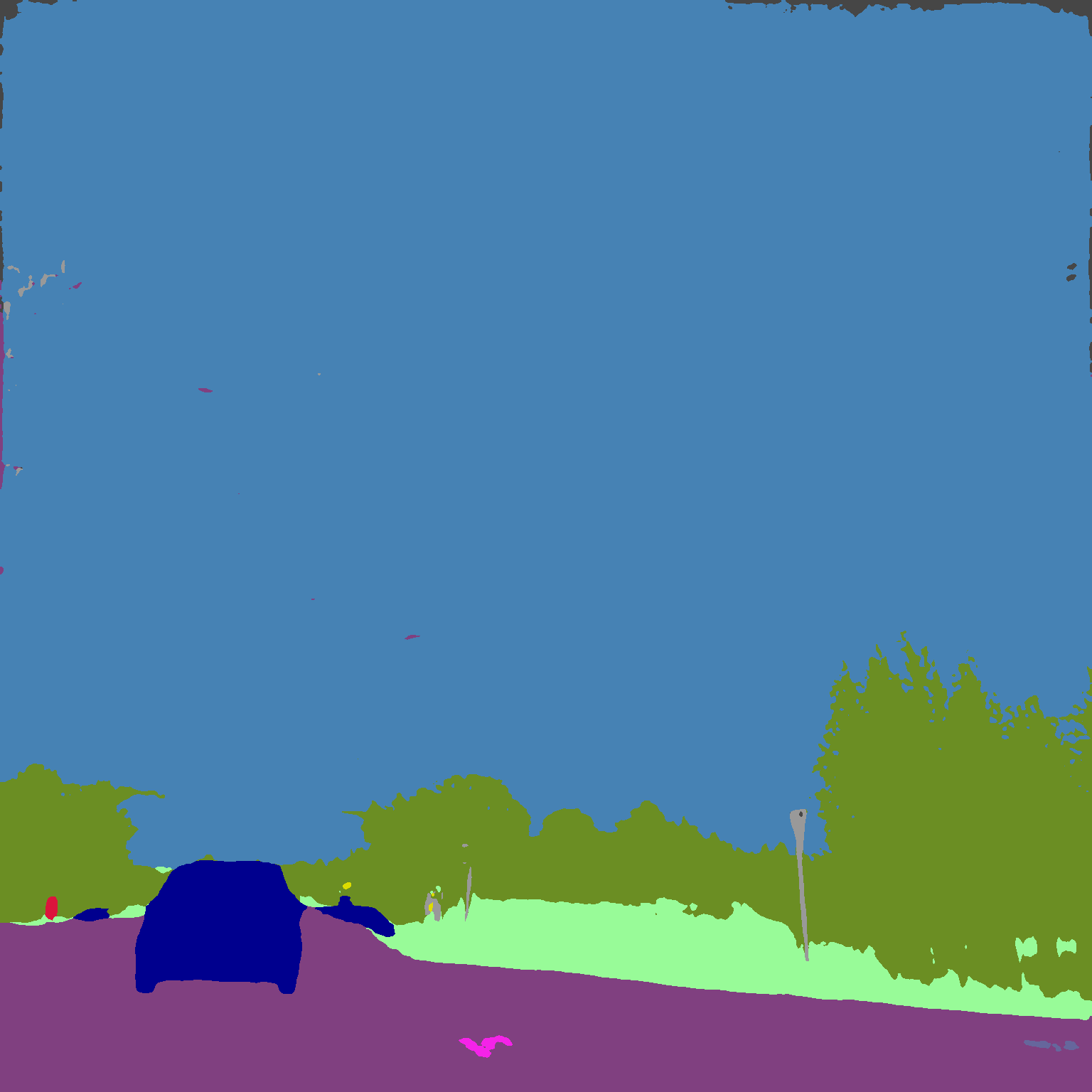} &
          \includegraphics[width=.12\textwidth, height=1.8cm]{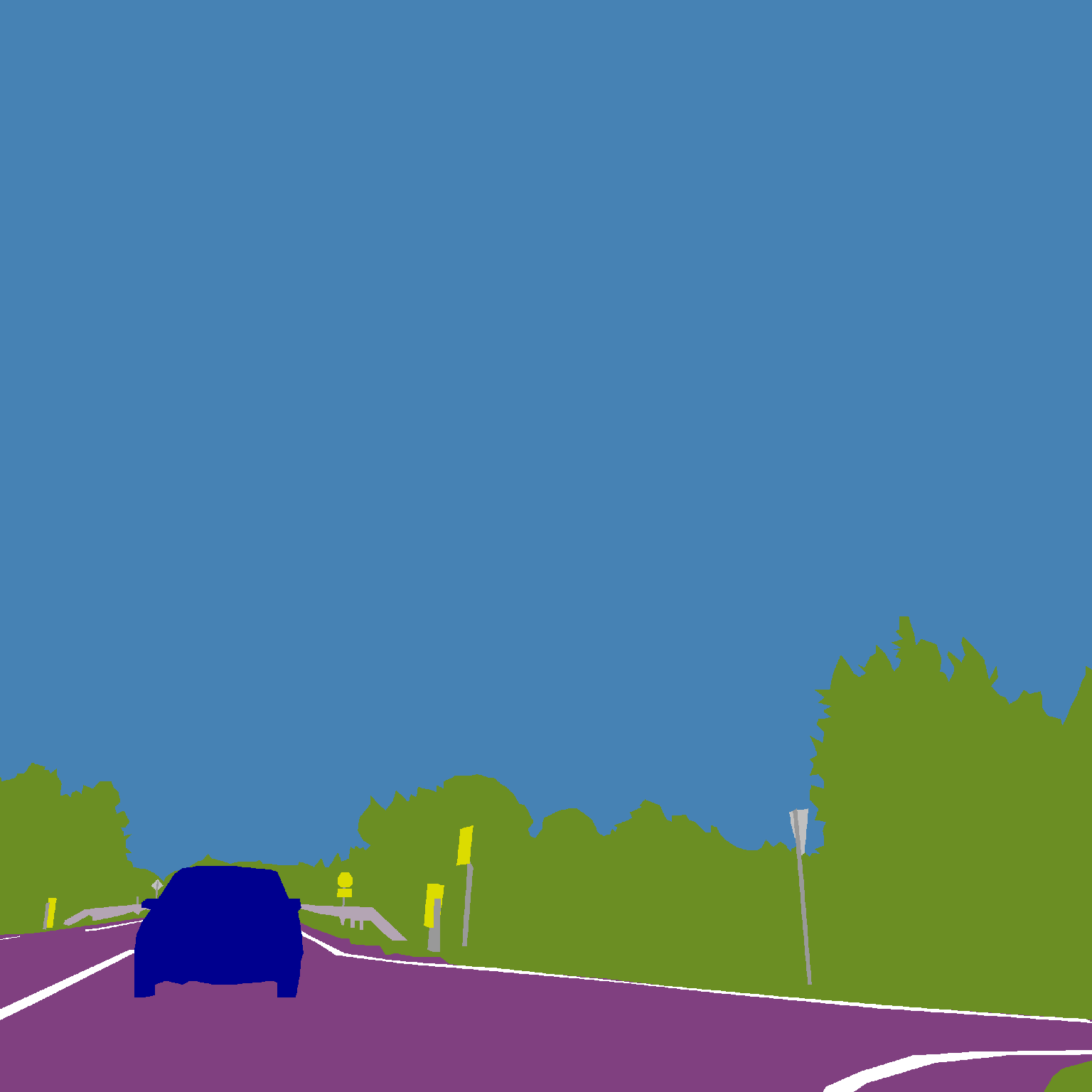} \\
          Input & DeepLabV3Plus~\cite{Chen2017RethinkingAC} & ISW~\cite{choi2021robustnet} & SHADE~\cite{Zhao2022StyleHallucinatedDC} & WildNet~\cite{Lee2022WildNetLD} & BlindNet~\cite{Ahn2024StyleBD} & \method{} & Ground Truth \\
          
      \end{tabular}
      \caption{\textbf{Qualitative comparison between DGSS methods trained on GTAV (G)~\cite{Richter2016PlayingFD} using DeepLabV3Plus~\cite{Chen2017RethinkingAC} with ResNet-50~\cite{He2015DeepRL}}. The first, second, and third rows show the predictions for \C{} (C)~\cite{Cordts2016TheCD}, \B{} (B)~\cite{Yu2018BDD100KAD}, and \M{} (M)~\cite{Neuhold2017TheMV}, respectively.
      }
      \label{fig:vis}
\end{figure*}

\section{Implementation Details}
\subsection{Datasets}
We conduct experiments on five urban scene semantic segmentation datasets that share 19 common scene categories.
\noindent\textbf{Cityscapes (C)}~\cite{Cordts2016TheCD} consists of high-resolution images captured from 50 distinct cities, where we exclusively use the finely annotated subset containing 2,975 training and 500 validation images.\newline  
\noindent\textbf{BDD-100K (B)}~\cite{Yu2018BDD100KAD} provides 7,000 training images and 1,000 validation images collected from various locations conditions.\newline 
\noindent\textbf{Mapillary (M)}~\cite{Neuhold2017TheMV} is a large-scale dataset containing 25,000 high-resolution street-view images sourced globally.\newline 
\noindent\textbf{GTA5 (G)}~\cite{Richter2016PlayingFD} includes 24,966 images simulated using the Grand Theft Auto V game engine, divided into 12,403 training, 6,382 validation, and 6,181 testing samples.\newline 
\noindent\textbf{SYNTHIA (S)}~\cite{Ros2016TheSD} consists of 9,400 photorealistic synthetic images, including 2,830 for testing. 

\subsection{Evaluation Protocols}
Following existing DGSS benchmarks~\cite{Pan2018TwoAO, Pan2019SwitchableWF, choi2021robustnet, peng2022semantic, Ding2023HGFormerHG, Ahn2024StyleBD}, we adopt a single-source domain generalization setting, where one dataset is used for training (source domain), and the remaining four datasets are treated as unseen target domains for evaluation. 
For a fair comparison, all experiments are built on DeepLabV3Plus~\cite{Chen2017RethinkingAC} (with ResNet-50~\cite{He2015DeepRL}) and Mask2Former~\cite{Cheng2021MaskedattentionMT} (with Swin-T and Swin-L~\cite{liu2021Swin}). 
We use the mean intersection over the union (mIoU)~\cite{Everingham2014ThePV} to measure the performance of segmentation.

\vspace{1mm} 
\subsection{Training Details}
To facilitate a fair evaluation, we exclusively employ Photometric Augmentation as the data augmentation strategy to generate pseudo-target domain samples. 
The model is trained with Adam Optimizer~\cite{Kingma2014AdamAM} with an initial learning rate of $1 \times 10^{-5}$, momentum parameters $(\beta_1, \beta_2) = (0.9, 0.999)$, and a batch size of 4 for 100 epochs. 
The channel of \LDP{} is set to $c' = 4$. In \DPE{}, we adopt a total diffusion timestep of $T = 4$, with the noise variance $\beta_t$ linearly increasing from $\beta_1 = 0.1$ to $\beta_T = 0.99$. 
The loss coefficients are adjusted to $(\lambda_\text{task}, \lambda_\text{sc}, \lambda_\text{prior}) = (0.5, 0.5, 1.0)$, balancing the contributions of different loss terms. 
All experiments are implemented in PyTorch and conducted on a single NVIDIA RTX 4090 GPU. More details of the implementation and hyperparameters analysis are provided in the Supplementary Material. 

%% file: Experimental_Results.tex
\section{Experiments}

\begin{table}
    \centering
    \scalebox{0.68}{
    \begin{tabular}{l|c|ccccc} 
        \toprule[1.3pt]
        \multirow{2}{*}{Method} & \multirow{2}{*}{Backbone} &   \multicolumn{5}{c}{Trained on Cityscapes (C)} \\
                                &                           &   B & M & G & S & Avg. \\ 
        \hline
        \hline
         DeepLabV3Plus~\cite{Chen2017RethinkingAC} & \multirow{14}{*}{Res50} & 44.96 & 51.68 & 42.55 & 23.29 & 40.62 \\
         IBN~\cite{Pan2018TwoAO}&  &  48.56&  57.04&  45.06& 26.14 & 44.20\\
         IW~\cite{Pan2019SwitchableWF}&  &  48.49&   55.82&  44.87& 26.10 & 43.82\\
         Iternorm~\cite{Huang2019IterativeNB}&  &  49.23&  56.26&  45.73& 25.98 & 44.30\\
         DRPC~\cite{Yue2019DomainRA}&  &  49.86&   56.34&  45.62& 26.58 & 44.60\\
         ISW~\cite{choi2021robustnet}&  &  50.73&  58.64&   45.00&  26.20 & 45.14\\
         GTR~\cite{peng2021global}&  &  50.75&   57.16&  45.79&  26.47 & 45.04\\
         DIRL~\cite{Xu2022DIRLDR}&  &  51.80&  - &   46.52& 26.50 & - \\
         SHADE~\cite{Zhao2022StyleHallucinatedDC}&  &  50.95&  \textcolor{magenta}{60.67}&  \textcolor{magenta}{48.61}&  27.62 & 46.96 \\
         SAW~\cite{Peng2022SemanticAwareDG}&  &  \textcolor{magenta}{52.95}&   59.81&   47.28& 28.32 & 47.09 \\
         WildNet~\cite{Lee2022WildNetLD}&  &  50.94&  58.79&  47.01& 27.95 & 46.17 \\
         DPCL~\cite{Yang2023GeneralizedSS}&  &  50.97&  58.59&  46.00& 25.85 & 45.34 \\
         BlindNet~\cite{Ahn2024StyleBD}&  &  51.84&  60.18&  47.97& \textcolor{magenta}{28.51} & \textcolor{magenta}{47.13} \\
         \rowcolor{LightCyan} \textbf{\method{}}&  &\textcolor{blue}{53.50}&\textcolor{blue}{62.93}&\textcolor{blue}{50.54}&\textcolor{blue}{30.68} & \textcolor{blue}{49.41}\\
         \hdashline
         Mask2Former~\cite{Cheng2021MaskedattentionMT}&  \multirow{3}{*}{Swin-T}&  51.30 &  65.30&   50.60&   \textcolor{magenta}{34.00} & 50.30 \\
         HGFormer~\cite{Ding2023HGFormerHG}& &  \textcolor{magenta}{53.40}&  \textcolor{blue}{66.90}&   \textcolor{magenta}{51.30}&   33.60 & \textcolor{magenta}{51.30}\\
         \rowcolor{LightCyan} \textbf{\method{}} &  & \textcolor{blue}{56.40} & \textcolor{magenta}{66.40} & \textcolor{blue}{55.20} & \textcolor{blue}{34.80} & \textcolor{blue}{53.20}\\
         \hdashline
         Mask2Former~\cite{Cheng2021MaskedattentionMT}& \multirow{4}{*}{Swin-L} &  60.10 &  72.20 &   57.80 &   42.40 & 58.13 \\
         HGFormer~\cite{Ding2023HGFormerHG}&  &  61.50 &  72.10 &   59.40 &   41.30 & 58.58 \\
         CMFormer~\cite{Bi2023LearningCM}&  &  \textcolor{magenta}{62.60}&  \textcolor{magenta}{73.60}&   \textcolor{magenta}{60.70}&   \textcolor{magenta}{43.00} & \textcolor{magenta}{59.98}\\
         \rowcolor{LightCyan} \textbf{\method{}} &  & \textcolor{blue}{63.00} & \textcolor{blue}{74.10} & \textcolor{blue}{63.20} & \textcolor{blue}{44.00} & \textcolor{blue}{61.08}\\
        \bottomrule[1.3pt]
    \end{tabular}}
    \caption{\textbf{Comparison with existing methods trained on Cityscapes (C)~\cite{Cordts2016TheCD}.} The best and second-best mIoU scores are \textcolor{blue}{blue} and \textcolor{magenta}{magenta}.}
    \label{tab:city}
\end{table}

\begin{table}
    \centering
    \scalebox{0.8}{
    \begin{tabular}{l|ccccc} 
        \toprule[1.3pt]
        \multirow{2}{*}{Method} &   \multicolumn{5}{c}{Trained on GTAV (G)} \\
                                & C & B & M & S & Avg.        \\ 
        \hline\hline
         DeepLabV3Plus~\cite{Chen2017RethinkingAC} & 28.95 & 25.14 & 28.18 & 26.23 & 27.13 \\
         IBN~\cite{Pan2018TwoAO}         &  33.85&  32.30&  37.75& 27.90 & 32.95\\
         IW~\cite{Pan2019SwitchableWF}  &  29.91&   27.48&  29.71& 27.61& 28.68\\
         Iternorm~\cite{Huang2019IterativeNB}&  31.81&  32.70&  33.88& 27.07 & 31.37\\
         DRPC~\cite{Yue2019DomainRA}&  37.42&  32.14&  34.12& 28.06 & 32.94\\
         ISW~\cite{choi2021robustnet}&  36.58&  35.20&  40.33& 28.30 & 35.10\\
         GTR~\cite{peng2021global}&  37.53&  33.75&  34.52& 28.17 & 33.49 \\
         DIRL~\cite{Xu2022DIRLDR}&  41.04&  39.15&  41.60& - & -\\
         SHADE~\cite{Zhao2022StyleHallucinatedDC}&  44.65&  39.28&  43.34& - & -\\
         SAW~\cite{Peng2022SemanticAwareDG}&  39.75&  37.34&  41.86& 30.79 & 37.44\\
         WildNet~\cite{Lee2022WildNetLD}&  44.62&   38.42&  46.09& 31.34 & 40.12\\
         AdvStyle~\cite{Zhong2022AdversarialSA}&  39.62&   35.54&  37.00&  - & -\\
         SPC~\cite{Huang2023StylePC}&  44.10&  40.46&  45.51& - & -\\
         DPCL~\cite{Yang2023GeneralizedSS}&  44.74&  40.59&  46.33& 30.81 & 40.62\\
         BlindNet~\cite{Ahn2024StyleBD}&  \textcolor{blue}{45.72}&  \textcolor{magenta}{41.32}&  \textcolor{magenta}{47.08}& \textcolor{magenta}{31.39} & \textcolor{magenta}{41.38} \\
         \rowcolor{LightCyan} \textbf{\method{}} & \textcolor{magenta}{45.40}& \textcolor{blue}{43.56}&\textcolor{blue}{49.45}&\textcolor{blue}{32.98} & \textcolor{blue}{42.85}\\
        \bottomrule[1.3pt]
    \end{tabular}
    }
    \caption{\textbf{Comparison with existing DGSS methods using a CNN-based network trained on GTAV (G)~\cite{Richter2016PlayingFD}.}}
    \label{tab:gta}
\end{table}

\subsection{Comparison with State-of-the-arts}
\label{sec:sota}
To access the effectiveness of \method{}, we compare our approach against a comprehensive set of state-of-the-art methods including both CNN-based techniques~\cite{Chen2017RethinkingAC, Pan2018TwoAO, Pan2019SwitchableWF, Huang2019IterativeNB, Yue2019DomainRA, choi2021robustnet, peng2021global, Xu2022DIRLDR, Zhao2022StyleHallucinatedDC, Peng2022SemanticAwareDG, Lee2022WildNetLD, Zhong2022AdversarialSA, Huang2023StylePC, Yang2023GeneralizedSS, Ahn2024StyleBD} and Mask2Former-based approaches~\cite{Cheng2021MaskedattentionMT, Ding2023HGFormerHG, Bi2023LearningCM}. 
We adhere to standard DGSS evaluation protocols and report results directly cited from prior works~\cite{Pan2019SwitchableWF, choi2021robustnet, Peng2022SemanticAwareDG, Bi2023LearningCM, Ahn2024StyleBD}.

\vspace{1mm} 
\noindent\textbf{\C{} Source Domain.}
\tabref{tab:city} shows that \method{} achieves the best overall performance across four unseen datasets, demonstrating strong generalization to diverse domains. 
\method{} effectively transfers from real-world datasets (C)~\cite{Cordts2016TheCD} to target domains with distinct visual distributions, such as game-rendered synthetic scenes (G)~\cite{Richter2016PlayingFD} and photo-realistic simulations (S)~\cite{Ros2016TheSD}. 
Notably, \method{} consistently improves performance across CNN-based and Mask2Former-based methods, highlighting its flexibility in enhancing existing segmentation networks. 

\vspace{1mm} 
\noindent\textbf{\G{} Source Domain.}
\tabref{tab:gta} demonstrates the effectiveness of \method{} in bridging the domain gap between synthetic and real-world datasets. 
Unlike existing DGSS methods that rely primarily on feature normalization, \method{} explicitly models latent domain priors, providing structured guidance for feature alignment and enhancing domain generalization. 
As shown in \figref{fig:vis}, qualitative comparisons indicate that \method{} effectively preserves detailed structural information—such as road boundaries and object contours—resulting in more precise and consistent segmentation outcomes.


\vspace{1mm} 
\noindent\textbf{Clear to Corruptions.}
We evaluate \method{} on ACDC~\cite{sakaridis2021acdc}, which introduces four real-world adverse conditions (rain, fog, night and snow). 
We employ DeepLabV3Plus~\cite{Chen2017RethinkingAC} and Mask2Former~\cite{Cheng2021MaskedattentionMT} as the segmentation network and train our method on Cityscapes~\cite{Cordts2016TheCD}. 
As shown in~\tabref{tab:acdc}, \method{} achieves consistent improvement under all adversarial conditions, demonstrating that \LDP{} effectively compensates for degraded features caused by image corruptions and enhances robustness in varied environments.

\begin{table}
    \centering
    \scalebox{0.7}{
    \begin{tabular}{l|c|ccccc} 
        \toprule[1.3pt]
        \multirow{2}{*}{Method} & \multirow{2}{*}{Backbone} & \multicolumn{5}{c}{Trained on Cityscape (C)} \\
                                &                     & Foggy & Night & Rain & Snow & Avg.        \\ 
        \hline\hline
         
         IBN~\cite{Pan2018TwoAO}&  \multirow{5}{*}{Res50}   &  63.80 &  21.20 &  50.40 & 49.60 & 43.70 \\
         Iternorm~\cite{Huang2019IterativeNB}&              &  63.30 &  23.80 &  50.10 & \textcolor{magenta}{49.90} & 45.30 \\
         IW~\cite{Pan2019SwitchableWF}&                     &  62.40 &  21.80 &  52.40 & 47.60 & 46.60 \\
         ISW~\cite{choi2021robustnet}&                      &  \textcolor{magenta}{64.30} &  \textcolor{magenta}{24.30} &  \textcolor{magenta}{56.00} & 49.80 & \textcolor{magenta}{48.10} \\
         \rowcolor{LightCyan} \textbf{\method{}}&  &\textcolor{blue}{68.28}&\textcolor{blue}{30.14}&\textcolor{blue}{59.23}&\textcolor{blue}{53.58} & \textcolor{blue}{52.81}\\
         \hdashline
         Mask2Former~\cite{Cheng2021MaskedattentionMT}&  \multirow{3}{*}{Swin-T}&  56.40 &  39.10 & 58.90 & 58.20 & 53.15 \\
         HGFormer~\cite{Ding2023HGFormerHG}& &  \textcolor{magenta}{58.50}&  \textcolor{magenta}{43.30}&   \textcolor{magenta}{62.00}&   \textcolor{magenta}{58.30} & \textcolor{magenta}{55.53}\\
         \rowcolor{LightCyan} \textbf{\method{}} &  & \textcolor{blue}{73.75} & \textcolor{blue}{47.75} & \textcolor{blue}{68.12} & \textcolor{blue}{62.25} & \textcolor{blue}{62.97}\\
         \hdashline
         Mask2Former~\cite{Cheng2021MaskedattentionMT}& \multirow{3}{*}{Swin-L} & 69.10 & \textcolor{magenta}{53.10} & 68.30 & 65.20 & 63.93 \\
         HGFormer~\cite{Ding2023HGFormerHG}&  & \textcolor{magenta}{69.90} & 52.70 & \textcolor{magenta}{72.00} & \textcolor{magenta}{68.60} & \textcolor{magenta}{65.80} \\
         \rowcolor{LightCyan} \textbf{\method{}} &  & \textcolor{blue}{80.72} & \textcolor{blue}{55.12} & \textcolor{blue}{73.13} & \textcolor{blue}{71.43} & \textcolor{blue}{70.10}\\
        \bottomrule[1.3pt]
    \end{tabular}
    }
    \caption{\textbf{Comparison with existing methods evaluated on ACDC~\cite{sakaridis2021acdc}.}}
    \label{tab:acdc}
\end{table}

\begin{table}
    \centering
    \scalebox{0.7}{
    \begin{tabular}{l|cccc|c} 
        \toprule[1.3pt]
        Methods & B & M & G & S & Avg. \\ 
        \hline
        Baseline             & 44.96 & 51.68 & 42.55 & 23.29 & 40.62 \\
        \method{} w/o \LPE{} & 49.59 & 57.63 & 46.34 & 27.19 & 45.19 \\
        \method{} w/o \DCM{} & 51.51 & \textcolor{magenta}{60.79} & \textcolor{magenta}{49.42} & \textcolor{magenta}{29.55} & \textcolor{magenta}{47.82} \\
        \method{} w/o \DPE{} & \textcolor{magenta}{51.98} & 60.33 & 49.38 & 28.86 & 47.64 \\
        \rowcolor{LightCyan} \textbf{\method{}} & \textcolor{blue}{53.50} & \textcolor{blue}{62.93} & \textcolor{blue}{50.54} & \textcolor{blue}{30.68} & \textcolor{blue}{49.41} \\
        \bottomrule[1.3pt]
    \end{tabular}
    }
    \caption{\textbf{Effectiveness of proposed modules in \method{}.}}
    \label{tab:modules}
\end{table}

\begin{table}
    \centering
    \scalebox{0.7}{
    \begin{tabular}{ccc|cccc|c} 
        \toprule[1.3pt]
        $\mathcal{L}_\text{task}$ & $\mathcal{L}_\text{sc}$ & $\mathcal{L}_\text{prior}$ & B & M & G & S & Avg. \\ 
        \hline
        \checkmark & - & -                      & 49.76 & 58.73 & 47.32 & 27.51 & 45.83 \\
        \checkmark & \checkmark & -             & 51.28 & 59.58 & 48.38 & 27.94 & 46.80 \\
        \checkmark & - & \checkmark             & \textcolor{magenta}{53.36} & \textcolor{magenta}{62.11} & \textcolor{magenta}{49.69} & \textcolor{magenta}{30.10} & \textcolor{magenta}{48.82} \\
        \rowcolor{LightCyan} \checkmark & \checkmark & \checkmark    & \textcolor{blue}{53.50} & \textcolor{blue}{62.93} & \textcolor{blue}{50.54} & \textcolor{blue}{30.68} & \textcolor{blue}{49.41} \\
        \bottomrule[1.3pt]
    \end{tabular}
    }
    \caption{\textbf{Analysis on the proposed loss functions.}}
    \label{tab:loss}
\end{table}

\subsection{Ablation Studies}
To validate the effectiveness of the proposed modules, we perform an ablation study. 
All experiments are conducted using DeepLabV3Plus~\cite{Chen2017RethinkingAC} with ResNet-50~\cite{He2015DeepRL} as the segmentation network. 
The model is trained on \C{} (C)~\cite{Cordts2016TheCD} and evaluated on \B{} (B)~\cite{Yu2018BDD100KAD}, \M{} (M)~\cite{Neuhold2017TheMV}, \G{} (G)~\cite{Richter2016PlayingFD}, and \SYN{} (S)~\cite{Ros2016TheSD} to assess generalization across diverse domain shifts.

\vspace{1mm} 
\noindent\textbf{Effectiveness of Proposed Modules.}
We consider several experimental configurations: (i) Baseline: vanilla DeepLabV3Plus~\cite{Chen2017RethinkingAC} trained without \method{}; (ii) \method{} w/o \LPE{}: where variational posterior is removed, resulting in \LDP{} estimation without cross-domain guidance; (iii) \method{} w/o \DCM{}: where \LDP{} is directly added to the original features without employing \DCM{}; (iv) \method{} w/o \DPE{}: where \DPE{} is replaced by a single-source \LPE{}, omitting the diffusion process; and (v) \method{}: which integrates all proposed modules. 
As shown in \tabref{tab:modules}, each module enhances segmentation performance, with \LPE{} playing a pivotal role in posterior regularization for structured \LDP{} modeling.

\begin{table}
    \centering
    \scalebox{0.75}{
    \begin{tabular}{c|cccccc} 
        \toprule[1.3pt]
        Number of $c'$ & 1 & 4 & 16 & 64 & 256 & 1024 \\ 
        \hline
        Avg. of mIoU & \textcolor{magenta}{48.74} & \textcolor{blue}{49.41} & 48.66 & 48.35 & 46.14 & 45.69 \\
        \bottomrule[1.3pt]
    \end{tabular}
    }
    \caption{\textbf{Ablation study on the number of channels for \LDP{}.}}
    \label{tab:LDP}
\end{table}

\begin{table}
    \centering
    \scalebox{0.75}{
    \begin{tabular}{c|cccccc} 
        \toprule[1.3pt]
        Number of Steps & 1 & 2 & 4 & 8 & 16 & 32 \\ 
        \hline
        Avg. of mIoU & 48.88 & 49.12 & 49.41 & 49.43 & \textcolor{magenta}{49.46} & \textcolor{blue}{49.48} \\
        \bottomrule[1.3pt]
    \end{tabular}
    }
    \caption{\textbf{Ablation study on the number of timestep in \DPE{}.}}
    \label{tab:steps}
\end{table}

\vspace{1mm} 
\noindent\textbf{Analysis of Loss Functions.}
\tabref{tab:loss} demonstrates that each loss term contributes to improved performance. 
The prior constraint loss, \(\mathcal{L}_\text{prior}\), enforces consistency between the \LDP{}s estimated by \LPE{} and \DPE{}, enabling \DPE{} to accurately infer \LDP{} solely from target samples. Additionally, the semantic consistency loss, \(\mathcal{L}_\text{sc}\), enhances feature alignment by constraining predictions from both source and pseudo-target samples, further bolstering domain generalization.

\vspace{1mm} 
\noindent\textbf{Impact of \LDP{} Channel Dimensions.}
\LDP{} serve as a variational prior for modeling domain shifts while guiding feature alignment in DGSS. 
The channel dimension of \LDP{} is crucial for capturing domain-specific variations. 
As shown in \tabref{tab:LDP}, a channel dimension of 4 yields the best overall performance. 
In contrast, higher channel dimensions introduce feature redundancy and increase the risk of overfitting, which limits further performance improvements.

\vspace{1mm} 
\noindent\textbf{Impact of steps for \DPE{}.}
\DPE{} iteratively refines \LDP{} through the denoising process, facilitating precise alignment across diverse target domains. 
As shown in \tabref{tab:steps}, performance gradually improves with an increasing number of diffusion steps. 
To achieve an optimal trade-off between efficiency and accuracy, we set the diffusion timestep to 4.

%% file: Conclusion.tex
\section{Conclusion}
This paper introduces PDAF, a probabilistic diffusion alignment framework designed for DGSS. By explicitly modeling latent domain prior (\LDP{}), \method{} captures domain-specific variations to facilitate robust feature alignment across unseen target domains. To achieve this, \method{} involves a latent prior extractor to infer cross-domain relationships, a domain compensation module to condition segmentation features on \LDP{} and a diffusion prior estimator to estimate \LDP{} through a denoising process, allowing robust generalization to unseen domains. Extensive experiments demonstrate the effectiveness of \method{}, improving the generalization of existing segmentation models across diverse environments, highlighting the robustness of \LDP{} modeling in domain generalized semantic segmentation. 

%% file: main.bbl
\begin{thebibliography}{76}
\providecommand{\natexlab}[1]{#1}
\providecommand{\url}[1]{\texttt{#1}}
\expandafter\ifx\csname urlstyle\endcsname\relax
  \providecommand{\doi}[1]{doi: #1}\else
  \providecommand{\doi}{doi: \begingroup \urlstyle{rm}\Url}\fi

\bibitem[Ahn et~al.(2024)Ahn, Yang, Choi, and Lim]{Ahn2024StyleBD}
Woojin Ahn, Geun~Yeong Yang, Hyun~Duck Choi, and Myo~Taeg Lim.
\newblock Style blind domain generalized semantic segmentation via covariance alignment and semantic consistence contrastive learning.
\newblock In \emph{CVPR}, 2024.

\bibitem[Balaji et~al.(2018)Balaji, Sankaranarayanan, and Chellappa]{Balaji2018MetaRegTD}
Yogesh Balaji, Swami Sankaranarayanan, and Rama Chellappa.
\newblock Metareg: Towards domain generalization using meta-regularization.
\newblock In \emph{NeurIPS}, 2018.

\bibitem[Bartoccioni et~al.(2023)Bartoccioni, Zablocki, Bursuc, P{\'e}rez, Cord, and Alahari]{bartoccioni2023lara}
Florent Bartoccioni, {\'E}loi Zablocki, Andrei Bursuc, Patrick P{\'e}rez, Matthieu Cord, and Karteek Alahari.
\newblock Lara: Latents and rays for multi-camera bird’s-eye-view semantic segmentation.
\newblock In \emph{Conference on robot learning}, 2023.

\bibitem[Benigmim et~al.(2024)Benigmim, Roy, Essid, Kalogeiton, and Lathuili{\`e}re]{benigmim2024collaborating}
Yasser Benigmim, Subhankar Roy, Slim Essid, Vicky Kalogeiton, and St{\'e}phane Lathuili{\`e}re.
\newblock Collaborating foundation models for domain generalized semantic segmentation.
\newblock In \emph{CVPR}, 2024.

\bibitem[Bi et~al.(2024)Bi, You, and Gevers]{Bi2023LearningCM}
Qi Bi, Shaodi You, and Theo Gevers.
\newblock Learning content-enhanced mask transformer for domain generalized urban-scene segmentation.
\newblock In \emph{AAAI}, 2024.

\bibitem[Chen et~al.(2025)Chen, Chen, Liu, Chiang, Kuo, Yang, et~al.]{chen2025unirestore}
I Chen, Wei-Ting Chen, Yu-Wei Liu, Yuan-Chun Chiang, Sy-Yen Kuo, Ming-Hsuan Yang, et~al.
\newblock Unirestore: Unified perceptual and task-oriented image restoration model using diffusion prior.
\newblock In \emph{CVPR}, 2025.

\bibitem[Chen et~al.(2017)Chen, Papandreou, Schroff, and Adam]{Chen2017RethinkingAC}
Liang-Chieh Chen, George Papandreou, Florian Schroff, and Hartwig Adam.
\newblock Rethinking atrous convolution for semantic image segmentation.
\newblock \emph{arXiv preprint arXiv:1706.05587}, 2017.

\bibitem[Chen et~al.(2022{\natexlab{a}})Chen, Chen, Yeh, Yang, Chang, Ding, and Kuo]{chen2022rvsl}
Wei-Ting Chen, I-Hsiang Chen, Chih-Yuan Yeh, Hao-Hsiang Yang, Hua-En Chang, Jian-Jiun Ding, and Sy-Yen Kuo.
\newblock Rvsl: Robust vehicle similarity learning in real hazy scenes based on semi-supervised learning.
\newblock In \emph{ECCV}, 2022{\natexlab{a}}.

\bibitem[Chen et~al.(2022{\natexlab{b}})Chen, Chen, Yeh, Yang, Ding, and Kuo]{chen2022sjdl}
Wei-Ting Chen, I-Hsiang Chen, Chih-Yuan Yeh, Hao-Hsiang Yang, Jian-Jiun Ding, and Sy-Yen Kuo.
\newblock Sjdl-vehicle: Semi-supervised joint defogging learning for foggy vehicle re-identification.
\newblock In \emph{AAAI}, 2022{\natexlab{b}}.

\bibitem[Cheng et~al.(2021)Cheng, Misra, Schwing, Kirillov, and Girdhar]{Cheng2021MaskedattentionMT}
Bowen Cheng, Ishan Misra, Alexander~G. Schwing, Alexander Kirillov, and Rohit Girdhar.
\newblock Masked-attention mask transformer for universal image segmentation.
\newblock In \emph{CVPR}, 2021.

\bibitem[Choi et~al.(2021)Choi, Jung, Yun, Kim, Kim, and Choo]{choi2021robustnet}
Sungha Choi, Sanghun Jung, Huiwon Yun, Joanne~T Kim, Seungryong Kim, and Jaegul Choo.
\newblock Robustnet: Improving domain generalization in urban-scene segmentation via instance selective whitening.
\newblock In \emph{CVPR}, 2021.

\bibitem[Cordts et~al.(2016)Cordts, Omran, Ramos, Rehfeld, Enzweiler, Benenson, Franke, Roth, and Schiele]{Cordts2016TheCD}
Marius Cordts, Mohamed Omran, Sebastian Ramos, Timo Rehfeld, Markus Enzweiler, Rodrigo Benenson, Uwe Franke, Stefan Roth, and Bernt Schiele.
\newblock The cityscapes dataset for semantic urban scene understanding.
\newblock In \emph{CVPR}, 2016.

\bibitem[Deng et~al.(2009)Deng, Dong, Socher, Li, Li, and Fei-Fei]{5206848}
Jia Deng, Wei Dong, Richard Socher, Li-Jia Li, Kai Li, and Li Fei-Fei.
\newblock Imagenet: A large-scale hierarchical image database.
\newblock In \emph{CVPR}, 2009.

\bibitem[Dhariwal and Nichol(2021)]{Dhariwal2021DiffusionMB}
Prafulla Dhariwal and Alex Nichol.
\newblock Diffusion models beat gans on image synthesis.
\newblock In \emph{NeurIPS}, 2021.

\bibitem[Ding et~al.(2023)Ding, Xue, Xia, Schiele, and Dai]{Ding2023HGFormerHG}
Jian Ding, Nan Xue, Guisong Xia, Bernt Schiele, and Dengxin Dai.
\newblock Hgformer: Hierarchical grouping transformer for domain generalized semantic segmentation.
\newblock In \emph{CVPR}, 2023.

\bibitem[Dou et~al.(2019)Dou, de~Castro, Kamnitsas, and Glocker]{Dou2019DomainGV}
Qi Dou, Daniel~Coelho de Castro, Konstantinos Kamnitsas, and Ben Glocker.
\newblock Domain generalization via model-agnostic learning of semantic features.
\newblock In \emph{NeurIPS}, 2019.

\bibitem[Everingham et~al.(2014)Everingham, Eslami, Gool, Williams, Winn, and Zisserman]{Everingham2014ThePV}
Mark Everingham, S.~M.~Ali Eslami, Luc~Van Gool, Christopher K.~I. Williams, John~M. Winn, and Andrew Zisserman.
\newblock The pascal visual object classes challenge: A retrospective.
\newblock \emph{IJCV}, 2014.

\bibitem[Ganin and Lempitsky(2015)]{ganin2015unsupervised}
Yaroslav Ganin and Victor Lempitsky.
\newblock Unsupervised domain adaptation by backpropagation.
\newblock In \emph{ICML}, 2015.

\bibitem[Ghifary et~al.(2015)Ghifary, Kleijn, Zhang, and Balduzzi]{Ghifary2015DomainGF}
Muhammad Ghifary, W. Kleijn, Mengjie Zhang, and David Balduzzi.
\newblock Domain generalization for object recognition with multi-task autoencoders.
\newblock In \emph{ICCV}, 2015.

\bibitem[Gong et~al.(2023)Gong, Danelljan, Sun, Mangas, and Van~Gool]{gong2023prompting}
Rui Gong, Martin Danelljan, Han Sun, Julio~Delgado Mangas, and Luc Van~Gool.
\newblock Prompting diffusion representations for cross-domain semantic segmentation.
\newblock \emph{arXiv preprint arXiv:2307.02138}, 2023.

\bibitem[He et~al.(2015)He, Zhang, Ren, and Sun]{He2015DeepRL}
Kaiming He, X. Zhang, Shaoqing Ren, and Jian Sun.
\newblock Deep residual learning for image recognition.
\newblock In \emph{CVPR}, 2015.

\bibitem[Ho et~al.(2020)Ho, Jain, and Abbeel]{Ho2020DenoisingDP}
Jonathan Ho, Ajay Jain, and P. Abbeel.
\newblock Denoising diffusion probabilistic models.
\newblock In \emph{NeurIPS}, 2020.

\bibitem[Hu et~al.(2023)Hu, Yang, Chen, Li, Sima, Zhu, Chai, Du, Lin, Wang, et~al.]{hu2023planning}
Yihan Hu, Jiazhi Yang, Li Chen, Keyu Li, Chonghao Sima, Xizhou Zhu, Siqi Chai, Senyao Du, Tianwei Lin, Wenhai Wang, et~al.
\newblock Planning-oriented autonomous driving.
\newblock In \emph{CVPR}, 2023.

\bibitem[Huang et~al.(2021)Huang, Guan, Xiao, and Lu]{Huang2021FSDRFS}
Jiaxing Huang, Dayan Guan, Aoran Xiao, and Shijian Lu.
\newblock Fsdr: Frequency space domain randomization for domain generalization.
\newblock In \emph{CVPR}, 2021.

\bibitem[Huang et~al.(2019)Huang, Zhou, Zhu, Liu, and Shao]{Huang2019IterativeNB}
Lei Huang, Yi Zhou, Fan Zhu, Li Liu, and Ling Shao.
\newblock Iterative normalization: Beyond standardization towards efficient whitening.
\newblock In \emph{CVPR}, 2019.

\bibitem[Huang et~al.(2023)Huang, Chen, Li, Li, Li, Song, Yan, and Xiong]{Huang2023StylePC}
Wei Huang, Chang~Wen Chen, Yong Li, Jiacheng Li, Cheng Li, Fenglong Song, Youliang Yan, and Zhiwei Xiong.
\newblock Style projected clustering for domain generalized semantic segmentation.
\newblock In \emph{CVPR}, 2023.

\bibitem[Ioffe and Szegedy(2015)]{Ioffe2015BatchNA}
Sergey Ioffe and Christian Szegedy.
\newblock Batch normalization: Accelerating deep network training by reducing internal covariate shift.
\newblock In \emph{ICML}, 2015.

\bibitem[Ji et~al.(2024)Ji, He, Qu, Tan, Qin, and Wu]{ji2024diffusion}
Yuxiang Ji, Boyong He, Chenyuan Qu, Zhuoyue Tan, Chuan Qin, and Liaoni Wu.
\newblock Diffusion features to bridge domain gap for semantic segmentation.
\newblock \emph{arXiv preprint arXiv:2406.00777}, 2024.

\bibitem[Jia et~al.(2024)Jia, Hoyer, Huang, Wang, Van~Gool, Schindler, and Obukhov]{jia2024dginstyle}
Yuru Jia, Lukas Hoyer, Shengyu Huang, Tianfu Wang, Luc Van~Gool, Konrad Schindler, and Anton Obukhov.
\newblock Dginstyle: Domain-generalizable semantic segmentation with image diffusion models and stylized semantic control.
\newblock In \emph{ECCV}, 2024.

\bibitem[Jiang et~al.(2024)Jiang, Mao, Pan, Han, and Zhang]{jiang2024scedit}
Zeyinzi Jiang, Chaojie Mao, Yulin Pan, Zhen Han, and Jingfeng Zhang.
\newblock Scedit: Efficient and controllable image diffusion generation via skip connection editing.
\newblock In \emph{CVPR}, 2024.

\bibitem[Kingma and Ba(2014)]{Kingma2014AdamAM}
Diederik~P. Kingma and Jimmy Ba.
\newblock Adam: A method for stochastic optimization.
\newblock In \emph{ICLR}, 2014.

\bibitem[Kingma and Welling(2013)]{Kingma2013AutoEncodingVB}
Diederik~P. Kingma and Max Welling.
\newblock Auto-encoding variational bayes.
\newblock In \emph{ICLR}, 2013.

\bibitem[Kingma et~al.(2021)Kingma, Salimans, Poole, and Ho]{Kingma2021VariationalDM}
Diederik~P. Kingma, Tim Salimans, Ben Poole, and Jonathan Ho.
\newblock Variational diffusion models.
\newblock \emph{arXiv preprint arXiv:2107.00630}, 2021.

\bibitem[Lee et~al.(2022)Lee, Seong, Lee, and Kim]{Lee2022WildNetLD}
Suhyeon Lee, Hongje Seong, Seongwon Lee, and Euntai Kim.
\newblock Wildnet: Learning domain generalized semantic segmentation from the wild.
\newblock In \emph{CVPR}, 2022.

\bibitem[Li et~al.(2017)Li, Yang, Song, and Hospedales]{Li2017LearningTG}
Da Li, Yongxin Yang, Yi-Zhe Song, and Timothy~M. Hospedales.
\newblock Learning to generalize: Meta-learning for domain generalization.
\newblock In \emph{AAAI}, 2017.

\bibitem[Li et~al.(2018{\natexlab{a}})Li, Pan, Wang, and Kot]{Li2018DomainGW}
Haoliang Li, Sinno~Jialin Pan, Shiqi Wang, and Alex~Chichung Kot.
\newblock Domain generalization with adversarial feature learning.
\newblock In \emph{CVPR}, 2018{\natexlab{a}}.

\bibitem[Li et~al.(2018{\natexlab{b}})Li, Tian, Gong, Liu, Liu, Zhang, and Tao]{Li2018DeepDG}
Ya Li, Xinmei Tian, Mingming Gong, Yajing Liu, Tongliang Liu, Kun Zhang, and Dacheng Tao.
\newblock Deep domain generalization via conditional invariant adversarial networks.
\newblock In \emph{ECCV}, 2018{\natexlab{b}}.

\bibitem[Liu et~al.(2024{\natexlab{a}})Liu, Zhuang, Gao, and Qin]{Liu2024CDFormerWhenDP}
Qingguo Liu, Chenyi Zhuang, Pan Gao, and Jie Qin.
\newblock Cdformer:when degradation prediction embraces diffusion model for blind image super-resolution.
\newblock In \emph{CVPR}, 2024{\natexlab{a}}.

\bibitem[Liu et~al.(2024{\natexlab{b}})Liu, Zhou, Liu, Hao, Fan, and Tian]{liu2024unbiased}
Yajing Liu, Shijun Zhou, Xiyao Liu, Chunhui Hao, Baojie Fan, and Jiandong Tian.
\newblock Unbiased faster r-cnn for single-source domain generalized object detection.
\newblock In \emph{CVPR}, 2024{\natexlab{b}}.

\bibitem[Liu et~al.(2021)Liu, Lin, Cao, Hu, Wei, Zhang, Lin, and Guo]{liu2021Swin}
Ze Liu, Yutong Lin, Yue Cao, Han Hu, Yixuan Wei, Zheng Zhang, Stephen Lin, and Baining Guo.
\newblock Swin transformer: Hierarchical vision transformer using shifted windows.
\newblock In \emph{ICCV}, 2021.

\bibitem[Motiian et~al.(2017)Motiian, Piccirilli, Adjeroh, and Doretto]{Motiian2017UnifiedDS}
Saeid Motiian, Marco Piccirilli, Donald~A. Adjeroh, and Gianfranco Doretto.
\newblock Unified deep supervised domain adaptation and generalization.
\newblock In \emph{ICCV}, 2017.

\bibitem[Neuhold et~al.(2017)Neuhold, Ollmann, Bul{\`o}, and Kontschieder]{Neuhold2017TheMV}
Gerhard Neuhold, Tobias Ollmann, Samuel~Rota Bul{\`o}, and Peter Kontschieder.
\newblock The mapillary vistas dataset for semantic understanding of street scenes.
\newblock In \emph{ICCV}, 2017.

\bibitem[Nilsson et~al.(2021)Nilsson, Pirinen, G{\"a}rtner, and Sminchisescu]{nilsson2021embodied}
David Nilsson, Aleksis Pirinen, Erik G{\"a}rtner, and Cristian Sminchisescu.
\newblock Embodied visual active learning for semantic segmentation.
\newblock In \emph{AAAI}, 2021.

\bibitem[Onozuka et~al.(2021)Onozuka, Matsumi, and Shino]{onozuka2021autonomous}
Yuya Onozuka, Ryosuke Matsumi, and Motoki Shino.
\newblock Autonomous mobile robot navigation independent of road boundary using driving recommendation map.
\newblock In \emph{Int. Conf. on Intel. Robots and Systems}, 2021.

\bibitem[Pan et~al.(2018)Pan, Luo, Shi, and Tang]{Pan2018TwoAO}
Xingang Pan, Ping Luo, Jianping Shi, and Xiaoou Tang.
\newblock Two at once: Enhancing learning and generalization capacities via ibn-net.
\newblock In \emph{ECCV}, 2018.

\bibitem[Pan et~al.(2019)Pan, Zhan, Shi, Tang, and Luo]{Pan2019SwitchableWF}
Xingang Pan, Xiaohang Zhan, Jianping Shi, Xiaoou Tang, and Ping Luo.
\newblock Switchable whitening for deep representation learning.
\newblock In \emph{ICCV}, 2019.

\bibitem[Peng et~al.(2021)Peng, Lei, Liu, Zhang, and Liua]{peng2021global}
Duo Peng, Yinjie Lei, Lingqiao Liu, Pingping Zhang, and Jun Liua.
\newblock Global and local texture randomization for synthetic-to-real semantic segmentation.
\newblock \emph{IEEE TIP}, 2021.

\bibitem[Peng et~al.(2022{\natexlab{a}})Peng, Lei, Hayat, Guo, and Li]{Peng2022SemanticAwareDG}
Duo Peng, Yinjie Lei, Munawar Hayat, Yulan Guo, and Wen Li.
\newblock Semantic-aware domain generalized segmentation.
\newblock In \emph{CVPR}, 2022{\natexlab{a}}.

\bibitem[Peng et~al.(2022{\natexlab{b}})Peng, Lei, Hayat, Guo, and Li]{peng2022semantic}
Duo Peng, Yinjie Lei, Munawar Hayat, Yulan Guo, and Wen Li.
\newblock Semantic-aware domain generalized segmentation.
\newblock In \emph{CVPR}, 2022{\natexlab{b}}.

\bibitem[Pernias et~al.(2023)Pernias, Rampas, Richter, Pal, and Aubreville]{Pernias2023WuerstchenAE}
Pablo Pernias, Dominic Rampas, Mats~L. Richter, Christopher~J. Pal, and Marc Aubreville.
\newblock Wuerstchen: An efficient architecture for large-scale text-to-image diffusion models.
\newblock \emph{arXiv preprint arXiv:2306.00637v2}, 2023.

\bibitem[Qu et~al.(2024)Qu, Zou, He, R{\"o}hrbein, Knoll, Chen, and Jiang]{qu2024lead}
Sanqing Qu, Tianpei Zou, Lianghua He, Florian R{\"o}hrbein, Alois Knoll, Guang Chen, and Changjun Jiang.
\newblock Lead: Learning decomposition for source-free universal domain adaptation.
\newblock In \emph{CVPR}, 2024.

\bibitem[Richter et~al.(2016)Richter, Vineet, Roth, and Koltun]{Richter2016PlayingFD}
Stephan~R. Richter, Vibhav Vineet, Stefan Roth, and Vladlen Koltun.
\newblock Playing for data: Ground truth from computer games.
\newblock In \emph{ECCV}, 2016.

\bibitem[Rombach et~al.(2021)Rombach, Blattmann, Lorenz, Esser, and Ommer]{Rombach2021HighResolutionIS}
Robin Rombach, A. Blattmann, Dominik Lorenz, Patrick Esser, and Bj{\"o}rn Ommer.
\newblock High-resolution image synthesis with latent diffusion models.
\newblock In \emph{CVPR}, 2021.

\bibitem[Ros et~al.(2016)Ros, Sellart, Materzynska, V{\'a}zquez, and L{\'o}pez]{Ros2016TheSD}
Germ{\'a}n Ros, Laura Sellart, Joanna Materzynska, David V{\'a}zquez, and Antonio~M. L{\'o}pez.
\newblock The synthia dataset: A large collection of synthetic images for semantic segmentation of urban scenes.
\newblock In \emph{CVPR}, 2016.

\bibitem[Sakaridis et~al.(2021)Sakaridis, Dai, and Van~Gool]{sakaridis2021acdc}
Christos Sakaridis, Dengxin Dai, and Luc Van~Gool.
\newblock Acdc: The adverse conditions dataset with correspondences for semantic driving scene understanding.
\newblock In \emph{CVPR}, 2021.

\bibitem[Seo et~al.(2019)Seo, Suh, Kim, Han, and Han]{Seo2019LearningTO}
Seonguk Seo, Yumin Suh, Dongwan Kim, Jongwoo Han, and Bohyung Han.
\newblock Learning to optimize domain specific normalization for domain generalization.
\newblock In \emph{ECCV}, 2019.

\bibitem[Song et~al.(2020)Song, Meng, and Ermon]{song2020denoising}
Jiaming Song, Chenlin Meng, and Stefano Ermon.
\newblock Denoising diffusion implicit models.
\newblock \emph{arXiv preprint arXiv:2010.02502}, 2020.

\bibitem[Venkat et~al.(2020)Venkat, Kundu, Singh, Revanur, et~al.]{venkat2020your}
Naveen Venkat, Jogendra~Nath Kundu, Durgesh Singh, Ambareesh Revanur, et~al.
\newblock Your classifier can secretly suffice multi-source domain adaptation.
\newblock \emph{NIPS}, 2020.

\bibitem[Wang et~al.(2024{\natexlab{a}})Wang, Yue, Zhou, Chan, and Loy]{wang2024exploiting}
Jianyi Wang, Zongsheng Yue, Shangchen Zhou, Kelvin~CK Chan, and Chen~Change Loy.
\newblock Exploiting diffusion prior for real-world image super-resolution.
\newblock \emph{IJCV}, 2024{\natexlab{a}}.

\bibitem[Wang et~al.(2018)Wang, Yu, Dong, and Loy]{wang2018recovering}
Xintao Wang, Ke Yu, Chao Dong, and Chen~Change Loy.
\newblock Recovering realistic texture in image super-resolution by deep spatial feature transform.
\newblock In \emph{CVPR}, 2018.

\bibitem[Wang et~al.(2024{\natexlab{b}})Wang, Pan, Shen, Shi, and Shi]{wang2024domain}
Zongbin Wang, Bin Pan, Shiyu Shen, Tianyang Shi, and Zhenwei Shi.
\newblock Domain generalization guided by large-scale pre-trained priors.
\newblock \emph{arXiv preprint arXiv:2406.05628}, 2024{\natexlab{b}}.

\bibitem[Wei et~al.(2024)Wei, Chen, Jin, Ma, Liu, Ling, Wang, Chen, and Zheng]{wei2024stronger}
Zhixiang Wei, Lin Chen, Yi Jin, Xiaoxiao Ma, Tianle Liu, Pengyang Ling, Ben Wang, Huaian Chen, and Jinjin Zheng.
\newblock Stronger fewer \& superior: Harnessing vision foundation models for domain generalized semantic segmentation.
\newblock In \emph{CVPR}, 2024.

\bibitem[Wu et~al.(2023)Wu, Zhao, Chen, Gu, Zhao, He, Zhou, Shou, and Shen]{wu2023datasetdm}
Weijia Wu, Yuzhong Zhao, Hao Chen, Yuchao Gu, Rui Zhao, Yefei He, Hong Zhou, Mike~Zheng Shou, and Chunhua Shen.
\newblock Datasetdm: Synthesizing data with perception annotations using diffusion models.
\newblock \emph{NeurIPS}, 2023.

\bibitem[Xia et~al.(2023)Xia, Zhang, Wang, Wang, Wu, Tian, Yang, and Van~Gool]{xia2023diffir}
Bin Xia, Yulun Zhang, Shiyin Wang, Yitong Wang, Xinglong Wu, Yapeng Tian, Wenming Yang, and Luc Van~Gool.
\newblock Diffir: Efficient diffusion model for image restoration.
\newblock In \emph{ICCV}, 2023.

\bibitem[Xu et~al.(2022)Xu, Yao, Jiang, Jiang, Chu, Han, Zhang, Wang, and Tai]{Xu2022DIRLDR}
Qi Xu, Lili Yao, Zhengkai Jiang, Guannan Jiang, Wenqing Chu, Wenhui Han, Wei Zhang, Chengjie Wang, and Ying Tai.
\newblock Dirl: Domain-invariant representation learning for generalizable semantic segmentation.
\newblock In \emph{AAAI}, 2022.

\bibitem[Yang et~al.(2023)Yang, Gu, and Sun]{Yang2023GeneralizedSS}
Liwei Yang, Xiang Gu, and Jian Sun.
\newblock Generalized semantic segmentation by self-supervised source domain projection and multi-level contrastive learning.
\newblock In \emph{AAAI}, 2023.

\bibitem[Yu et~al.(2018)Yu, Chen, Wang, Xian, Chen, Liu, Madhavan, and Darrell]{Yu2018BDD100KAD}
Fisher Yu, Haofeng Chen, Xin Wang, Wenqi Xian, Yingying Chen, Fangchen Liu, Vashisht Madhavan, and Trevor Darrell.
\newblock Bdd100k: A diverse driving dataset for heterogeneous multitask learning.
\newblock In \emph{CVPR}, 2018.

\bibitem[Yue et~al.(2019)Yue, Zhang, Zhao, Sangiovanni-Vincentelli, Keutzer, and Gong]{Yue2019DomainRA}
Xiangyu Yue, Yang Zhang, Sicheng Zhao, Alberto~L. Sangiovanni-Vincentelli, Kurt Keutzer, and Boqing Gong.
\newblock Domain randomization and pyramid consistency: Simulation-to-real generalization without accessing target domain data.
\newblock In \emph{ICCV}, 2019.

\bibitem[Zhang et~al.(2024{\natexlab{a}})Zhang, Su, Xu, and Jia]{zhang2024improving}
Haojie Zhang, Yongyi Su, Xun Xu, and Kui Jia.
\newblock Improving the generalization of segmentation foundation model under distribution shift via weakly supervised adaptation.
\newblock In \emph{CVPR}, 2024{\natexlab{a}}.

\bibitem[Zhang et~al.(2024{\natexlab{b}})Zhang, Liu, Tai, and Tang]{zhang2024c3net}
Juntao Zhang, Yuehuai Liu, Yu-Wing Tai, and Chi-Keung Tang.
\newblock C3net: Compound conditioned controlnet for multimodal content generation.
\newblock In \emph{CVPR}, 2024{\natexlab{b}}.

\bibitem[Zhang et~al.(2023)Zhang, Rao, and Agrawala]{zhang2023adding}
Lvmin Zhang, Anyi Rao, and Maneesh Agrawala.
\newblock Adding conditional control to text-to-image diffusion models.
\newblock In \emph{ICCV}, 2023.

\bibitem[Zhang and Tan(2025)]{zhang2025mamba}
Xin Zhang and Robby~T Tan.
\newblock Mamba as a bridge: Where vision foundation models meet vision language models for domain-generalized semantic segmentation.
\newblock In \emph{CVPR}, 2025.

\bibitem[Zhao et~al.(2025)Zhao, Li, Wang, Wu, Zang, Sebe, and Zhong]{zhao2025fishertune}
Dong Zhao, Jinlong Li, Shuang Wang, Mengyao Wu, Qi Zang, Nicu Sebe, and Zhun Zhong.
\newblock Fishertune: Fisher-guided robust tuning of vision foundation models for domain generalized segmentation.
\newblock In \emph{CVPR}, 2025.

\bibitem[Zhao et~al.(2023)Zhao, Chen, Chen, Bao, Hao, Yuan, and Wong]{zhao2023uni}
Shihao Zhao, Dongdong Chen, Yen-Chun Chen, Jianmin Bao, Shaozhe Hao, Lu Yuan, and Kwan-Yee~K Wong.
\newblock Uni-controlnet: All-in-one control to text-to-image diffusion models.
\newblock \emph{NeurIPS}, 2023.

\bibitem[Zhao et~al.(2022)Zhao, Zhong, Zhao, Sebe, and Lee]{Zhao2022StyleHallucinatedDC}
Yuyang Zhao, Zhun Zhong, Na Zhao, N. Sebe, and Gim~Hee Lee.
\newblock Style-hallucinated dual consistency learning for domain generalized semantic segmentation.
\newblock In \emph{ECCV}, 2022.

\bibitem[Zhong et~al.(2022)Zhong, Zhao, Lee, and Sebe]{Zhong2022AdversarialSA}
Zhun Zhong, Yuyang Zhao, Gim~Hee Lee, and N. Sebe.
\newblock Adversarial style augmentation for domain generalized urban-scene segmentation.
\newblock In \emph{NeurIPS}, 2022.

\end{thebibliography}
